\documentclass[twoside,11pt]{article}

\usepackage[preprint]{jmlr2e}

\usepackage{amsmath, bbm}
\usepackage{booktabs}
\usepackage{multirow}
\usepackage{float}
\usepackage{xr}

\newtheorem{assumption}{Assumption}

\makeatletter
\newcommand*{\addFileDependency}[1]{% argument=file name and extension
  \typeout{(#1)}
  \@addtofilelist{#1}
  \IfFileExists{#1}{}{\typeout{No file #1.}}
}
\makeatother

\newcommand*{\myexternaldocument}[1]{%
    \externaldocument{#1}%
    \addFileDependency{#1.tex}%
    \addFileDependency{#1.aux}%
}
\myexternaldocument{supplemental}

\jmlrheading{1}{2023}{1-48}{4/00}{10/00}{meila00a}{Giora Simchoni and Saharon Rosset}

\ShortHeadings{Integrating Random Effects in Deep Neural Networks}{Giora Simchoni and Saharon Rosset}
\firstpageno{1}

\begin{document}

\title{Integrating Random Effects in Deep Neural Networks}

\author{\name Giora Simchoni \email gsimchoni@tauex.tau.ac.il \\
        \name Saharon Rosset \email saharon@tauex.tau.ac.il \\
        \addr Department of Statistics and Operations Research\\
        Tel Aviv University\\
        Tel Aviv, Israel, 69978}

\editor{Kevin Murphy and Bernhard Sch{\"o}lkopf}

\maketitle

\begin{abstract}%   <- trailing '%' for backward compatibility of .sty file
Modern approaches to supervised learning like deep neural networks (DNNs) typically implicitly assume that observed responses are statistically independent. In contrast, correlated data are prevalent in real-life large-scale applications, with typical sources of correlation including spatial, temporal and clustering structures. These correlations are either ignored by DNNs, or ad-hoc solutions are developed for specific use cases. We propose to use the mixed models framework to handle correlated data in DNNs. By treating the effects underlying the correlation structure as random effects, mixed models are able to avoid overfitted parameter estimates and ultimately yield better predictive performance. The key to combining mixed models and DNNs is using the Gaussian negative log-likelihood (NLL) as a natural loss function that is minimized with DNN machinery including stochastic gradient descent (SGD). Since NLL does not decompose like standard DNN loss functions, the use of SGD with NLL presents some theoretical and implementation challenges, which we address.  Our approach which we call LMMNN is demonstrated to improve performance over natural competitors in various correlation scenarios on diverse simulated and real datasets. Our focus is on a regression setting and tabular datasets, but we also show some results for classification. Our code is available at https://github.com/gsimchoni/lmmnn.
\end{abstract}


\section{Introduction} \label{intro}
Linear mixed models (LMMs) and generalized linear mixed models (GLMMs) have long been researched in the statistical literature, with applications in medical statistics, geography, psychometry and more \citep[see e.g.][]{McCulloch2008}. \citet[chap.~2, Example 7]{searle1992variance} give a classic application of estimating the effect of three medications on blood pressure in patients from 15 randomly chosen clinics across New York City. In each clinic 20 patients are divided into 4 groups (three medications and a placebo), such that each patient is treated with a single treatment and the effect on blood pressure is measured. Estimating the effect of treatment while ignoring the correlation between two measurements of blood pressure from the same clinic, or treating each of the effects of clinics as fixed, might lead to overfitted estimates~\citep{Robinson1991}. When modeling these data using LMM, each clinic receives its own random effect (RE) in the model, which is a random variable with a common predefined zero-mean distribution and a variance component to estimate, reflecting the researcher’s assumption that the clinics participating in the experiment are a random sample taken from a population of clinics, and that they themselves are not of interest. The resulting treatment effect estimate should have lower variance than an estimate which ignores the correlation within each clinics’s measurements, and if a true treatment effect exists in the population, it would be easier to detect~\citep{McCulloch2008}.

However, even though this statistical principle has been well understood for years, it seems to have been ignored in modern machine learning approaches to statistical learning such as ensemble trees and deep neural networks (DNNs). Typically, within these frameworks, models assume observations to be statistically independent \citep[see e.g.][]{Sela2012}. There are numerous scenarios, where modeling data using LMM and GLMM might improve the predictive performance of modern machine learning tools. In our recent work~\citep{lmmnn_neurips} we focused on one such scenario of handling high-cardinality categorical features in a regression setting. Our approach, which we call LMMNN, uses the negative log-likelihood (NLL) as a natural loss function, on top of almost any DNN architecture to learn a pair of functions: fixed and random. Handling such clustered data by adapting mixed modeling methodology to be used within DNNs while minimizing some form of NLL is the subject of several other papers. These include MeNets~\citep{Xiong_2019_CVPR} and DeepGLMM~\citep{DeepGLMM} which are reviewed in Section~\ref{related:categorical}. Yet, none of the aforementioned papers, including our own, were concerned with more complex mixed effects correlation scenarios which are prevalent in modern modeling tasks. For example, in \citet{Duan2014} the authors discussed the challenge of imputing the traffic flow for missing freeway detectors at a certain period of time. The input to the network was the traffic flow of $m$ other such detectors, in this case $m = 15K$ detectors across the state of California. While the authors ignored the spatial relations between detectors relying on stacked auto-encoders (SAE) to encode and decode these data, a mixed effects DNN might posit a proper covariance structure on the data points in space, for example using a squared exponential kernel on the pairwise distances between detectors.

Another type of data for which LMM and GLMM could be beneficial is longitudinal data exhibiting temporal dependence. In a recent study \citet{Lin2019} tried to predict hospital readmission from electronic medical records (EMR) of hospital patients, where each patient is measured hourly for various metrics such as blood pressure, 48 hours before discharge. To handle the temporal correlation between these measurements \citeauthor{Lin2019} chose to use a LSTM-based recurrent neural network. Yet it is not clear that such a short time series necessitates such a complex model which was developed for longer and more varied sequences such as word sentences and paragraphs. A GLMM-inspired network which would model the binary result of readmission, could handle the blood pressure sequence by adding one or two additional variance components parameters to estimate, for an added random slope at time $t$, or perhaps an additional quadratic term at $t^2$.

As said, such treatments of correlated data in neural networks are rare, and there is a growing need to generalize approaches like LMMNN to handle this and other complex correlation settings. The current paper takes a leap forward from our previous paper \citep{lmmnn_neurips}, as we generalize LMMNN to more complex LMM scenarios and discuss at length theoretical issues of LMMNN convergence. The paper is organized as follows: The rest of Section~\ref{intro} reviews in short the standard LMM approach to regression and some typical covariance structures. Section~\ref{proposed} describes our approach to LMM in DNNs, LMMNN. In Section~\ref{theory} we further elaborate on the conditions and covariance matrices under which the stochastic gradient descent (SGD) approach used by LMMNN is promised to converge, building on theoretical work by \citet{Chen2020}. Section~\ref{related} gives a brief overview of other attempts at incorporating random effects in DNNs to handle correlated data. In Section~\ref{results} we show results on simulated as well as real datasets, demonstrating the usefulness of LMMNN in common DNN prediction tasks and its superiority over other common solutions to handle such datasets. Section~\ref{classification} introduces GLMM for classification settings and a preliminary but successful attempt at implementing this in the LMMNN spirit. Lastly in Section~\ref{discussion} we discuss directions for future research.

\subsection{LMM: A Short Review} \label{intro:review}

In a typical LMM setting $y \in \mathbb{R}^n$ is a dependent variable modeled by $X$ and $Z$, which are $n \times p$ and $n \times q$ model matrices respectively:
\begin{equation}\label{eq:classicLMM}
	y = X\beta + Zb + \varepsilon.
\end{equation}
Here $\beta \in \mathbb{R}^p$ is a vector of {\em fixed} model parameters or {\em effects}, $\varepsilon \in \mathbb{R}^n$ is normal i.i.d noise or $\varepsilon \sim \mathbb{N}\left(0, \sigma^2_e I\right)$, and $b \in \mathbb{R}^q$ is a vector of {\em random effects}, meaning random variables. Typically $b$ is assumed to have a multivariate normal distribution $\mathbb{N}\left(0, D\right)$ where $D$ is a $q \times q$ positive semi-definite matrix of appropriate structure, holding usually unknown {\em variance components} to be estimated, let these be $\psi$, so $D$ could be written as $D(\psi)$. The structure of this covariance matrix is up to the researcher but there are typically simplified structures used. It is further assumed that there is no dependence between the normal noise and the random effects, that is $\text{cov}\left(\varepsilon, b\right) = 0$. 

We write the marginal distribution of $y$ as:
\begin{equation}\label{eq:classicMarginal}
	y \sim \mathbb{N}\left(X\beta, V(\theta)\right),
\end{equation}
where $V(\theta) = ZD(\psi)Z' + \sigma^2_e I$ and $\theta$ is the vector of all variance components $[\sigma^2_e, \psi]$. To fit $\beta, \theta$ we use maximum likelihood estimation (MLE), where we maximize the log-likelihood or equivalently minimize the negative log-likelihood (NLL):
\begin{equation}\label{eq:classicNLL}
	NLL(\beta, \theta | y) = \frac{1}{2}\left(y - X\beta\right)'V(\theta)^{-1}\left(y - X\beta\right) + \frac{1}{2}\log{|V(\theta)|} + \frac{n}{2}\log{2\pi}
\end{equation}

To predict $\hat{y}_{te}$ in a machine learning scenario, where $\left(X, Z, y\right)$ are typically split into training and testing sets $\left(X_{tr}, Z_{tr}, y_{tr}\right)$ and $\left(X_{te}, Z_{te}, y_{te}\right)$, one would use $y$'s fitted conditional mean:
\begin{equation}\label{eq:classicPred}
	\hat{y}_{te} = X_{te}\hat{\beta} + Z_{te}\hat{b},
\end{equation}
where $\hat{\beta} = (X_{tr}'\hat{V}^{-1}X_{tr})^{-1}X_{tr}'\hat{V}^{-1}y_{tr}$ are the estimated fixed effects once the estimated variance components $\hat{\theta}$ are input into $V$, and:
\begin{equation}\label{eq:classicBLUP}
    \hat{b} = \hat{D}Z_{tr}'V(\hat{\theta})^{-1}\left(y_{tr} - X_{tr}\hat{\beta}\right)
\end{equation}
is the so called {\em best linear unbiased predictor} (BLUP), as $b$ are not actually parameters to be estimated, but random variables to be predicted.

The LMM framework may suffer from a few drawbacks. Sometimes, calculating \eqref{eq:classicPred} is not possible such as in the case of the random intercepts model as in Section~\ref{intro:structures:single} with a single categorical feature with $q$ levels, where $Z_{te}$ holds levels unseen before. In this case it is customary to use $y$'s marginal distribution and predict $\hat{y}_{te}$ to be $X_{te}\hat{\beta}$, without the random part. More difficulty may arise when computing the BLUP in \eqref{eq:classicBLUP} and the NLL in \eqref{eq:classicNLL} if $n$ is so large that inverting $V(\hat{\theta})$ is infeasible, though see Section~\ref{intro:structures} and comments at the end of Section~\ref{proposed} for considerable speedups when implementing these computations for specific covariance structures. Another major and obvious drawback of LMM is the limitation to linear relationships, and indeed non-linear mixed models have been developed \citep[see e.g.][]{nlme_Lindstrom}. Finally, basic LMM as presented here is targeted towards modeling continuous response $y$, with a conditional normal distribution as in \eqref{eq:classicMarginal}. When $y$ is not continuous (for example, binary as in two-class classification), the commonly used extension is GLMM \citep{McCulloch2008}. We return to this in Section~\ref{classification}, where we discuss adapting LMMNN to classification.

\subsection{LMM: Covariance Structures}\label{intro:structures}
There are a few typical specialized models used in LMM, stemming from different choices for covariance structure in $D(\psi)$. It is worth reviewing these here since in Section~\ref{results} we show many results using these specific models, on simulated and real datasets.

\subsubsection{Single categorical feature: random intercepts}
\label{intro:structures:single}
The random intercepts model is appropriate for a single RE categorical variable of $q$ levels. In our previous work \citep{lmmnn_neurips} we demonstrated how this model is especially useful for handling high-cardinality categorical features in DNNs. The $Z$ matrix of dimension $n \times q$ is a binary matrix where $Z_{ij}=1$ means that observation $i$ has level $j$ of the categorical variable, and $Z_{ij}=0$ otherwise, meaning each row has a single non-zero entry. Therefore, we can mark the $l$-th measurement of level $j$ ($j = 1, \dots, q; l = 1, \dots, n_j$) as $y_{lj}$ and write model \eqref{eq:classicLMM} in scalar form:
\begin{equation}\label{eq:randomIntercepts}
    y_{lj} = \beta_0 + \beta'x_{lj} + b_j + \varepsilon_{lj}
\end{equation}
This nicely shows how for each level $j$ of the categorical feature we have an additional {\em random intercept} term $b_j$, hence the model's name. The term $b_j$ is distributed $\mathbb{N}(0, \sigma^2_b)$, where $\sigma^2_b$ is a single variance component so $\psi = \sigma^2_b$, and $D(\psi) = \sigma^2_bI_q$ is diagonal, making $y$'s marginal covariance matrix $V(\theta)$ block diagonal, since $V(\theta) = \sigma^2_bZZ' + \sigma^2_eI_n$. This in turn allows to avoid its inversion when computing \eqref{eq:classicNLL} or \eqref{eq:classicBLUP}. In fact, it can be shown that for a given level $j$ the computation of the BLUP is reduced to:
\begin{equation}\label{eq:classicRandomInterceptsPrediction}
	\hat{b}_j = \frac{n_j\hat{\sigma}^2_b}{\hat{\sigma}^2_e + n_j\hat{\sigma}^2_b}\left(\bar{y}_{tr; j} - \overline{X_{tr}\beta}_j\right),
\end{equation}
where $(\hat{\sigma}^2_e, \hat{\sigma}^2_b)$ are the estimated variance components, $n_j$ is the number of observations in level $j$ and $\bar{y}_{tr; j}$ and $\overline{X_{tr}\beta}_j$ are the observed and predicted average values of $y$ in cluster $j$ respectively.

\subsubsection{Multiple categorical features}
\label{intro:structures:multiple}
In the case of $K$ categorical RE variables, each of $q_k$ levels, the $Z$ matrix may be seen as a concatenation of $K$ binary matrices $Z_k$ of dimension $n \times q_k$, to form a binary matrix of dimension $n \times M$, where $M = \sum_k q_k$. The vector of REs $b$ is of length $M$ and is distributed $\mathbb{N}(0, D(\psi))$ where $D(\psi)$ is of dimension $M \times M$. If there are correlations between the $K$ variables they would be considered as part of the variance components to estimate and appear in the off diagonal elements of $D(\psi)$. Otherwise $D(\psi)$ is diagonal and $\psi = [\sigma^2_{b1}, \dots, \sigma^2_{bK}]$. As for the marginal covariance matrix of $y$, even when the $K$ categorical variables are assumed uncorrelated, $V(\theta)$ is no longer block-diagonal:
\begin{equation}\label{eq:multipleCategoricalV}
    V(\theta) = \sum_k \sigma^2_{bk}Z_kZ'_k + \sigma^2_eI_n
\end{equation}

\subsubsection{Longitudinal data and repeated measures}
\label{intro:structures:longitudinal}
In many applications we see repeated measures of the same unit of interest, typically one of $q$ subjects who are being monitored for some continuous measure $y$ through time. In this case it is often assumed observations have temporal correlation, and the longitudinal LMM model is used to predict $y$ at different times. In scalar form for the $l$-th measurement of subject $j$ could be modeled with a polynomial of time $t_{lj}$:
\begin{equation}\label{eq:longitudinalScalar}
    y_{lj} = \beta_0 + \beta'x_{lj} + b_{0,j} + b_{1,j} \cdot t_{lj} + b_{2,j} \cdot t^2_{lj} + \dots + b_{K-1,j} \cdot t^{K - 1}_{lj} + \varepsilon_{lj}
\end{equation}
A measurement of subject $j$ ($j = 1, \dots, q$) at time $t_{lj}$ has a random intercept $b_{0,j}$, a random slope $b_{1,j}$, and so on until the polynomial order $K - 1$. Each $b_{k,j}$ term is distributed $\mathbb{N}(0, \sigma^2_{b,k})$. The model is also flexible enough to have fixed variables from $X$ varying in time or to include fixed terms in $\beta$ for time $t_{lj}$. Now assume $t$ is the full $n$-length vector of times. Let $Z_0$ be the $n \times q$ binary matrix where the $\left[l, j\right]$-th entry holds 1 if subject $j$ was measured at time $l$. The full $Z$ would be of dimension $n \times Kq$ for $K$ polynomial terms and $q$ subjects. $Z$ would be a concatenation of $K$ matrices: $[Z_0 \vdots Z_1 \vdots \dots \vdots Z_{K - 1}]$ where each $Z_k = diag(t^{k}) \cdot Z_0$ for $k = 0, \dots K - 1$. Note that on the $\left[l, j\right]$-th entry $Z_k$ will have $t^{k}$ if subject $j$ has measurement in time $t_l$ or 0 else. $b$ of length $Kq$ is still distributed normally, its covariance matrix $D(\psi)$ is of dimension $Kq \times Kq$ with $\sigma^2_{b,0}I_q, \dots, \sigma^2_{b,K-1}I_q$ on the diagonal. If the RE terms are correlated there are additional correlation parameters to estimate on its off-diagonal, otherwise $\psi = [\sigma^2_{b,0}, \dots, \sigma^2_{b,K-1}]$ and $D(\psi)$ is diagonal. In general it can be shown that $V(\theta)$, the marginal covariance matrix of $y$, is block-diagonal. We expand on this in Section~\ref{theory}.

\subsubsection{Kriging or spatial data}
\label{intro:structures:spatial}
Suppose some continuous measurement $y$ changes across a N-dimensional random field $\mathcal{S}$. For each element $s \in \mathcal{S}$ (say a point in space and time), $y(s)$ is the sum of a ``deterministic'' component $\mu$ and a ``stochastic'' component $e$, functions of the ``location'' element $s$ and other properties $x \in \mathbb{R}^p$ and we write: $y(s) = \mu(x, s) + e(s) + \varepsilon$. Here $\mu$ could be a constant mean or a $x'\beta$ regression-like sum which does not depend on element $s$, and $e(s)$ is usually an additive variable which is distributed Gaussian, with zero mean and some covariance matrix. Usually the covariance is assumed to decay as distance between elements $h_{ij} = |s_i - s_j|$ increases. If the covariance is isotropic, meaning it depends only on $h_{ij}$ and covariance decays in the same pattern in all directions, we could write: $cov(y(s_i), y(s_j)) = f(h_{ij})$, where $f$ is sometimes called the kernel function, typically denoted as $k(s_i, s_j)$. The most common kernel is the \textit{radial basis function} (RBF) kernel, or squared exponential:
\begin{equation}\label{eq:RBFKernel}
	\text{cov}(y(s_i), y(s_j)) = \tau^2 \cdot \exp\left(- \frac{h^2_{ij}}{2l^2}\right)
\end{equation}
where $\tau^2$ is a variance parameter and $l^2$ a ``range'' or ``lengthscale'' rate-of-decay parameter to estimate. As the distance $h_{ij}$ increases the covariance decreases, potentially very quickly, depending on the kernel used and parameter values.

The above describes the model behind kriging, Gaussian processes (GP) and spatial analysis, which are very similar at their core (see e.g. \cite{Rasmussen} and \cite{Cressie1993}). However it is also a description of \eqref{eq:classicLMM} with $Z_{n \times q}$ a binary matrix of $q$ locations, and $b$ having covariance matrix $D(\psi)$ of dimension $q \times q$:
\begin{equation}\label{eq:RBFD}
    D_{ij}(\psi) = \sigma^2_{b0} \cdot \exp\left(- \frac{|s_i - s_j|^2}{2\sigma^2_{b1}}\right),
\end{equation}
where $\psi = [\sigma^2_{b0}, \sigma^2_{b1}]$ and $s_i, s_j$ are again N-dimensional locations. Usually N is 2 (often latitude and longitude) or 3 (often latitude, longitude and time). Here, the marginal covariance matrix of $y$ does not have any sparse structure.

\section{LMMNN: Proposed Approach}\label{proposed}
We start with redefining model \eqref{eq:classicLMM} by allowing both fixed and random parts to have non-linear relations to $y$:
\begin{equation}\label{eq:lmmnnModel}
	y = f\left(X\right) + g\left(Z\right)b + \varepsilon,
\end{equation}
where $f$ and $g$ are non-linear complex functions which we fit using DNNs. Note that $f$ and $g$ are kept as general as possible, to allow any acceptable DNN architecture, including convolutional and recurrent neural networks, as previously demonstrated in \citet{lmmnn_neurips}. An additional example to what $g$ could be is given in Section~\ref{results:simulated:spatial} for the spatial data case, where we pass the 2-D locations $s_i, s_j$ through a multilayer perceptron (MLP) which has 1000 neurons in its final layer. Thus, $g$ here is embedding of the 2-D locations to dimension 1000.

\begin{figure}
	\centering
	\includegraphics[width=0.6\linewidth]{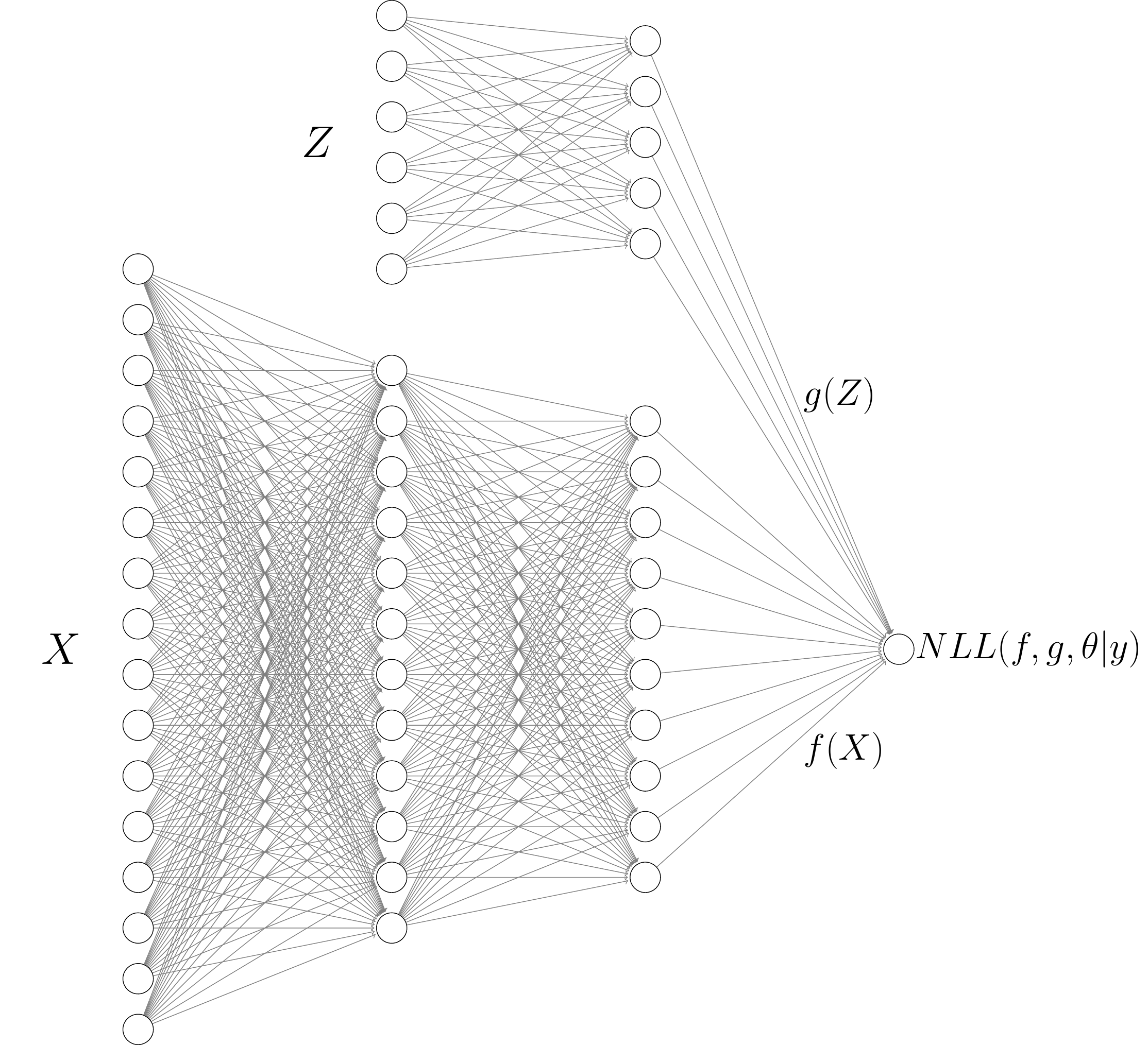}
	\caption{Schematic description of LMMNN using a simple deep MLP for fitting $f$ and $g$, and combining outputs with the NLL loss layer, in a single-stage training.}
	\label{fig:lmmnn_nlp}
\end{figure}

Next we modify the NLL loss criterion \eqref{eq:classicNLL} to include the DNN outputs $f$ and $g$:
\begin{equation}\label{eq:lmmnnNLL}
	NLL(f, g, \theta | y) = \frac{1}{2}\left(y - f\left(X\right)\right)'V(g, \theta)^{-1}\left(y - f\left(X\right)\right) + \frac{1}{2}\log{|V(g, \theta)|} + \frac{n}{2}\log{2\pi},
\end{equation}
where $V(g, \theta) = g(Z)D(\psi)g(Z)' + \sigma^2_e I_n$. We call DNNs using this NLL loss criterion LMM neural networks or LMMNN. See Figure~\ref{fig:lmmnn_nlp} for a schematic description of LMMNN, in the case $f$ and $g$ are approximated with a simple MLP. Note how $f$ and $g$ can be represented using the same network architecture, two different architectures, and in many real data experiments we found it useful to have $g$ as the identity function, that is to say, not learning any transformation for the data in $Z$.

At each epoch we use SGD on mini-batches to optimize the network's weights including the variance components $\theta$ which are treated as additional network parameters. For a mini-batch $\xi$ of size $m$ comprised of $(X_\xi, Z_\xi, y_\xi)$ we choose to define a version of the NLL criterion in \eqref{eq:lmmnnNLL}, using the inverse of the sub-matrix $V(g, \theta)_\xi = g(Z_\xi)D(\psi)g(Z_\xi)' + \sigma^2_e I_m$ instead of the sub-matrix of the inverse $(V(g, \theta)^{-1})_\xi$ as formal SGD would require (see discussion in Section~\ref{theory}):
\begin{equation}\label{eq:lmmnnSGDNLL}
	NLL_\xi(f, g, \theta | y_\xi) = \frac{1}{2}\left(y_\xi - f\left(X_\xi\right)\right)'V(g, \theta)_\xi^{-1}\left(y_\xi - f\left(X_\xi\right)\right) + \frac{1}{2}\log{|V(g, \theta)_\xi|} + \frac{m}{2}\log{2\pi}.
\end{equation}
The partial derivative of $NLL_\xi$ with respect to the variance components can be written explicitly:
\begin{equation}\label{eq:lmmnnSGDGradient}
\frac{\partial NLL_\xi}{\partial \theta} = -\frac{1}{2}\left(y_\xi - f\left(X_\xi\right)\right)'V_{\xi}^{-1}\frac{\partial V_\xi}{\partial \theta}V_{\xi}^{-1}\left(y_\xi - f\left(X_\xi\right)\right) + \frac{1}{2}\text{tr}\left(V_{\xi}^{-1}\frac{\partial V_\xi}{\partial \theta}\right),
\end{equation}
where we further shorten $V(g, \theta)_\xi$ as $V_\xi$ and the $\frac{\partial V_\xi}{\partial \theta}$ expressions might further be simplified. In practice, we use existing DNN machinery to fit the network, mainly those of back-propagation and SGD.

It is worth emphasizing at this stage looking at \eqref{eq:lmmnnSGDNLL} and \eqref{eq:lmmnnSGDGradient} that for each mini-batch $\xi$ the $V_{\xi}$ inversion and computation of log-determinant no longer involve a matrix of size $n \times n$ but a matrix of size $m \times m$ where $m$ is the batch size and typically $m \ll n$. This ``inversion in parts'' is the key element behind LMMNN's scalability, and therefore we further expand on it in the next Section and conduct experiments to demonstrate this scalability in Section~\ref{results:simulated:scale}. We further note that even with this decrease in dimensionality a smart implementation does not necessitate an actual inversion of $V_\xi$ in $\eqref{eq:lmmnnSGDNLL}$. Rather, if we mark $e = y_\xi - f\left(X_\xi\right)$, we need to solve a linear system of equations $V_\xi x = e$ to get $V_{\xi}^{-1}\left(y_\xi - f\left(X_\xi\right)\right)$ directly, which further speeds up computations and stability and allows for larger batch sizes.

While training is performed on $(X_{tr}, Z_{tr}, y_{tr})$, prediction of $y_{te}$ from $(X_{te}, Z_{te})$ is made using:
\begin{equation}\label{eq:lmmnnPred}
	\hat{y}_{te} = \hat{f}\left(X_{te}\right) + \hat{g}\left(Z_{te}\right)\hat{b},
\end{equation}
where $\hat{f}$ and $\hat{g}$ are the outputs of the DNNs used to approximate $f$ and $g$, and $\hat{b}$ is the modified version of the BLUP from \eqref{eq:classicBLUP}:
\begin{equation}\label{eq:lmmnnBLUP}
    \hat{b} = D(\hat{\psi})\hat{g}\left(Z_{tr}\right)'V(\hat{g}, \hat{\theta})^{-1}\left(y_{tr} - \hat{f}\left(X_{tr}\right)\right).
\end{equation}
Now $V(\hat{g}, \hat{\theta})$ is again of dimension $n \times n$ and one needs to calculate its inverse once. In case of the random intercepts model with a single categorical feature with $q$ levels and $g$ is the identity function, the formula in \eqref{eq:classicRandomInterceptsPrediction} can be accommodated as in \citet{lmmnn_neurips} and no inversion is necessary. In the case of multiple categorical features, the random slopes model or in general a longitudinal repeated-measures model and $g$ is the identity function, $V(\hat{\theta})$ is relatively sparse and we can take advantage of that. We mark $e = y_{tr} - \hat{f}\left(X_{tr}\right)$ and solve the linear system of equations $V(\hat{\theta}) x = e$ to get $V(\hat{\theta})^{-1}\left(y_{tr} - \hat{f}\left(X_{tr}\right)\right)$ directly. It is only when $V(\hat{g}, \hat{\theta})$ is not sparse, such as in the case when $g$ is not the identity function or when using the spatial model, and $n$ is very large, that we need to resort to different solutions for computing the inverse. In our implementation we find a simple sampling approach works well, other more sophisticated sampling approaches or sparse approximations such as the inducing points method \citep{JMLR:v6:quinonero-candela05a} may be used.

\section{LMMNN: Justifying the SGD Mini-batch Approximation}\label{theory}

\begin{figure}
	\centering
	\includegraphics[width=1.0\linewidth]{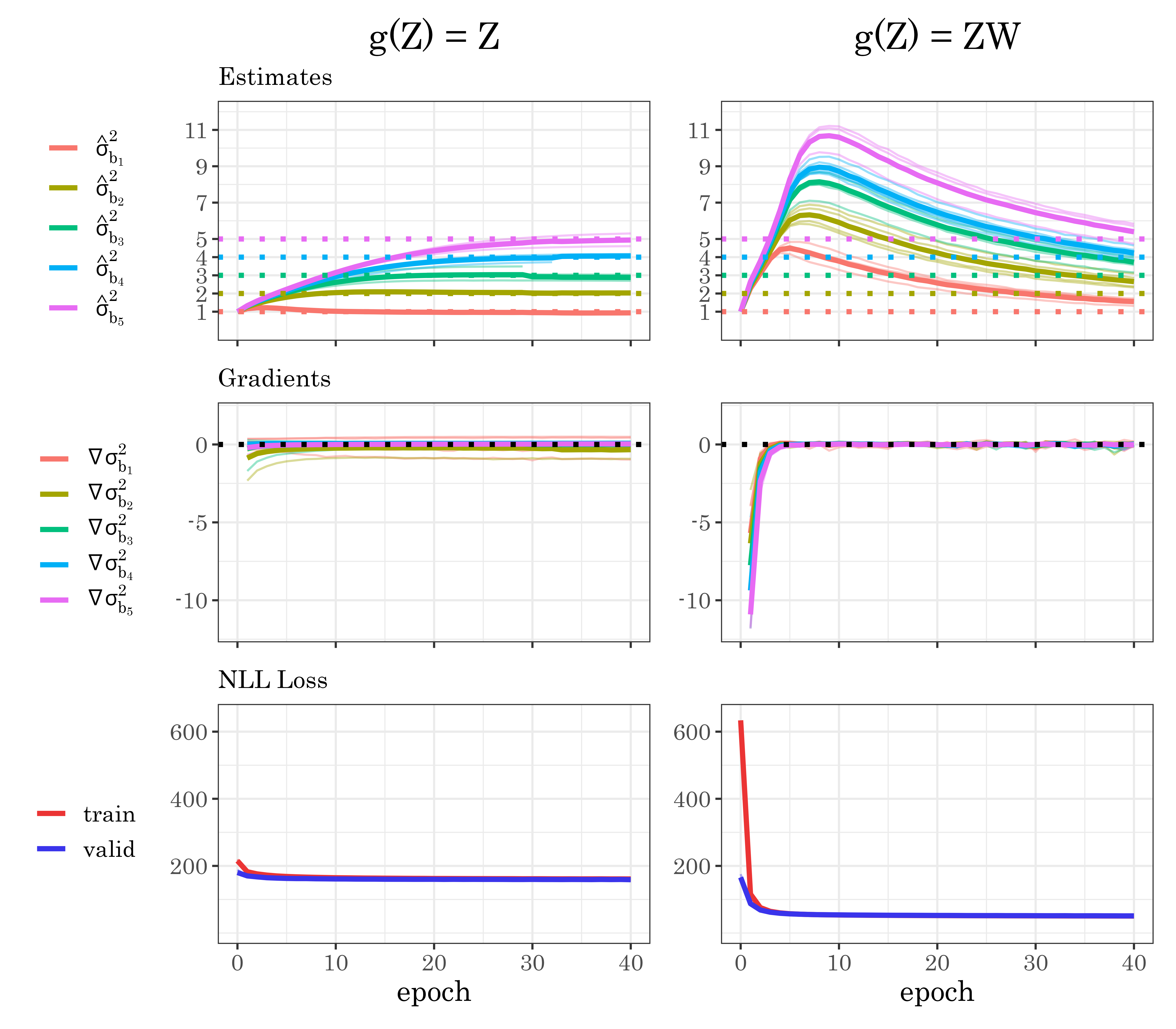}
	\caption{A LMMNN simulation with 5 uncorrelated categorical features each with $q = 1000$ and $\sigma^2_{bj} = j$ for $j = 1, \dots, 5$. $n = 100000$, $\sigma^2_e = 1$, there are $p = 10$ fixed features in $X$ and $f(X)$ and network architecture are as described in Section~\ref{results:simulated}. From top to bottom: $\sigma^2_{bj}$ estimates, $\sigma^2_{bj}$ gradients and NLL through epochs. The experiment was repeated five times, and the five results are shown as light lines, bold lines are average. Left: $g(Z) = Z$, Right: $g(Z) = ZW$, where $W$ is a $5,000 \times 500$ random $\mathbb{U}(-1,1)$ matrix.}
	\label{fig:lmmnn_convergence}
\end{figure}

In Section~\ref{proposed} we explicitly define in \eqref{eq:lmmnnSGDNLL} $NLL_\xi$ -- the NLL version using a mini-batch $\xi$ of size $m$. In each batch iteration, we calculate the inverse of the $m \times m$ sub-matrix $V(g, \theta)_\xi$ instead of the sub-matrix of the $n \times n$ inverse $(V(g, \theta)^{-1})_\xi$. This ``inversion in parts'' is the key element behind LMMNN's scalability as demonstrated in Section~\ref{results:simulated:scale}, however it bears some justification as it does not in general result in the full $n \times n$ inverse for any symmetric matrix $V$, unless $V$ is block-diagonal with blocks of size $m$. To demonstrate, in Figure~\ref{fig:lmmnn_convergence} we profile LMMNN's performance in terms of variance components estimates and gradients, and NLL loss, in a multiple high-cardinality categorical features scenario. Here $n = 100000$ observations are simulated according to model \eqref{eq:lmmnnModel}, in identical manner to simulations in Section~\ref{results:simulated}. There are $K = 5$ categorical RE features, each with $q = 1000$ levels, so $Z$ is of dimension $100000 \times 5000$. There are $p = 10$ fixed features in $X$, and $f(X)$ is a complex non-linear function as in \eqref{eq:f}. $g(Z)$ is either the identity function (left) or a linear mapping to a lower dimension (right), $g(Z) = ZW$ where $W$ is a $5000 \times 500$ random matrix with values sampled from a $\mathbb{U}(-1,1)$ distribution. We use SGD with $NLL_\xi$ approximation as in \eqref{eq:lmmnnSGDNLL}, a simple MLP architecture, and record each $\hat{\sigma}^2_{bj}$ ($j = 1, \dots, 5$) estimate and gradient at the end of each epoch. As described in Section~\ref{intro:structures:single} the $V(g, \theta)$ marginal covariance matrix is not block-diagonal, and with $g(Z) = ZW$ it is not even sparse. Yet, it is clear that LMMNN's use of SGD, and in particular the inversion of $V(g, \theta)$ and calculating its log-determinant $\log{|V(g, \theta)|}$ from \eqref{eq:lmmnnNLL} ``in parts" works well in the sense of estimates converging to their true parameters, gradients approaching zero and NLL loss decreasing. This Section's purpose is to offer intuition and some mathematical rigor to this phenomenon.

\subsection{Block-diagonal covariance matrix: when the gradient decomposes}\label{theory:deomposition}
Consider the case of a simple random intercepts model: a single categorical feature with $q$ levels each having $n_j$ observations $(X_j, Z_j, y_j)$, where $j = 1, \dots q$, and let $g$ be the identity function. As said above in this setting $V(\theta) = \sigma^2_b ZZ' + \sigma^2_e I$ is a block diagonal matrix and we can write $V(\theta) = diag(V_1, ..., V_q)$ where each $V_j$ block is of size $n_j \times n_j$ and $V_j = \sigma^2_b J_{n_j} + \sigma^2_e I_{n_j}$ where $J_{n_j}$ is a $n_j \times n_j$ all $1$s matrix. This means we can write the inverse in \eqref{eq:lmmnnNLL} as block diagonal as well, $V(\theta)^{-1} = diag(V^{-1}_1, ..., V^{-1}_q)$, and the log determinant in \eqref{eq:lmmnnNLL} as a sum of log determinants: $\log{|V(\theta)|} = \sum_{j = 1}^{q}{\log|V_j|}$. The NLL in \eqref{eq:lmmnnNLL} can now be written as a sum: $NLL(f, \theta | y) = \sum_{j = 1}^{q}{\frac{1}{2}\left(y_j - f\left(X_j\right)\right)'V_j^{-1}\left(y_j - f\left(X_j\right)\right) + \frac{1}{2}\log{|V_j|} + \frac{n_j}{2}\log{2\pi}}$. Most importantly, the full variance components gradient in \eqref{eq:lmmnnSGDGradient} can be decomposed into a sum of gradients: 
\begin{equation}\label{eq:randomInterceptsGradientDecomposition}
    \frac{\partial NLL}{\partial \theta} = \sum_{j = 1}^{q}\left[-\frac{1}{2}\left(y_j - f\left(X_j\right)\right)'V_j^{-1}\frac{\partial V_j}{\partial \psi}V_j^{-1}\left(y_j - f\left(X_j\right)\right) \\
        + \frac{1}{2}\text{tr}\left(V_j^{-1}\frac{\partial V_j}{\partial \psi}\right)\right]
\end{equation}
    
Thus if say $n_j = m$ for all $j$ and $m$ is a reasonable batch size, we can choose our mini-batches as the levels of the RE variable. For each mini-batch $\xi_k$, $(X_{\xi_k}, Z_{\xi_k}, y_{\xi_k})$ \textit{are} $(X_j, X_j, y_j)$ without stochasticity, and computing the gradient in parts and summing is identical to computing the whole gradient. If $n_j \neq m$ for all $j$ but all $n_j$ are small, we could have the batch size vary for each $j$. 

There are additional cases where the gradient naturally decomposes. For the case of random intercepts in GLMM see Section~\ref{classification}. Another case is the longitudinal model \eqref{eq:longitudinalScalar}, where $g$ is the identity function and $Z$ of dimension $n \times Kq$ and $Z_0, \dots, Z_{K-1}$ are defined in Section~\ref{intro:structures:longitudinal}. $D(\psi)$ is of dimensions $Kq \times Kq$ and we can decompose it to sub-matrices:
    
$D(\psi) =
\begin{pmatrix}
	\sigma^2_{b_0}I_q & \rho_{0,1}\sigma_{b_0}\sigma_{b_1} I_q  & \dots & \rho_{0,K-1}\sigma_{b_0}\sigma_{b_{K-1}} I_q\\
	\rho_{0,1}\sigma_{b_0}\sigma_{b_1} I_q &  \sigma^2_{b_1}I_q & \dots & \rho_{1,K-1}\sigma_{b_1}\sigma_{b_{K-1}} I_q \\
	\vdots & \vdots & \ddots & \vdots \\
	\rho_{0, K-1}\sigma_{b_0}\sigma_{b_K} I_q & \rho_{1, K-1}\sigma_{b_1}\sigma_{b_{K-1}} I_q  & \dots & \sigma^2_{b_{K-1}}I_q \\
\end{pmatrix}$
	
Or more compactly:
	
$D(\psi) =
\begin{pmatrix}
	D_{0,0} & \dots & D_{0,K-1}\\
	D_{1,0} & \dots & D_{1,K-1} \\
	\vdots & \ddots & \vdots \\
	D_{K-1,0} & \dots & D_{K-1,K-1} \\
\end{pmatrix}$
    
Now we can compose $V(\theta)$ into a sum of matrices:
\begin{equation}\label{eq:longitudinalDeomposing-V}
    V(\theta) = ZD(\psi)Z' + \sigma^2_eI_n = \sum_{l = 0}^{K-1}\sum_{m = 0}^{K-1}{Z_lD_{l,m}Z_m'} + \sigma^2_eI_n
\end{equation}
    
If $Z_0$ is sorted, in the sense that all of subject $j$'s measurements are in adjacent rows and subjects are ordered from 1 to $q$, then every $Z_k$ is sorted and each of the $Z_lD_{l,m}Z_m'$ matrices is block-diagonal \textit{with the same blocks}. Since $\sigma^2_eI_n$ is diagonal, $V(\theta)$ is also block-diagonal. Therefore, the decomposition of the full gradient in \eqref{eq:lmmnnSGDGradient} to the sum of $q$ subjects sub-gradients, will also hold.
	
For the multiple uncorrelated categorical random intercepts model, $V(\theta)$ would not in general be block-diagonal as explained in Section~\ref{intro:structures:multiple}. A more limiting but not uncommon structure of the categorical features is when they are \textit{nested}, for example the first feature is which school a student goes to and the second is which class in that school she goes to. In this case $V(\theta)$ will be block-diagonal, the block sizes corresponding to the highest level in the categorical variables hierarchy, that is the school in this example, and the gradient can be decomposed.

\subsection{Block-diagonal approximation of covariance matrix}
\label{theory:approximation}
\begin{figure}
	\centering
	\includegraphics[width=1.0\linewidth]{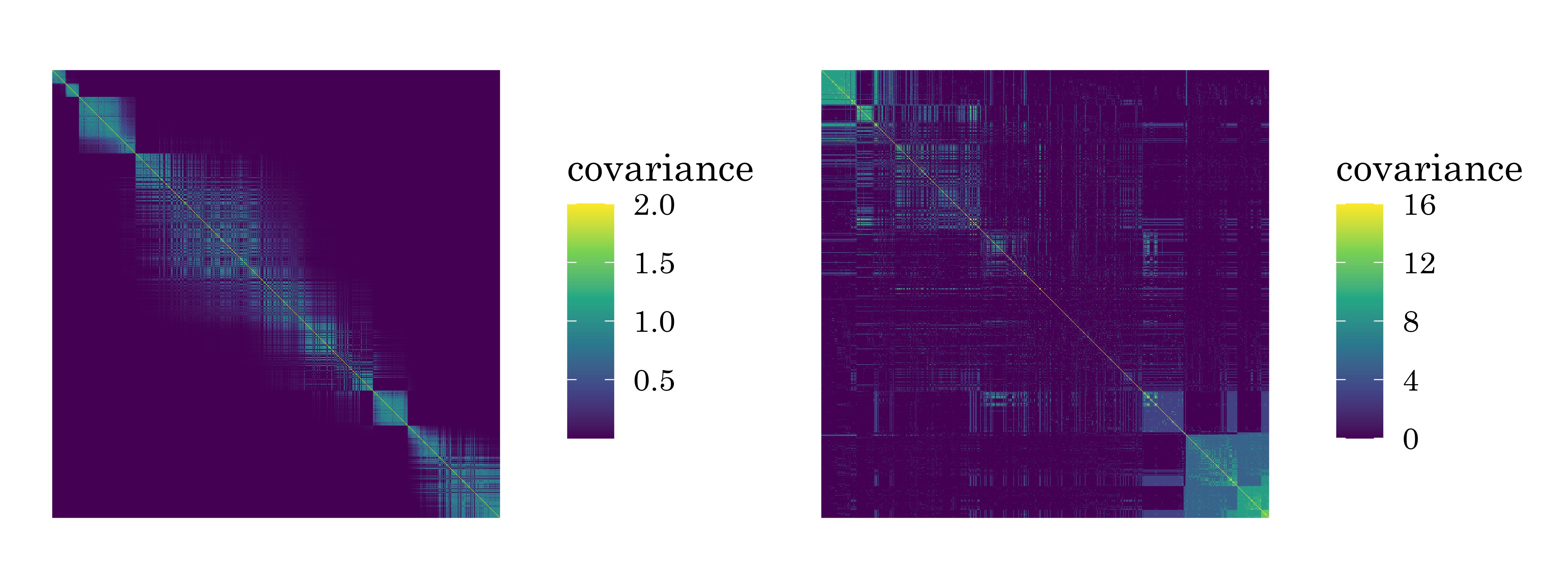}
	\caption{The marginal covariance matrix $V(\theta)$ for a random sample of $n = 1000$ UK Biobank subjects with cancer history. Left: RE feature is subject's location on the UK map (total $q = 900$ locations in sample), a simple RBF kernel $D(\psi)$ as in \eqref{eq:RBFD} is used with $\sigma^2_{b0} = \sigma^2_{b1} = 1$, locations are sorted according to first PC weight from PCA performed on the Euclidean distance matrix. Right: RE features are 5 categorical variables: diagnosis ($q = 338$ in sample), operation ($q = 304$ in sample), treatment ($q = 211$ in sample), cancer type ($q = 151$ in sample), tumor histology ($q = 104$ in sample). $\sigma^2_{bk} = k$, and data is sorted according to the first PC weight from PCA performed on $V(\theta)$ without specific order.}
	\label{fig:ukb_V_matrices}
\end{figure}

In Figure~\ref{fig:ukb_V_matrices} we can see actual covariance matrices $V(\theta)$ for a sample of $n = 1000$ UK Biobank patients with cancer history upon admission. For a detailed description of the UK Biobank data see Appendix~\ref{appx:realdetails}. The model on the left is the spatial model with $q = 900$ locations across the UK in the sample, and a simple RBF kernel $D(\psi)$ with $\sigma^2_{b0} = \sigma^2_{b1} = 1$. The model on the right is the multiple categorical model with $K = 5$ high-cardinality features: diagnosis, operation, treatment, cancer type and tumor histology. Clearly these $1000 \times 1000$ matrices are not block-diagonal, but one might conjecture that block-diagonal approximations of them would be useful in calculating their inverses and log-determinants. We find that using mini-batch gradient descent on the sorted data does just that.

Furthermore, the spatial model with RBF kernel as in \eqref{eq:RBFD} is of particular interest in this regard. As $\sigma^2_{b1}$ -- the lengthscale parameter -- gets smaller, the $D(\psi)$ kernel becomes diagonal and $V(\theta)$ the marginal covariance matrix becomes $\sigma^2_{b0}ZZ' + \sigma^2_eI_n$, where $Z$ is binary of dimension $n \times q$ is as defined in Section~\ref{intro:structures:spatial}. In other words $V(\theta)$ is block-diagonal at the limit $\sigma^2_{b1} \rightarrow 0$.

Finally, we would like to offer that this approximation of $V(\theta)$ with block-diagonal matrices that LMMNN in effect does, is reminiscent of a work by \cite{BickelLevina2008}, who proved that \textit{banding} a covariance matrix from a wide variety of classes is useful in many senses, including calculating its inverse. Specifically, for symmetric covariance matrix $\Sigma = \{m_{ij}\}$, define the $k$-banding operator $B_k(\Sigma) = [m_{ij} \mathbbm{1}(|i - j| \le k)]$. Since the $k$-banding operator is essentially capping small covariances between distant variables to zero, it is a form of regularization. \cite{BickelLevina2008} give an upper bound on $||B_k(\Sigma) - \Sigma||$ as well as on $||B_k(\Sigma)^{-1} - \Sigma^{-1}||$, where $||\cdot||$ is the matrix $L2$ norm, under some mild conditions. They comment it is ideal in the situation where $\Sigma$ is sorted in such a way that $|i - j| > k \Rightarrow m_{ij} = 0$, as in our description above. More theoretical work is needed to achieve bounds similar to \cite{BickelLevina2008} for the block-diagonal approximation in our settings of interest.

\subsection{Applying Chen et al. (2020) theorems}
\label{theory:sgd}
A recent work by \citet{Chen2020} denoted sgGP dealt with a model very similar to the spatial model presented in Section~\ref{intro:structures:spatial}, a zero-mean Gaussian process (GP) trained with a neural network's mini-batch SGD. The authors managed to bypass the question of inversion ``in parts'' and offer theoretical bounds on the variance components estimates and on the NLL gradient magnitude as the iterates progress. This is of relevance to our current discussion when we cannot show that the gradient decomposes (Section~\ref{theory:deomposition}) or that inversion in parts is valid (Section~\ref{theory:approximation}). Using this result we can show that the full gradient of the LMMNN loss in \eqref{eq:lmmnnNLL} converges to 0 using SGD for the spatial covariance and multiple categorical features scenarios, where the covariance matrix is not block-diagonal, thus concluding that LMMNN should reach at least a local minimum for these scenarios as well.

Here, the model is not dependent on any ``fixed'' features $X$, so it can be written as:
\begin{equation}
    \label{eq:ZeroMeanSpatial}
    \begin{aligned}
	    &y = Zb + \varepsilon,\\
	    &\varepsilon \sim \mathbb{N}(0, \sigma^2_eI_n),\\
	    &b \sim \mathbb{N}(0, D(\psi)),
    \end{aligned}
\end{equation}
where $D(\psi)$ is the GP standard RBF kernel from \eqref{eq:RBFD}, which the authors mark as the kernel function $k(\cdot, \cdot)$. To be consistent with \citet{Chen2020} mark $\theta = [\theta_1, \theta_2] = [\sigma^2_{b0}, \sigma^2_e]$. Note that the order we write these parameters is reversed here, and that \cite{Chen2020} knowingly leave out $\sigma^2_{b_1}$ -- the lengthscale parameter -- since as they write it is inside the exponent in \eqref{eq:RBFD}, therefore it would be difficult to take into account in their proof, but they use SGD to fit it nonetheless. 

For a full description of \citet{Chen2020}'s results see their paper. Here we bring their main assumption on the covariance matrix and their second theorem, bounding the NLL gradient magnitude:

\begin{assumption}[Exponential eigendecay, \citet{Chen2020}]\label{assum1}
	The eigenvalues of kernel function $k(\cdot; \cdot)$ w.r.t. probability measure $\mathbb{P}$ are $\{Ce^{-bj}\}_{j = 0}^\infty$, where $C \leq 1$ is regarded as a constant \footnote{When the authors write ``w.r.t. probability measure $\mathbb{P}$" they refer to a work by \citet{JMLR:v7:braun06a}, where this exponential decay of the kernel matrix eigenvalues is shown assuming $X_1, ..., X_n$ on which the kernel matrix is calculated, are a random sample from some probability space $\mathcal{X}$, with probability measure $\mathbb{P}$}
\end{assumption}
This fits the RBF kernel $D(\psi)$ or $k(\cdot, \cdot)$. The authors of sgGP comment that polynomial decay is also valid, and indeed in an extended work \citep{Chen2021} they also treat this case. This fast eigendecay quality of the covariance matrix is used to bound the trace in \eqref{eq:lmmnnSGDGradient} and eventually to bound the full gradient.

\begin{theorem}[Convergence of full gradient, \citet{Chen2020}]{\label{theo2}}
	\normalfont
	The \textit{full} gradient is bounded:\\
	For ${\frac{3}{2\gamma} \leq \alpha_1 \leq \frac{2}{\gamma}}$, ${\gamma = \frac{1}{4\theta^2_{\max}}}$, and ${0 < \varepsilon < C\frac{\log\log m}{\log m}}$ w.p. at least ${1- CK\exp\{-cm^{2\varepsilon}\}}$,
	\begin{equation}
		||\nabla NLL(\theta^K)||^2_2 \leq C\bigg[\frac{G^2}{K + 1} + m^{-\frac{1}{2} + \varepsilon}\bigg]
	\end{equation}

	Where $\alpha_1$ is the initial learning rate of SGD, $m$ is the batch size, $\theta_{min}, \theta_{max}$ are lower and upper bounds on both true variance components in $\theta$ (Assumption 2, \citet{Chen2020}), $G$ is an upper bound on the stochastic gradient (Assumption 3, \citet{Chen2020}) and $c, C > 0$ depend only on $\theta_{\min}, \theta_{\max}, b$. Most importantly, $K$ is the number of SGD iterations, so the gradient's magnitude should approach zero.
\end{theorem}

The above theorem is proven not only for a single spatial RBF kernel $k(\cdot, \cdot)$ with fast eigendecay, but also for the sum $\sum_l \sigma^2_lk_l(\cdot, \cdot)$ of $L$ general kernels each having fast eigendecay. We would naturally like to see if we can apply \citet{Chen2020} theorems to the covariance structures often encountered in LMMNN other than the RBF kernel, most importantly for structures for which the covariance matrix $V(\theta)$ is not block-diagonal. This leaves us with the multiple categorical case, which can indeed be considered as the sum of $L$ kernels as can be seen by \eqref{eq:multipleCategoricalV}. In Appendix~\ref{appx:multiple} we show how each of these kernel matrices may in fact present polynomial or even exponential eigendecay, which makes \citet{Chen2020} and \citet{Chen2021} bounds apply to this scenario as well.

\section{Related Methods} \label{related}
We will now describe some previous approaches to handling correlated data in neural networks, focusing on those which we later use in Section~\ref{results} to compare our approach to.

\subsection{Categorical features in DNNs}
\label{related:categorical}
The most prominent approach to using categorical features in any machine learning framework is one-hot encoding (OHE). If variable $v$ has $q$ distinct levels, OHE would add $q$ binary features $z_1, ..., z_q$, one for each level, with $z_{li} = 1$ if observation $i$ has level $l$ in feature $v$, and 0 otherwise. While OHE is deterministic, fast and explainable, it is hard to scale. As $q$ reaches $10000$ and more, even when using sparse data structures to store such wide datasets, many algorithms are challenged by this huge number of features. Features weights resulting from OHE also tend to carry little information and have no way of expressing complex relations between categories, for example similarity between categories.

Entity embeddings improve on OHE, by mapping each of the categorical feature's $q$ levels into a Euclidean space of a low dimensionality $d$ (Typically $d \ll q$, see e.g. \citet{guo2016entity}). After it had been one-hot encoded, the feature enters a neural network, and using the network's loss function and back propagation, a dictionary or a lookup-table $E$ of dimension $q \times d$ is learned, which is essentially a collection of $q$ vector representations or ``embeddings''. Thus if two levels are ``similar'', this would be reflected by their vector representations being close. These vectors may also later be re-used via transfer learning where the representation learned for one task can serve for other tasks, see e.g. \citet{NIPS2005_bf2fb7d1}. Entity embeddings have sparse implementations in a way which allows $q$ to scale. However the $E$ lookup table consumes much space, it may need to be learned for each new task and the resulting representations are usually hard to interpret.

A recent attempt at treating categorical features or clustering variables as RE in DNNs has been made by \citet{Xiong_2019_CVPR} and \citet{10.1007/978-3-030-20351-1_8}. The authors propose the following model named MeNets to learn fixed effects $\beta$ and random effects $b$:
\begin{equation}\label{eq:MeNets}
    y = f(X)\beta + f(X)b + \varepsilon.
\end{equation}
Here, the RE features are necessarily {\em learned}, by the same neural network that is applied to the fixed features to learn $f$ using standard squared loss and SGD. In LMMNN, in contrast, we allow for a different transformation $g$ which can also be the identity function. In order to learn $\beta$ and $b$ the authors use variational expectation maximization (V-EM) combined with SGD: An E-Step in which $\hat{\beta}, \hat{b}$ are updated while minimizing the standard squared loss with a DNN, followed by an M-Step where the variance components $\hat{\theta}$ are updated so as to maximize a NLL loss similar to \eqref{eq:lmmnnNLL}. MeNets is relevant for (and was demonstrated on) a single categorical feature with $q$ levels treated as RE with diagonal prior, which is a crucial limitation in comparison to LMMNN which generalizes to a wide variety of common covariance scenarios and combinations of these. In addition, MeNets uses two-stage training with two different loss functions, while LMMNN uses a single training stage with a single loss function. Furthermore, MeNets makes it necessary to invert all $q$ levels $n_j \times n_j$ matrices in each SGD iteration, hence for some datasets it may not even be feasible (when $q$ is ultra-high and/or when there are many small categories and one huge category which is very common in Pareto-like data). Hence, MeNets is slow per iteration (5 times longer than LMMNN on average) and in our experience also slow to converge, see runtime tables in Appendix~\ref{results:simulated:single} and results on real datasets in our previous paper \citep{lmmnn_neurips}.

\subsection{Longitudinal data in DNNs}
\label{related:longitudinal}
The go-to approach to feeding DNNs with temporal data is using recurrent neural networks (RNN), with structured cells such as LSTM \citep{lstm} suitable for remembering and forgetting previous data, in order to predict upcoming data. RNN with LSTM cells are typically used in the field of natural language processing, where sentences, paragraphs and even full documents can be thought of as long time series being fed into the DNN. However, RNN with LSTM cells may not be suitable for longitudinal data, such as growth curves and repeated measures, which tend to be very short and irregular time series exhibiting simple temporal dependence. Such data are often encountered in EMR where a patient is being followed for several hospitalization sessions, at a varied schedule (see Section~\ref{results} for simulated and real datasets which demonstrate this).

\citet{DeepGLMM} is the only work we know of which takes inspiration from LMM explicitly for handling temporal data in DNNs. These authors base their work on a very specific LMM model, in which each subject $i$ is repeatedly measured at the same set of times ${t_1,...t_T}$ for some response $y_{i,t_j}$ ($j=1,\dots,T$), which can be continuous as well as discrete, as modeled by generalized linear models (GLM). In such a model it makes sense to not only have a random intercept for each subject but also a random slope $a_{i}$. In a similar fashion to MeNets the authors propose to learn a set of features from a neural network $z_{it; j} = z(x_{it; j})$ where $j = 1, \dots, m$, the units in the last hidden layer, and have a random slope $a_{ij}$ for each unit, as well as a random intercept $a_{i0}$. In the GLM framework we model not $y$ but $\mu = E(y|x)$, via some link function $g$, for instance the logit function for binary $y$, and the authors get:
\begin{equation}\label{eq:DeepGLMM1}
    g(\mu_{it}) = \beta_0 + a_{i0} + (\beta_1 + a_{i1})z_{it; 1} + \dots + (\beta_m + a_{im})z_{it; m} = f(x_{it}, w, \beta + a_i),
\end{equation}
where $w$ are the network parameters. The authors further note that the fixed and random parts of the model can be separated such that the random part is linear with the appropriate input:
\begin{equation}\label{eq:DeepGLMM2}
    g(\mu_{it}) = f(x^{(1)}_{it}, w, \beta^{(1)}) + (\beta^{(2)} + a_i)'x^{(2)}_{it}.
\end{equation}
Here $x^{(1)}$ and $x^{(2)}$ are the fixed and random features expected to have nonlinear and linear effects respectively, and $\beta^{(1)}$ and $\beta^{(2)}$ are the linear fixed and random effects respectively. \citet{DeepGLMM} then write the likelihood for \eqref{eq:DeepGLMM2}, which is intractable, therefore they use a Bayesian approach based on variational approximation.

We note that \eqref{eq:DeepGLMM2} is similar to our criterion in \eqref{eq:lmmnnModel}, when $g$ is the identity function and $y$ is linear in $Z$, the RE features matrix. However, the variational approximation algorithm proposed in DeepGLMM, which combines numerous elements such as importance sampling, factor covariance, variable selection and choice of priors, makes it challenging to implement, let alone use as a ``plug-in'' for different DNN architectures and covariance structures as we strive to do. Finally, as with MeNets, DeepGLMM has been demonstrated in a very limited context. The number of subjects and number of time steps are both small, in the simulated as well as the real data experiments.

\subsection{Spatial data in DNNs}
\label{related:spatial}
In contrast to the few DNN adaptations of LMM for clustered and longitudinal data, when it comes to modeling spatial data there are many theoretical papers, most dealing with scaling Gaussian processes. We already expanded on sgGP \citep{Chen2020}, in this section we also explore papers which appear to be the SOTA in this field -- deep kernel learning (DKL) and stochastic variational deep kernel learning (SVDKL), originating from the same authors \citep{DKL, SVDKL}. These approaches are in wide use since they also have mature implementations in the GPyTorch library \citep{gardner2018gpytorch}. DKL applies a kernel function on the data features after they have been transformed via a DNN. Instead of fitting $k(x_i, x_j|\theta)$ where $\theta$ are the kernel parameters, we fit $k(g(x_i,w), g(x_j,w)|\theta)$, where $g$ is the DNN architecture and $w$ are the DNN weights. All parameters $w, \theta$ are jointly learned through minimizing NLL. The real ingenuity of DKL, however, comes from replacing the kernel matrix $K$ (or covariance matrix $V$ in our case) needed for NLL computation and derivation, by the KISS-GP covariance matrix \citep{KISS-GP}:
\begin{equation}\label{eq:KISS-GP}
    K \approx MK_UM,
\end{equation}
where $M$ is a sparse matrix of interpolation weights and $K_U$ is the kernel matrix $K$ evaluated at $m$ inducing points $U$. All downstream computations become substantially more efficient, to the extent that even if $g$ is the identity function (like we use it in Section~\ref{results:simulated:spatial}), DKL scales to datasets with millions of observations, without learning on mini-batches. SVDKL, in contrast, allows for mini-batch training and is even more scalable. \citet{SVDKL} use variational inference to optimize a factorized approximation of the NLL, thus bypassing the issue of decomposing the actual NLL gradient and allowing the use of SGD. The use of variational inference, combined with a fast sampling scheme, makes SVDKL suitable in classification settings as well. Both DKL and SVDKL however are based on approximations to the NLL, and are focused on scaling GPs for regression in general as opposed to handling specific correlations within the data features, such as temporal correlation in longitudinal datasets or within-cluster correlations in high-cardinality categorical features. In Section~\ref{results} we compare LMMNN's performance to SVDKL, and indeed find that for spatial data SVDKL gives comparable results to LMMNN including runtime, however when spatial data and categorical variables are both present, LMMNN can take advantage of the covariance structure induced by both random effects types (See Tables~\ref{tab:simulated:spatialcategorical:mse}, \ref{tab:real:spatial:mse}).

In addition to theoretically sound approaches, there are also numerous practical solutions for handling spatio-temporal data in DNNs, for varied applications such as crime and traffic prediction \citep{ST-ResNet, spatial-traffic} and weather forecasting \citep{spatial-weather}. For an extensive review see \citet{spatial-review}. One of those practical solutions which may work for 2-D coordinates features which are in our focus, is treating those coordinates as points on 2-D maps or images, and feeding them into a convolutional neural network (CNN). Once those images go through a standard series of convolutions and max pooling, their output could be flattened and concatenated to the output of a standard MLP for the other features, and entered into a standard loss function. In essence this strategy {\em embeds} those location features into a $d$-length Euclidean space, in a way which preserves spatial structure. As can be seen in Section~\ref{results:simulated:spatial} these embeddings are considerably more useful in prediction than embeddings which are the result of treating $q$ locations as a set of $q$ levels of a regular categorical feature, however they are still generally inferior to the approach of using a random field covariance structure in LMMNN and CNN is considerably slower (See Tables~\ref{tab:simulated:spatial:mse}, \ref{tab:real:spatial:mse} and Section~\ref{results:simulated:scale}).

\subsection{Relation to Multitask Learning}
\label{related:mtl}
The relation between LMM and multitask learning (MTL) has been addressed since the very early days of MTL. \citet{Bakker2003} applied MTL for the goal of predicting students' test results for a collection of 139 UK schools, each treated as a ``task'' -- a problem which could very well have been dealt with the random intercepts model discussed in Section~\ref{intro:structures:single}. The NLL loss in \eqref{eq:lmmnnNLL} is in fact reminiscent of many contemporary DNN losses used for performing MTL at scale. As an example consider the paper by \citet{Zhao2019}, which presents the AdaReg algorithm. The AdaReg loss aims at regularizing the weights of a DNN in an adaptive data-dependent way to ``borrow statistical strength'' from one another \citep{efron_2010}, and it is demonstrated to work on MTL applications, where typically datasets for each task are small. For simplicity consider a single hidden layer network in which $x_i \in \mathbb{R}^p$ is mapped into $d$ neurons via matrix $W$ of order $d \times p$, before a non-linear activation function is applied and a final linear layer produces prediction $\hat{y}_i \in \mathbb{R}$. Let $W$ have a matrix-variate normal prior $W \sim \mathcal{MN}(0, \Sigma_r, \Sigma_c)$, where $\Sigma_r, \Sigma_c$ are row and column covariance matrices of order $d \times d$ and $p \times p$ respectively, and define $\Omega_r := \Sigma_r^{-1}; \Omega_c = \Sigma_c^{-1}$, that is the row and column precision matrices. Then, the AdaReg loss seeks to find $W, \Omega_r, \Omega_c$ which minimize the loss:
\begin{equation}
    \label{eq:AdaReg}
    \begin{aligned}
        \text{Loss}_{AR} &= \frac{1}{2n}\Sigma_i(\hat{y}_i - y_i)^2 + \lambda||\Omega^{1/2}_rW\Omega^{1/2}_c||^2_F - \lambda(p\log|\Omega_r| + d\log|\Omega_c|) \\
        & \text{s.t. } uI_d \preceq \Omega_r \preceq vI_d, uI_p \preceq \Omega_c \preceq vI_p,
    \end{aligned}
\end{equation}
where $\lambda$ is a constant, $0 \leq u \leq v; uv = 1$, and the constraints are added to make the loss well formulated. The relation between \eqref{eq:AdaReg} and NLL loss \eqref{eq:lmmnnNLL} is not clear at first sight. Specifically, the NLL loss was reached by integrating out the RE $b$ to reach the marginal distribution of $y$ and writing its marginal negative likelihood. A different route could have been to write the joint negative likelihood of $y, b$ and minimize this loss to achieve predictions for both $y$ and $b$. Let $\hat{y} = f(X) + g(Z)b$, then:
\begin{equation}
    \label{eq:Joint1}
    \text{Loss}_{joint} = \frac{1}{2\sigma^2_e}(y - \hat{y})'(y - \hat{y}) + \frac{1}{2}b'Db + \frac{1}{2}\log|\sigma^2_eI_n| + \frac{1}{2}\log|D|.
\end{equation}
In fact, it has been shown that the loss in \eqref{eq:Joint1} will produce for $b$ the same BLUP estimate as in \eqref{eq:lmmnnBLUP} \citep{Robinson1991}. Finally, in order to reach \eqref{eq:AdaReg} one needs only assume $\sigma^2_e$ is known, and consider ``$b$'' as a RE \textit{matrix} $W$ rather then a vector, coming from a prior distribution as above with full covariance matrices on its rows and columns.
	
AdaReg could therefore be thought of as fitting a specific LMM with a proper covariance prior. LMMNN in turn represents a different approach to the one taken by AdaReg: instead of jointly optimizing for the ``RE matrix'' $W$ and ``variance components'' $\Sigma_r, \Sigma_c$, we choose to first optimize for $\Sigma_r, \Sigma_c$ and then plug those estimates to predict $W$. We leave this direction for AdaReg for future research, and note that AdaReg and MTL might prove useful for some of the scenarios examined in this paper. There are however a few critical differences between MTL and LMMNN, which make LMMNN more useful especially for large tabular datasets. First, the LMMNN approach scales to a much higher number of ``tasks'' ($q$ in our notation). Indeed most MTL DNN architectures have $q$ neurons in their final output layer, one for each task, whereas here we consider datasets where $q$ can reach tens of thousands and even 469K in one of the real datasets used in Section~\ref{results}. Second, it isn't always natural to model each of a categorical feature's levels as a ``task''. Some categorical features can only be thought of as ``just another feature'' for tabular datasets, such as the doctor in a large electronic healthcare records dataset. Finally, some of the scenarios examined in this paper would be very hard to tackle with MTL. The multiple high-cardinality categorical features is one such scenario (a combination of a doctor, a medicine, a treatment, etc.). The combination of RE features of different types is another scenario which LMMNN would handle much more naturally than MTL, as we demonstrate in Section~\ref{results:real:spatial}, with datasets having both a high-cardinality categorical feature and spatial features.

\section{Results}\label{results}
In this Section we present an extensive set of experiments demonstrating LMMNN's performance compared to other well-tested approaches. In Section~\ref{results:simulated} we apply LMMNN to a series of simulated datasets derived from the different dependence scenarios discussed in Section~\ref{intro:structures}. In Section~\ref{results:real} we apply it to real datasets from various applications, exhibiting similar dependence structures. All experiments in this paper were implemented in Python using Keras \citep{chollet2015keras} and Tensorflow \citep{tensorflow2015-whitepaper}, run on Google Colab with NVIDIA Tesla V100 GPU machines, and are publicly available in \url{https://github.com/gsimchoni/lmmnn}.

\subsection{Simulated Data}\label{results:simulated}
\subsubsection{Single categorical feature: random intercepts}
\label{results:simulated:single}
\begin{figure}
	\centering
	\includegraphics[width=1.0\linewidth]{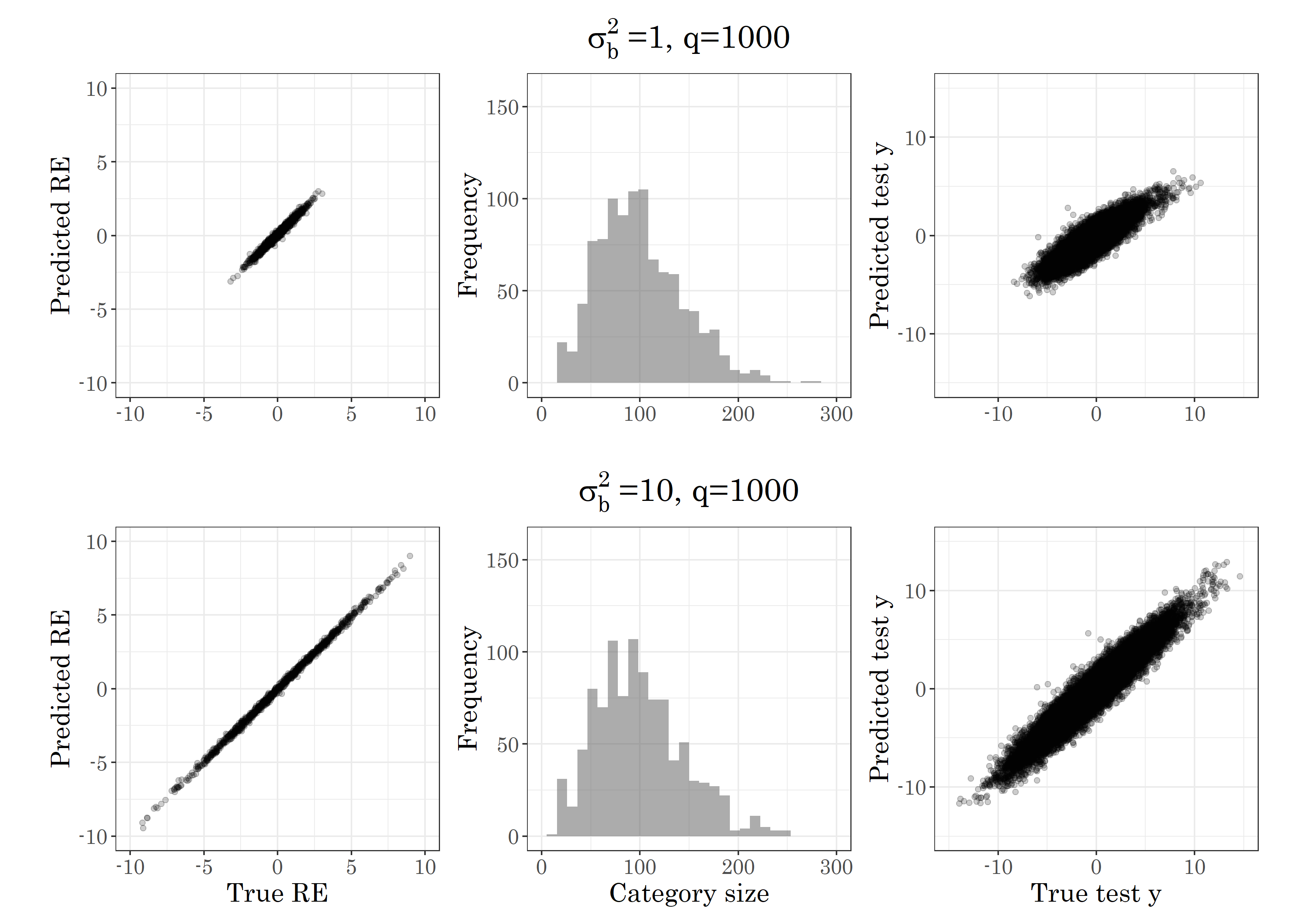}
	\caption{Simulation results with a single categorical feature when $n = 100000, g(Z) = Z, q = 1000, \sigma^2_b = 1$ (top) and $\sigma^2_b = 10$ (bottom)}
	\label{fig:sim_viz}
\end{figure}

We start by simulating the model in \eqref{eq:lmmnnModel} with a single categorical feature with $q$ levels and variance $\sigma^2_b$, where $q$ is varied in $\{100, 1000, 10000\}$ and $\sigma^2_b$ is varied in $\{0.1, 1, 10\}$. $n = 100000$ and $\sigma^2_e = 1$ always. The $q$ levels are not evenly distributed among the $n$ observations, rather we use a multinomial distribution sampling where the $q$ probabilities are obtained by sampling $q$ $\text{Poisson}(30)$ random variables, and standardizing them to sum to 1 (see category level sizes distribution in Figure~\ref{fig:sim_viz}). There are 10 fixed features in $X$ sampled from a $\mathbb{U}(-1,1)$ distribution, non-linearly related to $y$:
\begin{equation}\label{eq:f}
    y = (X_1 + \dots + X_{10}) \cdot \cos(X_1 + \dots + X_{10}) + 2 \cdot X_1 \cdot X_2 + g(Z)b + \varepsilon,
\end{equation}
where $Z$ is of dimension $n \times q$ as described in Section~\ref{intro:structures:single}. $g(Z)$ is either the identity function or $g(Z) = ZW$, where $W$ is a linear transformation $W_{q \times d}$ with values sampled from a $\mathbb{U}(-1, 1)$ distribution, and $d = 0.1 \cdot q$, or $g(Z)$ is a non-linear function $(Z_i \cdot W') * \cos(Z_i \cdot W')$, where $Z_i$ is the $i$th row of $Z$, $*$ is elementwise multiplication and $g$ is applied rowwise. We perform 5 iterations for each $(q, \sigma^2_b, g)$ combination (27 combinations in total), in which we sample the data, randomly split it into training (80\%) and testing (20\%), train our models to predict $\hat{y}_{te}$ and compare the bottom-line MSEs in predicting $y_{te}$. We compare its MSE to those of R's lme4 package results (i.e. standard LMM) \citep{lme4}, MeNets, OHE, entity embeddings and ignoring the categorical feature in $Z$ altogether. We use the same DNN architecture for all neural networks, that is 4 hidden layers with {100, 50, 25, 12} neurons, a Dropout of 25\% in each, a ReLU activation and a final output layer with a single neuron. When $g(Z)$ is not the identity function we use an embedding layer on $Z$ to learn $g$ (in case $g(Z) = ZW$ it is the ``correct'' transformation to use and in case $g(Z)$ is not linear it is ``incorrect'' and thus more challenging). The loss we use is mean squared error (MSE) loss for OHE, embeddings and ignoring the RE, and NLL for LMMNN and MeNets (as mentioned above, MeNets uses squared loss for estimating fixed effects and NLL for variance components only). In all experiments in this paper we use a batch size of 100 and an early stopping rule where training is stopped if no improvement in 10\% validation loss is seen within 10 epochs, up to a maximum of 500 epochs. For prediction in LMMNN, in case $g(Z) = Z$ the formula in \eqref{eq:classicRandomInterceptsPrediction} is used adjusted for LMMNN output $\hat{f}(X_{tr})$, and when $g(Z) = ZW$ we sample 10000 observations when calculating \eqref{eq:lmmnnBLUP}, in order to avoid inverting $V(\theta)$ which is of dimension $80000 \times 80000$. We initialize both $\hat{\sigma}^2_e, \hat{\sigma}^2_b$ to be 1.0 where appropriate: R's lme4 and LMMNN, and compare the resulting final estimates for these two methods.

\begin{table}
  \caption{Simulated model with a single categorical feature, mean test MSEs and standard errors in parentheses. Bold results are non-inferior to the best result in a paired t-test. Hence, LMMNN is significantly better than all competitors in all scenarios.}
  \label{tab:simulated:single: mse}
  \centering
  \begin{tabular}{ll|llllll}
\toprule
\multicolumn{8}{c}{\textbf{g(Z) = Z}} \\
\midrule
$\sigma^2_b$ & $q$ & Ignore & OHE & Embeddings & lme4 & MeNets & LMMNN \\
	\midrule
0.1 & $10^2$   & 1.24 (.01) & 1.18 (.02) & 1.16 (.01) & 2.93 (.03) & 1.16 (.02) & \textbf{1.10 (.01)}  \\
    & $10^3$  & 1.22 (.02) & 1.28 (.00) & 1.21 (.01) & 2.93 (.02) & 1.33 (.06) & \textbf{1.09 (.01)} \\
    & $10^4$ & 1.22 (.01) & 1.57 (.02) & 1.58 (.01) & 2.96 (.02) & 1.65 (.26) & \textbf{1.18 (.01)} \\
\midrule
1 & $10^2$   & 2.09 (.10) & 1.23 (.03) & 1.18 (.01) & 2.93 (.02) & 1.18 (.02) & \textbf{1.10 (.00)} \\
    & $10^3$  & 2.15 (.03) & 1.36 (.02) & 1.28 (.02) & 2.94 (.02) & 1.53 (.17) & \textbf{1.10 (.01)} \\
    & $10^4$ & 2.15 (.03) & 1.70 (.02) & 1.67 (.01) & 3.22 (.02) & 1.60 (.06) & \textbf{1.24 (.01)} \\
\midrule
10 & $10^2$ & 10.8 (.45) & 1.55 (.07) & 1.55 (.06) & 2.93 (.02) & 1.85 (.22) & \textbf{1.11 (.01)} \\
    & $10^3$  & 11.1 (.15) & 1.60 (.02) & 1.65 (.07) & 2.93 (.03) & 2.01 (.17) & \textbf{1.09 (.01)} \\
    & $10^4$ & 11.2 (.06) & 2.37 (.07) & 2.12 (.04) & 3.32 (.02) & 2.80 (.36) & \textbf{1.29 (.01)} \\
	\midrule
	\multicolumn{8}{c}{\textbf{g(Z) = ZW}} \\
	\midrule
    0.1 & $10^2$   & 1.48 (.08) & 1.19 (.01) & \textbf{1.17 (.03)} & 2.91 (.02) & 1.25 (.08) & \textbf{1.15 (.02)} \\
        & $10^3$  & 4.45 (.16) & 1.40 (.02) & 1.39 (.03) & 2.95 (.02) & 1.44 (.06) & \textbf{1.25 (.01)} \\
        & $10^4$ & 36.1 (.7) & 3.95 (.25) & 3.34 (.07) & 3.42 (.04) & 7.35 (1.95) & \textbf{2.40 (.03)} \\
    \midrule
    1 & $10^2$  & 4.48 (.71) & 1.39 (.06) & 1.37 (.04) & 2.88 (.02) & 1.40 (.11) & \textbf{1.12 (.01)} \\
        & $10^3$  & 34.6 (2.2) & 2.20 (.21) & 2.51 (.24) & 2.96 (.05) & 7.00 (1.9) & \textbf{1.28 (.01)} \\
        & $10^4$ & 332.6 (9.7) & 13.8 (1.9) & 15.21 (2.7) & \textbf{4.29 (.10)} & 143.3 (32.5) & \textbf{4.49 (.10)} \\
    \midrule
    10 & $10^2$ & 35.9 (3.3) & 2.36 (.10) & 2.87 (.27) & 2.90 (.02) & 12.03 (3.03) & \textbf{1.14 (.02)} \\
        & $10^3$ & 381.9 (16.9) & 9.3 (1.7) & 15.1 (2.6) & 2.96 (.03) & 163.7 (17.9) & \textbf{1.31 (.03)} \\
        & $10^4$ & 3365.9 (42.3) & 81.3 (16.3) & 153.5 (14.4) & \textbf{13.8 (1.2)} & 2880.6 (463.5) & \textbf{13.9 (1.6)} \\
    \midrule
    \multicolumn{8}{c}{\textbf{g(Z) = ZW * cos(ZW)}} \\
	\midrule
    0.1 & $10^2$ & 1.27 (.01) & 1.19 (.02) & \textbf{1.16 (.01)} & 2.93 (.01) & \textbf{1.14 (.00)} & \textbf{1.14 (.02)} \\
        & $10^3$ & 2.93 (.21) & 1.39 (.02) & 1.36 (.03) & 2.92 (.02) & 1.69 (.09) & \textbf{1.28 (.02)} \\
        & $10^4$ & 19.1 (.46) & 2.87 (.11) & 2.54 (.06) & 3.36 (.03) & 3.71 (.81) & \textbf{2.25 (.02)} \\
    \midrule
    1 & $10^2$ & 2.91 (.26) & 1.23 (.01) & 1.25 (.02) & 2.91 (.02) & 1.48 (.11) & \textbf{1.12 (.01)} \\
        & $10^3$ & 21.0 (.94) & 1.96 (.10) & 2.22 (.17) & 2.95 (.02) & 2.82 (.40) & \textbf{1.26 (.02)} \\
        & $10^4$ & 178.9 (4.7) & 7.37 (.80) & 8.31 (1.2) & \textbf{3.91 (.11)} & 78.8 (19.5) & \textbf{3.51 (.16)} \\
    \midrule
    10 & $10^2$ & 23.8 (3.5) & 1.86 (.22) & 1.96 (.10) & 2.90 (.02) & 2.20 (.30) & \textbf{1.13 (.02)} \\
        & $10^3$ & 161.1 (10.2) & 6.05 (.72) & 10.5 (1.2) & 2.92 (.01) & 79.9 (10.7) & \textbf{1.32 (.02)} \\
        & $10^4$ & 1797.4 (48.0) & 36.3 (4.0) & 90.7 (9.7) & \textbf{13.6 (2.3)} & 725.1 (75.3) & \textbf{14.1 (.89)} \\
\bottomrule
\end{tabular}

\end{table}

Table~\ref{tab:simulated:single: mse} summarizes the test MSE results and Table~\ref{tab:simulated:single:sigmas} in Appendix~\ref{appx:tables:variance} summarizes the estimated variance components results. As can be seen LMMNN reaches the smallest test MSE on average and with a considerable gap from the other methods, when standard errors are taken into account. This is particularly true when RE variance $\sigma^2_b$ and cardinality $q$ are high and when $g(Z)$ isn't the identity function. As for the estimated variance components $\hat{\sigma}^2_e, \hat{\sigma}^2_b$, LMMNN reaches a good estimation for both when $g(Z) = Z$, while R's lme4 reaches a poor estimation for $\sigma^2_e$ without adding appropriate non-linear and interaction terms, resulting in worse prediction performance. When $g(Z)$ is not the identity function LMMNN struggles to reach good estimates for $\sigma^2_e, \sigma^2_b$, but they are still considerably better than R's lme4. Here we note that when $g(Z) = ZW$ we found that additional training of the network until the variance components estimates converge may sometimes lead to improved estimates. Finally Table~\ref{tab:simulated:single:times} in Appendix~\ref{appx:tables:times} summarizes mean runtime and number of epochs, and in Figure~\ref{fig:sim_viz} we show predicted RE and $\hat{y}_{te}$ versus true RE and $y_{te}$ in two of the scenarios.

\subsubsection{Multiple categorical features}
\label{results:simulated:multiple}

Table~\ref{tab:simulated:multiple:mse} and Table~\ref{tab:simulated:multiple:sigmas} in Appendix~\ref{appx:tables:variance} summarize a simulation where $K = 3$ high-cardinality categorical features are used. We have as above $n = 100000$, $q_1 = 1000, q_2 = 2000, q_3 = 3000$, so $Z$ is of dimension $100000 \times 6000$ in a model identical to \eqref{eq:f}. We keep $\sigma^2_e = 1$, vary $[\sigma^2_{b1}, \sigma^2_{b2}, \sigma^2_{b3}]$ in $(0.3, 3.0)$ and $g(Z)$ is one of the three options as in the single categorical simulation, a total of 24 combinations. We use the same MLP architecture and training details as in previous simulation, where here MeNets is no longer applicable.

As can be seen LMMNN is the clear winner in terms of mean test MSE and in terms of variance components estimates. Its performance is especially impressive as the RE variance components increase, failing well-tested solutions like OHE and entity embeddings. As with a single categorical feature, when $g(Z)$ is not the identity function LMMNN's variance components estimates are no longer accurate but they are much closer to the true values than those of R's lme4. Mean running times and number of training epochs are summarized in Table~\ref{tab:simulated:multiple:times} in Appendix~\ref{appx:tables:times}.

\begin{table}
  \caption{Simulated model with 3 categorical features, with $q_1 = 1000, q_2 = 2000, q_3 = 3000$. Mean test MSEs and standard errors in parentheses. Bold results are non-inferior to the best result in a paired t-test.}
  \label{tab:simulated:multiple:mse}
  \centering
  \begin{tabular}{lll|lllll}
\toprule
\multicolumn{8}{c}{\textbf{g(Z) = Z}} \\
\midrule
$\sigma^2_{b_1}$ & $\sigma^2_{b_2}$ & $\sigma^2_{b_3}$ & Ignore & OHE & Embed. & lme4 & LMMNN \\
	\midrule
    0.3 & 0.3 & 0.3 & 2.06 (.01) & 1.62 (.01) & 1.48 (.01) & 3.04 (.01) & \textbf{1.16 (.01)}  \\
    &  & 3.0 & 4.85 (.04) & 1.87 (.01) & 1.63 (.02) & 3.05 (.01) & \textbf{1.17 (.01)}  \\
    & 3.0 & 0.3 & 4.72 (.05) & 1.83 (.02) & 1.60 (.02) & 3.05 (.01) & \textbf{1.15 (.01)}  \\
    &  & 3.0 & 7.61 (.11) & 2.05 (.02) & 1.79 (.04) & 3.12 (.01) & \textbf{1.18 (.02)}  \\
	\midrule
    3.0 & 0.3 & 0.3 & 4.89 (.07) & 1.79 (.04) & 1.61 (.04) & 3.02 (.01) & \textbf{1.16 (.01)}  \\
    &  & 3.0 & 7.62 (.13) & 2.00 (.04) & 1.81 (.03) & 3.05 (.03) & \textbf{1.16 (.02)}  \\
    & 3.0 & 0.3 & 7.36 (.14) & 1.93 (.03) & 1.70 (.02) & 3.05 (.02) & \textbf{1.15 (.02)}  \\
    &  & 3.0 & 10.2 (.14) & 2.17 (.03) & 1.92 (.05) & 3.07 (.01) & \textbf{1.17 (.01)}  \\
	\midrule
	\multicolumn{8}{c}{\textbf{g(Z) = ZW}} \\
	\midrule
    0.3 & 0.3 & 0.3 & 62.0 (.88) & 4.36 (.22) & 3.65 (.24) & 3.12 (.03) & \textbf{1.90 (.04)}  \\
    &  & 3.0 & 333.9 (9.23) & 12.8 (1.69) & 15.5 (1.12) & 3.17 (.02) & \textbf{1.96 (.03)}  \\
    & 3.0 & 0.3 & 242.5 (6.63) & 11.2 (.71) & 12.8 (1.71) & 3.19 (.02) & \textbf{1.92 (.02)}  \\
    &  & 3.0 & 509.4 (18.1) & 13.2 (1.8) & 25.3 (1.61) & 3.18 (.02) & \textbf{2.41 (.02)}  \\
	\midrule
    3.0 & 0.3 & 0.3 & 151.3 (7.38) & 7.67 (.73) & 8.14 (.92) & 3.13 (.02) & \textbf{1.93 (.05)}  \\
    &  & 3.0 & 429.6 (10.4) & 17.1 (1.98) & 22.8 (2.68) & 3.18 (.02) & \textbf{2.31 (.05)}  \\
    & 3.0 & 0.3 & 358.7 (18.3) & 16.82 (1.23) & 21.5 (2.04) & 3.19 (.02) & \textbf{2.05 (.04)}  \\
    &  & 3.0 & 611.3 (14.1) & 23.7 (2.6) & 31.4 (2.02) & 3.25 (.03) & \textbf{2.50 (.06)}  \\
    \midrule
	\multicolumn{8}{c}{\textbf{g(Z) = ZW * cos(ZW)}} \\
	\midrule
    0.3 & 0.3 & 0.3 & 32.1 (.72) & 3.34 (.04) & 2.71 (.10) & 3.12 (.02) & \textbf{1.77 (.03)} \\
    &  & 3.0 & 187.2 (10.8) & 6.60 (.69) & 11.2 (.78) & 3.13 (.01) & \textbf{1.89 (.02)} \\
    & 3.0 & 0.3 & 123.3 (3.5) & 7.00 (.52) & 7.07 (.68) & 3.10 (.01) & \textbf{1.88 (.01)} \\
    &  & 3.0 & 280.4 (11.2) & 12.3 (.53) & 14.1 (1.7) & 3.16 (.03) & \textbf{2.15 (.02)} \\
	\midrule
    3.0 & 0.3 & 0.3 & 90.2 (4.4) & 6.37 (.35) & 4.53 (.48) & 3.12 (.02) & \textbf{1.85 (.02)} \\
    &  & 3.0 & 217.1 (4.4) & 7.41 (.58) & 11.5 (1.4) & 3.19 (.02) & \textbf{2.05 (.01)} \\
    & 3.0 & 0.3 & 172.7 (11.8) & 8.95 (1.8) & 8.86 (1.4) & 3.13 (.02) & \textbf{1.93 (.02)} \\
    &  & 3.0 & 316.8 (5.2) & 11.0 (1.5) & 15.1 (2.0) & 3.21 (.02) & \textbf{2.30 (.04)} \\
\bottomrule
\end{tabular}
\end{table}

\subsubsection{Longitudinal data and repeated measures}
\label{results:simulated:longitudinal}
For the longitudinal model we take a model very similar to \eqref{eq:longitudinalScalar}, except now $y$ is related to $X$ via the non-linear function $f$ shown in $\eqref{eq:f}$, and $K = 3$ so time $t$ has intercept, linear and quadratic terms:
\begin{equation}\label{eq:longitudinalNonLinear}
    y_{ij} = f(x_{ij}) + b_{0,j} + b_{1,j} \cdot t_{ij} + b_{2,j} \cdot t^2_{ij} + \varepsilon_{ij}
\end{equation}
We sample a variable number of $n_j$ measurements from each of $q = 10000$ subjects, the total being $n = 100000$ as before. $t$ is taken from a sequence of $\max{n_j}$ equally sized steps between 0 and 1. If $\max{n_j} = 6$ for example, the possible sequence is $[0, 0.2, 0.4, 0.6, 0.8, 1]$, and a subject with $n_j = 2$ will have measurements in times $0$ and $0.2$, while a subject with $n_j = 6$ will have measurements in times $[0, 0.2, 0.4, 0.6, 0.8, 1]$. To challenge LMMNN we also add two of the possible three correlations: between the intercept and slope terms $\rho_{01}$, between the intercept and quadratic terms $\rho_{02}$, but not between the slope and quadratic terms. This gives a total of 6 variance components to estimate: $\theta = [\sigma^2_{e}, \sigma^2_{b_0}, \sigma^2_{b_1}, \sigma^2_{b_2}, \rho_{01}, \rho_{02}]$. We fix $\sigma^2_e$ at 1 as before, we fix $\rho_{01} = \rho_{02}$ at 0.3 and vary $[\sigma^2_{b_0}, \sigma^2_{b_1}, \sigma^2_{b_2}]$ in $(0.3, 3.0)$. To make the simulation more realistic we not only include a ``Random'' mode where the data is split randomly to 80\% training and 20\% testing sets, but also a ``Future'' mode where the testing set are those 20\% observations which occur latest in time $t$ across all $n$ observations, meaning that the model is only trained on past observations. This means a total of 16 experiments. As before, we compare LMMNN's results to ignoring the temporal dependence, one-hot encoding the $q$ patients, embedding them and using standard LMM in R's lme4 package. All training details and networks baseline architectures are identical to those described in Section~\ref{results:simulated:single}. Here we also compare LMMNN's results to performing LSTM on these short time series, where the LSTM architecture was chosen via performing grid search on optional parameters and choosing a single LSTM layer with 5 neurons.

Table~\ref{tab:simulated:longitudinal:mse} and Table~\ref{tab:simulated:longitudinal:sigmas} in Appendix~\ref{appx:tables:variance} summarize the mean test MSE and estimated variance components results. As can be seen LMMNN's performance is superior to all other methods, and especially that of standard LMM with R's lme4. The Future mode is generally more challenging to all methods, but LMMNN still performs best by a considerable margin. Looking at the variance components results, the ``higher'' the term the more challenging it is for LMMNN to reach a good estimate (namely, estimating $\sigma^2_{b2}$ and $\rho_{02}$ versus estimating $\sigma^2_{b0}$ and $\rho_{01}$). Its estimates are still much better than those of R's lme4. Mean running times and number of training epochs are summarized in Table~\ref{tab:simulated:longitudinal:times} in Appendix~\ref{appx:tables:times}.

\begin{table}
  \caption{Simulated model with longitudinal data for $q = 10000$ subjects. Mean test MSEs and standard errors in parentheses. Bold results are non-inferior to the best result in a paired t-test.}
  \label{tab:simulated:longitudinal:mse}
  \centering
  \begin{tabular}{lll|llllll}
\toprule
\multicolumn{9}{c}{\textbf{Mode: Random}} \\
\midrule
$\sigma^2_{b0}$ & $\sigma^2_{b1}$ & $\sigma^2_{b2}$ & Ignore & OHE & Embed. & lme4 & LSTM & LMMNN \\
	\midrule
    0.3 & 0.3 & 0.3 & 1.47 (.01) & 1.61 (.01) & 1.63 (.01) & 3.18 (.03) & 1.40 (.01) & \textbf{1.23 (.01)} \\
    &  & 3.0 & 1.51 (.01) & 1.63 (.01) & 1.64 (.01) & 3.15 (.04) & 1.44 (.01) & \textbf{1.23 (.02)} \\
    & 3.0 & 0.3 & 1.67 (.03) & 1.65 (.01) & 1.66 (.01) & 3.18 (.04) & 1.58 (.03) & \textbf{1.25 (.01)} \\
    &  & 3.0 & 1.73 (.03) & 1.68 (.01) & 1.66 (.01) & 3.15 (.02) & 1.63 (.02) & \textbf{1.26 (.03)} \\
	\midrule
    3.0 & 0.3 & 0.3 & 4.29 (.04) & 1.87 (.02) & 1.80 (.02) & 4.55 (.24) & 4.23 (.03) & \textbf{1.29 (.02)} \\
    &  & 3.0 & 4.44 (.04) & 1.95 (.02) & 1.88 (.03) & 5.00 (.54) & 4.35 (.04) & \textbf{1.26 (.01)} \\
    & 3.0 & 0.3 & 4.58 (.03) & 1.96 (.04) & 1.85 (.01) & 5.06 (.26) & 4.50 (.05) & \textbf{1.27 (.01)} \\
    &  & 3.0 & 4.72 (.04) & 1.96 (.05) & 1.88 (.03) & 5.10 (.36) & 4.55 (.02) & \textbf{1.29 (.01)} \\
	\midrule
	\multicolumn{9}{c}{\textbf{Mode: Future}} \\
	\midrule
    0.3 & 0.3 & 0.3 & 1.65 (.02) & 1.74 (.02) & 1.72 (.02) & 3.38 (.06) & 1.49 (.01) & \textbf{1.27 (.01)}  \\
    &  & 3.0 & 1.75 (.03) & 1.84 (.02) & 1.83 (.02) & 3.44 (.05) & 1.65 (.02) & \textbf{1.36 (.02)} \\
    & 3.0 & 0.3 & 2.17 (.08) & 2.01 (.03) & 2.01 (.06) & 3.60 (.05) & 2.12 (.04) & \textbf{1.43 (.03)} \\
    &  & 3.0 & 2.29 (.05) & 2.04 (.03) & 2.11 (.03) & 3.69 (.07) & 2.17 (.03) & \textbf{1.47 (.02)} \\
	\midrule
    3.0 & 0.3 & 0.3 & 4.58 (.04) & 1.94 (.03) & 1.93 (.04) & 4.64 (.49) & 4.43 (.06) & \textbf{1.29 (.02)} \\
    &  & 3.0 & 4.90 (.05) & 2.17 (.05) & 2.06 (.07) & 4.73 (.44) & 4.71 (.05) & \textbf{1.35 (.01)} \\
    & 3.0 & 0.3 & 5.51 (.07) & 2.20 (.04) & 2.14 (.03) & 5.17 (.45) & 5.29 (.11) & \textbf{1.43 (.02)} \\
    &  & 3.0 & 5.56 (.08) & 2.23 (.06) & 2.25 (.04) & 4.82 (.54) & 5.47 (.10) & \textbf{1.47 (.02)} \\
\bottomrule
\end{tabular}
\end{table}

\subsubsection{Spatial data}
\label{results:simulated:spatial}
For spatial data we use the standard model:
\begin{equation}\label{eq:spatialNonLinear}
    y_{ij} = f(x_{ij}) + b_{j} + \varepsilon_{ij},
\end{equation}
where $b_j$ is a 2-D location random effect with zero mean and covariance matrix $D(\psi)$ as described in Section~\ref{intro:structures:spatial} with the RBF kernel in \eqref{eq:RBFD}, and $f$ is non-linear as shown in \eqref{eq:f}. We sample $q$ 2-D locations from the $\mathbb{U}(-10, 10) \times \mathbb{U}(-10, 10)$ grid, where $q$ is varied in $\{100, 1000, 10000\}$. We sample a variable number of measurements from each of the $q$ locations, the total being $n = 100000$ as before. We fix $\sigma^2_e$ at 1 and vary the RBF kernel variance components $[\sigma^2_{b0}, \sigma^2_{b1}]$ in $(0.1, 1, 10)$, for a total of 27 combinations. Here we exclude results for ignoring the spatial correlation for brevity and since they are clearly the worst. We further tried to perform standard kriging using R's gstat package yet it failed to scale to this magnitude of problem. For LMMNN we used two approaches: LMMNN-R was trained assuming a standard RBF kernel, and LMMNN-E was trained without such assumption, demonstrating an additional use of a non-linear $g(Z)$ as described in Section~\ref{proposed}. LMMNN-E passes the 2-D locations $s_i, s_j$ through a standard MLP with 6 layers of $(1000, 500, 200, 100, 500, 1000)$ neurons, before entering a standard NLL layer as if it were a single RE feature of dimension 1000 with a single variance parameter $\sigma^2_{b_0}$. As for SOTA methods, we compared LMMNN to using DKL and SVDKL with 500 inducing points as described in Section~\ref{related:spatial} and run in GPyTorch. A standard baseline MLP for the fixed features is used as the mean of a multivariate normal distribution and a standard RBF kernel for the 2-D locations as its covariance, fitted via NLL minimization. We report here only the SVDKL results for brevity and since they didn't differ that much from those of DKL. We also compared our approach to using a CNN on locations treated as images, as described in Section~\ref{related:spatial}. For CNN we used a standard architecture of four 2D convolutions layers with $[32, 64, 32, 16]$ filters and a kernel of size 2, separated by max pooling, concatenated with a standard baseline MLP for the fixed features. All other training details such as batch size, hardware and baseline MLP architectures are identical to those described in Section~\ref{results:simulated:single}.

\begin{figure}
	\centering
	\includegraphics[width=1.0\linewidth]{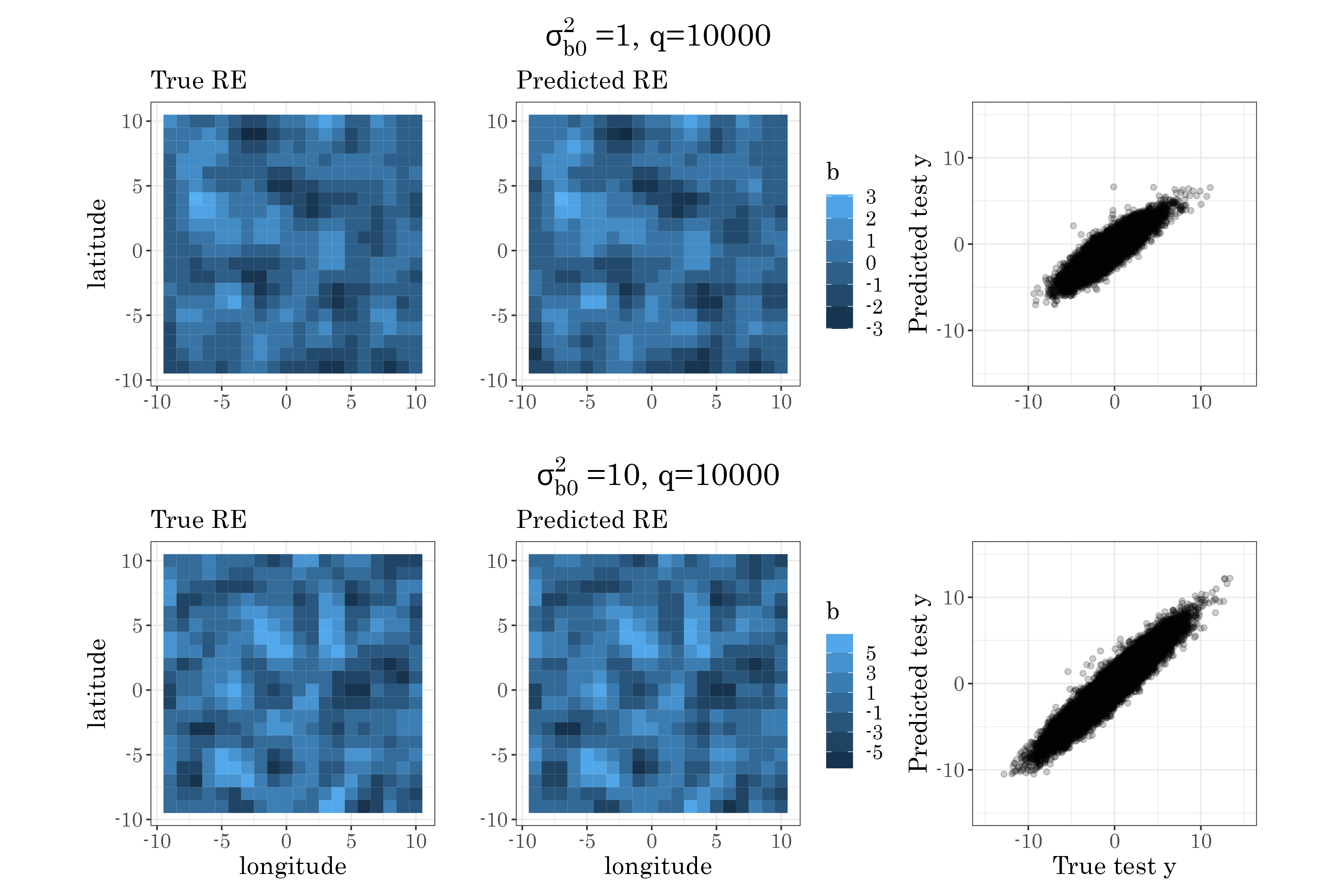}
	\caption{Spatial data simulation results with $q = 10000$ locations, $n = 100000, \sigma^2_{b1} = 1$, and $\sigma^2_{b0} = 1$ (top) and $\sigma^2_{b0} = 10$ (bottom)}
	\label{fig:sim_spatial_viz}
\end{figure}

Table~\ref{tab:simulated:spatial:mse} summarizes the mean test MSE results. LMMNN's main competition is SVDKL (and DKL) performing similar in most experiments, but it performs better in the extreme scenarios of a very low lengthscale $\sigma^2_{b1} = 0.1$, a medium to high scaling variance $\sigma^2_{b0} = 1$ or $10$ and a large $q$. LMMNN is also faster than SVDKL by a typical factor of 2-5 as can be seen in Table~\ref{tab:simulated:spatial:times} in Appendix~\ref{appx:tables:times}, and in those extreme scenarios even by a factor of 10, where DKL reaches the limit of 500 epochs. It is also interesting to note that LMMNN without assuming a known RBF kernel (LMMNN-E) but passing the locations through a deep embedding network, also performs quite well in most experiments. In Table~\ref{tab:simulated:spatial:sigmas} in Appendix~\ref{appx:tables:variance} we present LMMNN's variance components estimates, where it finds estimating the lengthscale $\sigma^2_{b1}$ considerably more challenging. We also show in Figure~\ref{fig:sim_spatial_viz} predicted RE and $\hat{y}_{te}$ versus true RE and $y_{te}$ for two spatial scenarios.

\begin{table}
  \caption{Simulated model with spatial data with a RBF kernel. Mean test MSEs and standard errors in parentheses. Bold results are non-inferior to the best result in a paired t-test.}
  \label{tab:simulated:spatial:mse}
  \centering
  \begin{tabular}{lll|llllll}
\toprule
$\sigma^2_{b0}$ & $\sigma^2_{b1}$ & $q$ & OHE & Embed. & CNN & SVDKL & LMMNN-E & LMMNN-R \\
	\midrule
    0.1 & 0.1 & $10^2$ & 1.22 (.01) & 1.25 (.00) & 1.19 (.02) & \textbf{1.09 (.01)} & 1.22 (.02) & \textbf{1.10 (.02)} \\
    & & $10^3$ & 1.30 (.01) & 1.29 (.02) & 1.18 (.02) & \textbf{1.14 (.02)} & 1.20 (.02) & \textbf{1.13 (.01)} \\
    & & $10^4$ & 1.54 (.01) & 1.60 (.01) & 1.26 (.02) & \textbf{1.17 (.01)} & 1.23 (.01) & \textbf{1.17 (.01)}  \\
    \cline{2-9}
    & 1.0 & $10^2$ & 1.21 (.02) & 1.20 (.02) & 1.18 (.02) & \textbf{1.12 (.02)} & \textbf{1.13 (.01)} & \textbf{1.11 (.01)} \\
    & & $10^3$ & 1.29 (.01) & 1.27 (.01) & 1.17 (.01) & 1.14 (.01) & 1.19 (.03) & \textbf{1.10 (.01)}  \\
    & & $10^4$ & 1.55 (.01) & 1.60 (.01) & 1.22 (.02) & \textbf{1.10 (.01)} & 1.23 (.01) & \textbf{1.10 (.01)} \\
    \cline{2-9}
    & 10.0 & $10^2$ & 1.23 (.01) & 1.22 (.01) & 1.18 (.02) & \textbf{1.10 (.02)} & \textbf{1.12 (.01)} & \textbf{1.10 (.02)} \\
    & & $10^3$ & 1.28 (.01) & 1.28 (.02) & 1.16 (.01) & \textbf{1.12 (.01)} & \textbf{1.12 (.02)} & \textbf{1.12 (.01)} \\
    & & $10^4$ & 1.55 (.01) & 1.62 (.01) & 1.19 (.01) & \textbf{1.10 (.01)} & \textbf{1.11 (.01)} & \textbf{1.12 (.02)} \\
    \midrule
    1.0 & 0.1 & $10^2$ & 1.26 (.02) & 1.27 (.02) & 1.25 (.04) & \textbf{1.13 (.02)} & \textbf{1.12 (.01)} & \textbf{1.14 (.02)} \\
    & & $10^3$ & 1.35 (.01) & 1.34 (.01) & \textbf{1.28 (.02)} & \textbf{1.26 (.03)} & \textbf{1.26 (.02)} & \textbf{1.29 (.07)} \\
    & & $10^4$ & 1.70 (.01) & 1.73 (.01) & 1.42 (.02) & 1.45 (.02) & 1.66 (.02) & \textbf{1.30 (.02)} \\
    \cline{2-9}
    & 1.0 & $10^2$ & 1.28 (.02) & 1.27 (.02) & 1.21 (.02) & \textbf{1.10 (.01)} & \textbf{1.11 (.01)} & \textbf{1.10 (.01)} \\
    & & $10^3$ & 1.33 (.01) & 1.34 (.02) & 1.27 (.02) & \textbf{1.12 (.01)} & 1.18 (.02) & \textbf{1.13 (.02)} \\
    & & $10^4$ & 1.68 (.01) & 1.73 (.01) & 1.31 (.01) & \textbf{1.11 (.01)} & 1.18 (.01) & \textbf{1.16 (.01)} \\
    \cline{2-9}
    & 10.0 & $10^2$ & 1.28 (.01) & 1.29 (.03) & 1.20 (.02) & \textbf{1.11 (.01)} & \textbf{1.13 (.01)} & \textbf{1.11 (.02)} \\
    & & $10^3$ & 1.34 (.01) & 1.30 (.02) & 1.22 (.02) & \textbf{1.09 (.03)} & \textbf{1.10 (.01)} & \textbf{1.10 (.01)} \\
    & & $10^4$ & 1.62 (.01) & 1.68 (.02) & 1.24 (.03) & \textbf{1.11 (.01)} & \textbf{1.11 (.01)} & \textbf{1.09 (.01)} \\
    \midrule
    10.0 & 0.1 & $10^2$ & 1.66 (.05) & 1.72 (.02) & 1.32 (.03) & \textbf{1.11 (.01)} & 1.17 (.02) & \textbf{1.09 (.00)} \\
    & & $10^3$ & 1.67 (.05) & 1.86 (.09) & 2.12 (.16) & 1.38 (.02) & 1.52 (.02) & \textbf{1.24 (.02)} \\
    & & $10^4$ & 2.33 (.04) & 2.45 (.07) & 2.73 (.09) & 2.38 (.06) & 2.92 (.33) & \textbf{1.57 (.02)} \\
    \cline{2-9}
    & 1.0 & $10^2$ & 1.64 (.07) & 1.81 (.06) & 1.34 (.06) & 1.15 (.02) & \textbf{1.11 (.01)} & \textbf{1.09 (.01)} \\
    & & $10^3$ & 1.62 (.04) & 1.75 (.06) & 1.63 (.06) & \textbf{1.12 (.02)} & 1.25 (.00) & \textbf{1.14 (.01)} \\
    & & $10^4$ & 2.35 (.09) & 2.50 (.14) & 1.74 (.03) & \textbf{1.15 (.02)} & 1.30 (.01) & \textbf{1.15 (.01)} \\
    \cline{2-9}
    & 10.0 & $10^2$ & 1.57 (.04) & 1.56 (.06) & 1.29 (.06) & \textbf{1.12 (.02)} & \textbf{1.12 (.01)} & \textbf{1.11 (.01)} \\
    & & $10^3$ & 1.62 (.04) & 1.81 (.07) & 1.49 (.06) & \textbf{1.14 (.02)} & \textbf{1.14 (.01)} & \textbf{1.12 (.01)} \\
    & & $10^4$ & 2.14 (.04) & 2.20 (.07) & 1.53 (.08) & \textbf{1.13 (.02)} & \textbf{1.13 (.01)} & \textbf{1.12 (.01)} \\
\bottomrule
\end{tabular}
\end{table}

\subsubsection{Combination of spatial data and multiple categorical features}
\label{results:simulated:combo}
For our final simulation we wanted to use a combination of spatial data and multiple high-cardinality features, such as often seen in various tabular data applications (See the Airbnb and Craigslist cars examples in Section~\ref{results:real:spatial}). Here we have two uncorrelated categorical features with random terms $b_j$ and $c_k$ and a spatial 2-D location feature with term $d_l$:
\begin{equation}\label{eq:spatialComboModel}
    y_{ijkl} = f(x_{ijkl}) + b_{j} + c_{k} + d_{l} + \varepsilon_{ijkl},
\end{equation}
where $f$ is as before, both categorical features have $q = 3000$ levels, and the spatial feature has $q = 10000$ 2-D locations from the $\mathbb{U}(-10, 10) \times \mathbb{U}(-10, 10)$ grid. We thus estimate 5 variance components: $\theta = [\sigma^2_{e}, \sigma^2_{b}, \sigma^2_{c}, \sigma^2_{d_0}, \sigma^2_{d_1}]$, where $\sigma^2_{b}, \sigma^2_{c}$ are the variances of the two categorical features and $\sigma^2_{d_0}, \sigma^2_{d_1}$ are the location feature's RBF kernel variances. We vary $[\sigma^2_{b}, \sigma^2_{c}, \sigma^2_{d_0}]$ in $(0.3, 3)$ for a total of 8 combinations, where $\sigma^2_e$ and $\sigma^2_{d_1}$ are fixed at 1 and in total $n = 100000$ as before. Here we compare LMMNN to ignoring the RE features, one-hot encoding each of them and embedding each of them. All training details and baseline MLP architectures are identical to those described in Section~\ref{results:simulated:single}.

Table~\ref{tab:simulated:spatialcategorical:mse} and Table~\ref{tab:simulated:spatialcategorical:sigmas} in Appendix~\ref{appx:tables:variance} summarize the mean test MSE and estimated variance components results. As can be seen LMMNN's performance is best by a margin, and it also reaches excellent variance components estimates.

\begin{table}
  \caption{Simulated model with two high-cardinality categorical features and a spatial feature with 2-D locations with a RBF kernel. Mean test MSEs and standard errors in parentheses. Bold results are non-inferior to the best result in a paired t-test.}
  \label{tab:simulated:spatialcategorical:mse}
  \centering
  \begin{tabular}{lll|llll}
\toprule
$\sigma^2_{b}$ & $\sigma^2_{c}$ & $\sigma^2_{d_0}$ & Ignore & OHE & Embed. & LMMNN \\
	\midrule
  0.3 & 0.3 & 0.3 & 2.05 (.03) & 1.85 (.02) & 1.78 (.01) & \textbf{1.38 (.02)} \\
  & & 3.0 & 2.98 (.07) & 2.24 (.03) & 2.02 (.03) & \textbf{1.42 (.02)} \\
  & 3.0 & 0.3 & 4.82 (.05) & 2.12 (.03) & 2.05 (.04) & \textbf{1.70 (.03)} \\
  & & 3.0 & 5.58 (.04) & 2.51 (.03) & 2.28 (.03) & \textbf{1.68 (.01)} \\
	\midrule
  3.0 & 0.3 & 0.3 & 4.76 (.05) & 2.12 (.04) & 2.01 (.02) & \textbf{1.72 (.01)} \\
  & & 3.0 & 5.67 (.04) & 2.61 (.04) & 2.19 (.04) & \textbf{1.73 (.02)} \\
  & 3.0 & 0.3 & 7.51 (.03) & 2.38 (.02) & \textbf{2.18 (.02)} & \textbf{2.12 (.03)} \\
  & & 3.0 & 8.40 (.08) & 2.74 (.03) & 2.50 (.05) & \textbf{2.13 (.02)} \\
\bottomrule
\end{tabular}
\end{table}

\subsubsection{Scalability of LMMNN}
\label{results:simulated:scale}
\begin{figure}
	\centering
	\includegraphics[width=1.0\linewidth]{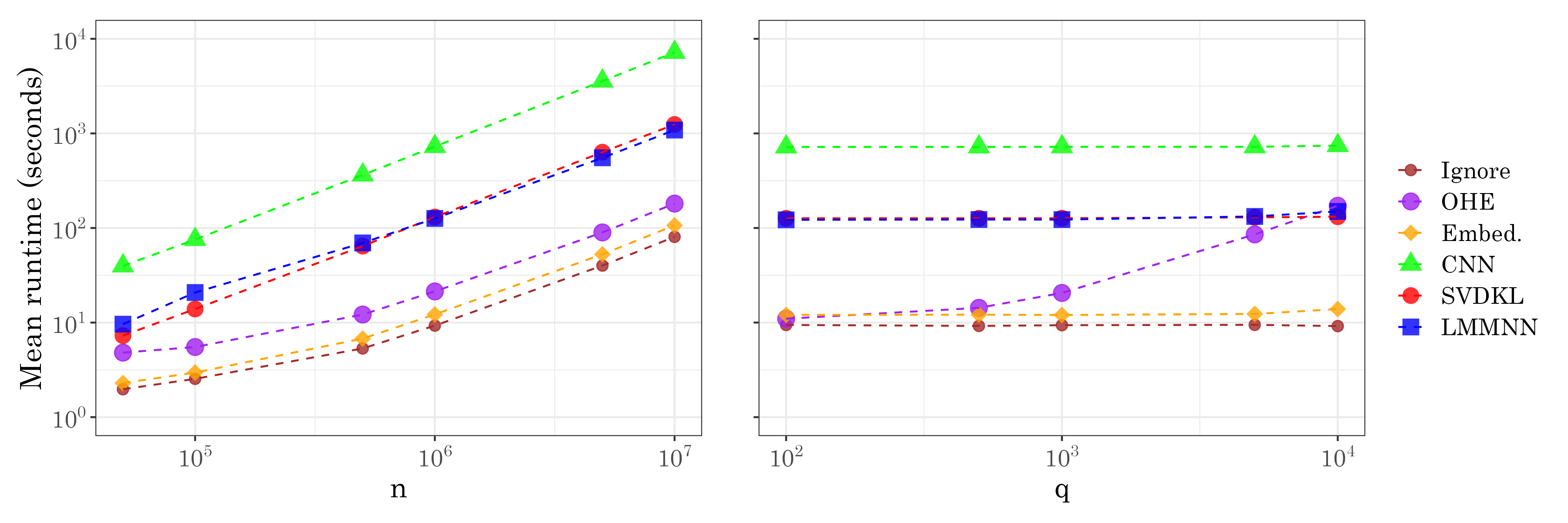}
	\caption{Testing the scalability of LMMNN in the spatial data scenario, running 50 epochs, a batch size of 1000 and all else as described in Section~\ref{results:simulated:spatial}. Left: $q$ is fixed on 1000 location and $n$ is varied; Right: $n$ is fixed on 100000 and $q$ is varied. Note both axes are in the $\log_{10}$ scale.}
	\label{fig:spatial_increase_n_q}
\end{figure}

We give detailed results regarding mean runtime and number of epochs for each and every experiment in Appendix~\ref{appx:tables:times}. However, to demonstrate the scalability of LMMNN compared to other methods we choose to add here an additional set of experiments on one of the challenging covariance scenarios - the spatial data case. In Figure~\ref{fig:spatial_increase_n_q} (left) we record the mean runtime for running 50 epochs of LMMNN and other methods where we keep $q$ fixed on 1000 locations and vary the total number of observations $n$ from 1000 to 1 million. In Figure~\ref{fig:spatial_increase_n_q} (right) we keep $n$ fixed on 100000 observations and vary the number of locations $q$ from 100 to 10000. Both plots show how LMMNN in the spatial scenario scales similar to SVDKL and much better than treating locations as images and applying CNN. Furthermore, LMMNN suffers little in performance when $q$ is increasing, as opposed to OHE, for which a vanilla implementation hardly scales for $q$ over 10000. Similar profiles can be seen for the rest of the covariance scenarios discussed in this paper.

\subsection{Real Data}\label{results:real}
\subsubsection{Multiple categorical features}
\label{results:real:multiple}
We show a number of real tabular datasets with two to five high-cardinality categorical features. For additional examples using a single categorical feature see our previous paper \citep{lmmnn_neurips}. Table~\ref{tab:real:multiple:desc} describes key characteristics of these datasets, with $q$ ranging from 14 to 72K. For more details and where to obtain these publicly available datasets see Appendix~\ref{appx:realdetails}. For all datasets we used a MLP with two hidden layers of 10 and 3 neurons, and a 5-CV procedure. All other details including batch size and early stopping are identical to those described in Section~\ref{results:simulated:single}. Table~\ref{tab:real:multiple:mse} summarizes the mean test MSE results, where LMMNN performs the best with lme4 in close second. Notice that in the UKB-blood example, where there seems to be little advantage to using the categorical features, OHE and entity embeddings tend to overfit and perform worse than ignoring those features, while LMMNN does not, due to its ability to fit very low variance components to these features, thus performing a type of regularization. Finally we note for the UKB-blood example we tried modeling other blood substance levels for cancer patients, such as protein, calcium, glucose and vitamin D -- in all LMMNN achieved the best mean test MSE. Additional summaries of mean running times and number of epochs appear in Table~\ref{tab:real:multiple:times} in Appendix~\ref{appx:tables:times}, and plots of category size distribution and predicted $y_{te}$ versus true appear in Figure~\ref{fig:real_viz_categorical} in Appendix~\ref{appx:realplots}.

\begin{table}
  \caption{Real datasets with $K$ categorical features: summary table}
  \label{tab:real:multiple:desc}
  \centering
  \begin{tabular}{lllllll}
    \toprule
    Dataset & $n$ & $K$ & $p$ & categorical & $q$ & $y$ \\
    \midrule
    \multirow{2}{*}{}
    Imdb   & 86K & 2 & 159 & director & 38K & Movie avg. score (1-10)\\
           &     &   &   & movie type & 1.7K & \\
    \midrule
    \multirow{2}{*}{}
    News   & 81K & 2 & 176 & source & 5.4K & News item FB \\
           &     &   &   & title & 72K & no. of shares (log)\\
    \midrule
    \multirow{3}{*}{}
    InstEval   & 73K & 3 & 3 & student & 2.9K & Teacher ratings (1-5)\\
               &     &   &   & teacher & 1.1K & \\
               &     &   &   & department & 14 & \\
    \midrule
    \multirow{4}{*}{}
    Spotify   & 28K & 4 & 14 & artist & 10K & Song danceability (0-1)\\
           &     &   &   & album & 22K & \\
           &     &   &   & playlist & 2.3K & \\
           &     &   &   & subgenre & 553 & \\
    \midrule
    \multirow{5}{*}{}
    UKB-blood   & 42K & 5 & 19 & treatment & 1.1K & Cancer patient \\
           &     &   &   & operation & 2.0K &  Triglycerides level\\
           &     &   &   & diagnosis & 2.1K &  (mmol/L, standardized)\\
           &     &   &   & cancer type & 446 & \\
           &     &   &   & histology & 359 & \\
    \bottomrule
 \end{tabular}
\end{table}
\begin{table}
  \caption{Real datasets with $K$ categorical features: Mean test MSEs and standard errors in parentheses. Bold results are non-inferior to the best result in a paired t-test.}
  \label{tab:real:multiple:mse}
  \centering
  \begin{tabular}{llllll}
    \toprule
    Dataset & Ignore & OHE & Embed. & lme4 & LMMNN \\
    \midrule
    Imdb       & 1.44 (.01) & -- & 1.26 (.12) & \textbf{0.99 (.01)} & \textbf{0.97 (.01)} \\
    News       & 3.22 (.02) & -- & 1.89 (.02) & \textbf{1.80 (.01)} & \textbf{1.81 (.02)} \\
    InstEval   & 1.77 (.01) & 1.48 (.01) & 1.50 (.01) & \textbf{1.45 (.01)} & \textbf{1.45 (.00)} \\
    Spotify    & .015 (.002) & -- & .016 (.001) & .011 (.000) & \textbf{.009 (.000)} \\
    UKB-blood  & 0.88 (.01) & 1.01 (.01) & 1.04 (.02) & 0.88 (.01) & \textbf{0.86 (.01)} \\
    \bottomrule
 \end{tabular}
\end{table}

\subsubsection{Longitudinal data and repeated measures}
\label{results:real:longitudinal}
Table~\ref{tab:real:longitudinal:desc} summarizes key features of some datasets in which $q$ units of measurement are repeatedly measured through time. $q$ varies from about 1000 stores in the Rossmann dataset with 25-32 monthly measures of total sales, to almost 470K patients in the UK Biobank dataset, with 1-4 measurements of systolic blood pressure (SBP). For more details and where to obtain these publicly available datasets see the Appendix~\ref{appx:realdetails}. As in Section~\ref{results:simulated:longitudinal} for each dataset we either perform a random 5-CV where 80\% of the data is used to predict 20\% of the data (Random mode), or perform 5-CV on past 80\% observations, to predict the latest 20\% observations (Future mode). For the Rossmann and AUimport datasets we use the four layer MLP architecture used in simulations in Section~\ref{results:simulated} and random terms in $t$ up to a quadratic with no correlations between these terms. For the UKB-SBP dataset we use the two layer MLP architecture used in Section~\ref{results:real:multiple}, with random terms in $t$ up to linear (a.k.a random slopes) and no correlations between these terms. Table~\ref{tab:real:longitudinal:mse} summarizes the mean test MSE results and as can be seen LMMNN performs best. R's lme4 performs considerably better than any DNN approach, but it is inferior to LMMNN which has the benefit of introducing non-linearity to the fixed features. Additional summaries of mean running times and number of epochs appear in Table~\ref{tab:real:longitudinal:times} in Appendix~\ref{appx:tables:times}, and plots of the distribution of number of repeated measures $n_j$ and predicted $y_{te}$ versus true appear in Figure~\ref{fig:real_viz_longitudinal} in Appendix~\ref{appx:realplots}.

\begin{table}
  \caption{Longitudinal datasets with $q$ measurement units: summary table}
  \label{tab:real:longitudinal:desc}
  \centering
  \begin{tabular}{llllllll}
    \toprule
    Dataset & $n$ & $n_j$ & $p$ & unit & $q$ & $t$ & $y$ \\
    \midrule
    Rossmann   & 33K & 25-31 & 23 & store & 1.1K & 2013-2015 (mon.) & Total \$ sales (in 100K) \\
    AUimport & 125K & 1-29 & 8 & commodity & 5K & 1988-2016 (year) & Total \$ import (log) \\
    UKB SBP & 528K & 1-4 & 50 & person & 469K & 38-83 (age) & Systolic BP (in 100s) \\
    \bottomrule
 \end{tabular}
\end{table}
\begin{table}
  \caption{Longitudinal datasets with $q$ measurement units: Mean test MSEs and standard errors in parentheses. Bold results are non-inferior to the best result in a paired t-test.}
  \label{tab:real:longitudinal:mse}
  \centering
  \begin{tabular}{lllllll}
    \toprule
    \multicolumn{7}{c}{\textbf{Mode: Random}} \\
    \midrule
    Dataset & Ignore & OHE & Embed. & lme4 & LSTM & LMMNN \\
    \midrule
    Rossmann & .179 (.01) & .052 (.01) & .052 (.01) & .013 (.00) & .505 (.01) & \textbf{.010 (.00)} \\
    AUimport & 7.78 (.70) & 4.91 (.30) & 3.35 (.45) & \textbf{0.72 (.01)} & 8.44 (.35) & \textbf{0.71 (.01)} \\
    UKB SBP & .0321 (.00) & -- & .0327 (.00) & .0310 (.00) & -- & \textbf{.0307(.00)} \\
    \midrule
    \multicolumn{7}{c}{\textbf{Mode: Future}} \\
    \midrule
    Rossmann & .215 (.01) & .067 (.01) & .087 (.02) & .026 (.00) & .336 (.00) & \textbf{.020 (.00)} \\
    AU Import & 7.69 (.48) & 5.60 (1.22) & 3.70 (.12) & 1.77 (.00) & 11.7 (1.1) & \textbf{1.48 (.02)} \\
    UKB SBP & .0387 (.00) & -- & .0396 (.00) & .0383 (.00) & -- & \textbf{.0379 (.00) } \\
    \bottomrule
 \end{tabular}
\end{table}

\subsubsection{Spatial data and spatial-categorical combinations}
\label{results:real:spatial}
Table~\ref{tab:real:spatial:desc} summarizes key features of some datasets in which $q$ geographical locations are repeatedly measured for different quantities. $q$ varies from about 1.2K locations in Japan where radiation was measured by the Safecast organization, to 12K locations across the United States where used cars were sold through Craigslist. The first three datasets come from the US census and the CDC, where each of 3K counties has a few census tract-level measurements of mean annual income, asthma rate in adults and PM2.5 particles. Two of the datasets also fit our spatial and categorical combination scenario: the Craigslist cars dataset, which has 15K cars models, and the Airbnb dataset from \citet{kalehbasti2019airbnb} which has NYC Airbnb listings from 40K hosts. For more details and where to obtain these publicly available datasets see Appendix~\ref{appx:realdetails}.

As usual, a 5-CV procedure is performed where 80\% of the data is used to predict 20\% of the data. For all datasets we use a simple two layer MLP with 10 and 3 neurons, ReLU activation and train until no improvement is seen in 10 epochs, in 10\% validation set. As in simulations, LMMNN-E is the LMMNN version without assuming a RBF kernel, where the 2-D locations pass through a standard MLP with 7 layers of $(100, 50, 20, 10, 20, 50, 100)$ neurons, before entering a standard NLL layer as if it were a single RE feature of dimension 100 with a single variance parameter. As can be seen LMMNN assuming a RBF kernel (LMMNN-R) achieves the best or not inferior from the best mean test MSE. When in addition to the spatial data we take into account high-cardinality features such as a car's model, in a single covariance structure, the improvement in test MSE is substantial. The mean test MSE achieved for the Airbnb dataset is far better than the best test MSE (0.147) reported by \citet{kalehbasti2019airbnb}, who also tried using boosting and support vector machines. More details such as mean running times appear in Table~\ref{tab:real:spatial:times} in Appendix~\ref{appx:tables:times}, and plots of the distribution of $n_j$ measurements in location and predicted $y_{te}$ versus true appear in Figure~\ref{fig:real_viz_spatial} in Appendix~\ref{appx:realplots}.

\begin{table}
  \caption{Spatial datasets with $q$ locations and an optionally high-cardinality categorical feature: summary table}
  \label{tab:real:spatial:desc}
  \centering
  \begin{tabular}{llllllll}
    \toprule
    Dataset & $n$ & $n_j$ & $p$ & $q$ locations & categorical & $y$ \\
    \midrule
    Income & 71K & 1-2K & 30 & 3K US counties & -- & Ann. income \$ (log) \\
    Asthma & 69K & 1-2K & 31 & 3K US counties & -- & Adult asthma \% \\
    AirQuality & 71K & 1-2K & 32 & 3K US counties & -- & PM2.5 1/1/2016 (log) \\
    Radiation & 650K & 1-40K & 3 & 1.2K Japan locs. & -- & CPM (log) \\
    Airbnb & 50K & 1-404 & 196 & 2.8K NYC locs. & host (40K) & Price \$ (log) \\
    Cars & 97K & 1-632 & 73 & 12K US locs. & model (15K) & Price \$ (log) \\
    \bottomrule
 \end{tabular}
\end{table}
\begin{table}
  \caption{Spatial datasets with $q$ locations: Mean test MSEs and standard errors in parentheses. Bold results are non-inferior to the best result in a paired t-test.}
  \label{tab:real:spatial:mse}
  \centering
  \begin{tabular}{lllllll}
    \toprule
    \multicolumn{7}{c}{\textbf{Without high-cardinality categorical features}} \\
    \midrule
    Dataset & Ignore & Embed. & CNN & SVDKL & LMMNN-E & LMMNN-R \\
    \midrule
    Income  & .034 (.00) & .032 (.00) & .032 (.00) & \textbf{.030 (.00)} & \textbf{.027 (.00)} & \textbf{.028 (.00)} \\
    Asthma  & .352 (.01) & .226 (.01) & .259 (.01) & .240 (.01) & .258 (.01) & \textbf{.209 (.00)} \\
    AirQuality  & .285 (.02) & .260 (.04) & .163 (.06) & .044 (.01) & .088 (.02) & \textbf{.035 (.00)} \\
    Radiation  & .354 (.01) & .254 (.02) & .251 (.01) & \textbf{.217 (.00)} & \textbf{.222 (.00)} & \textbf{.219 (.00)} \\
    Airbnb     & .156 (.00) & .196 (.01) & .154 (.00) & .151 (.00) & \textbf{.148 (.00)} & .150 (.00) \\
    Cars       & .152 (.00) & .118 (.00) & .137 (.00) & .149 (.00) & .136 (.00) & \textbf{.109 (.00)} \\
    \midrule
    \multicolumn{7}{c}{\textbf{With high-cardinality categorical features}} \\
    \midrule
    Airbnb     & .156 (.00) & .177 (.01) & -- & -- & -- & \textbf{.139 (.00)} \\
    Cars       & .152 (.00) & .092 (.00) & -- & -- & -- & \textbf{.084 (.00)} \\
    \bottomrule
 \end{tabular}
\end{table}

\section{Classification Setting: A Prelude}\label{classification}
In this section we start with revisiting the random intercepts model. Let $y_{ij} | b_j$ be the $i$-th measurement of cluster $j$, which is dependent on some random intercept $b_j$. $j = 1, \dots, q$ and $i = 1, \dots, n_j$, where $n_j$ is as before the number of observations for cluster $j$, and we usually assume $b_j \sim \mathbb{N}(0, \sigma^2_b)$, where $\sigma^2_b$ is a variance component as before. Let us develop the marginal NLL from scratch, writing $f_Y$, $f_b$ and $f_{Y|b}$ for $y$'s, $b$'s and $y|b$'s distribution functions respectively:
\begin{equation}\label{eq:glmm-nll1}
    \begin{aligned}
    NLL(\sigma^2_b|y) &= -\log{L(\sigma^2_b | y)} = -\log{\prod_{ij}{f_Y(y_{ij})}} \\
 &= -\log{\prod_{ij}{\int f_{Y | b}(y_{ij}|b_j)f_b(b)\,db}} = -\sum_{j=1}^{q}\log\{\prod_{i=1}^{n_j}{\int f_{Y | b}(y_{ij}|b_j)f_b(b_j)\,db_j}\}
    \end{aligned}
\end{equation}
Previously we utilized the assumption of $f_{Y|b}, f_b$ distributed normal, therefore the marginal $f_Y$ was normal as well, and the integral in \eqref{eq:glmm-nll1} could be written in closed form. When dealing with generalized linear mixed models (GLMM), however, where the response $y$ is far from normal, we see the marginal NLL contains an integral over the RE which is difficult to write in closed form and to minimize over the variance component parameters. In some cases however, such as random intercepts with a single categorical variable and a binary response variable $y$, we can approximate the NLL with Gauss-Hermite quadrature \citep{McCulloch2008}. Having done that, we can proceed within the LMMNN framework, to handle high-cardinality categorical features in DNNs for classification settings as well.

A binary response $y_{ij} | b_j \in \{0, 1\}$ is usually modeled with a Bernoulli distribution. We write $y_{ij}|b_j \sim \mathbb{B}(p_{ij})$, where $p_{ij}$ is the expectation of $y_{ij}|b_j$ in $[0, 1]$. Replacing in \eqref{eq:glmm-nll1} the Bernoulli distribution function for $f_{Y|b}$ and the normal distribution for $f_b$ we get:
\begin{equation}\label{eq:glmm-nll2}
    NLL(\sigma^2_b, p_{ij}|y) = -\sum_{j=1}^{q}\log\{\prod_{i=1}^{n_j}{\int p_{ij}^{y_{ij}}(1 - p_{ij})^{1-y_{ij}}\frac{e^{-b_j^2 / 2\sigma^2_b}}{\sqrt{2\pi\sigma^2_b}}\,db_j}\}
\end{equation}

Now in GLMM one models not the expectation $p_{ij}$ directly. Instead, a link function $\eta(p_{ij})$ is used, which maps $p_{ij}$ into $(-\infty, +\infty)$. For some explaining variables $x_{ij} \in \mathbb{R}^p$ we write $\eta(p_{ij}) = x'_{ij}\beta + b_j$, where $\beta \in \mathbb{R}^p$ are fixed parameters to estimate. In the LMMNN framework we write:
\begin{equation}\label{eq:glmm-link-func}
    \eta(p_{ij}) = f(x_{ij}) + b_j,
\end{equation}
where $f$ is a non-linear function which we model via a DNN. As before, the RE $b_j$ might pass through another function $g$, modeled by the same or different network. Now mark $f(x_{ij}) = f_{ij}$ and introduce the $\text{logit}$ function, which is the most common link function for a Bernoulli response variable. The model in \eqref{eq:glmm-link-func} becomes:
\begin{equation}\label{eq:link-func-logit}
    \log{\frac{p_{ij}}{1 - p_{ij}}} = f_{ij} + b_j = \eta_{ij},
\end{equation}

Back to the NLL in \eqref{eq:glmm-nll2}, after some algebraic manipulation, we can write:
\begin{equation}\label{eq:glmm-nll3}
    NLL(\sigma^2_b, f | y) = -\sum_{j=1}^{q}\log\{\int \exp\{\sum_i{y_{ij}\eta_{ij} - \log(1 + e^{\eta_{ij}})}\}\frac{e^{-b_j^2 / 2\sigma^2_b}}{\sqrt{2\pi\sigma^2_b}}\,db_j\},
\end{equation}

For using Gauss-Hermite quadrature we need each of the $q$ integrals to be of form $\int{h(v)e^{-v^2}\,dv}$. Define:
\begin{equation}\label{eq:glmm-quad-h}
    \begin{aligned}
        h_j(b_j) &= \exp\{\sum_{i = 1}^{n_j}{y_{ij}\eta_{ij} - \log(1 + e^{\eta_{ij}})}\} \\
        h^*_j(z) &= h_j(\sqrt{2}\sigma_b z) / \sqrt{\pi}
    \end{aligned},
\end{equation}
where $b_j$ enters $h_j$ via $\eta_{ij}$, the logits. Then:
\begin{equation}
    NLL(\sigma^2_b, f | y) = -\sum_{j=1}^{q}\log\{\int h^*_j(v_j)e^{-v_j^2}\,dv_j\},
\end{equation}
where $v_j = b_j / \sqrt{2}\sigma_b$.

Now we can use Gauss-Hermite quadrature to approximate each of the $q$ integrals with a sum over $K$ elements:
\begin{equation}
    \int h^*_j(v_j)e^{-v_j^2}\,dv_j \approx \sum_{k = 1}^{K}{h^*_j(x_k)w_k},
\end{equation}
where $x_k$ is the $k$th zero of $H_n(x)$, the Hermite polynomial of degree $n$, and both $x_k, w_k$ can be obtained from any mathematical software (not to be confused with the $x_{ij}$ covariates!). The approximation should be better the higher we take $K$. The NLL now simplifies to a relatively simple sum:

\begin{equation}\label{eq:glmm-nll4}
    NLL(\sigma^2_b, f | y) \approx -\sum_{j=1}^{q}\log\{\sum_{k = 1}^{K}{\exp\left[\sum_{i = 1}^{n_j}\left(y_{ij}(f_{ij} + \sqrt{2}\sigma_b x_k) - \log{(1 + e^{f_{ij} + \sqrt{2}\sigma_b x_k})}\right)\right]\frac{w_k}{\sqrt{\pi}}}\}
\end{equation}
For prediction of $b$, we use quadrature in a very similar way, following \citet{McCulloch2008}. Finally, note that the NLL and therefore its gradient can be naturally decomposed to $q$ separate computations, each on the group of $n_j$ observations for cluster $j$, thus using gradient descent in mini-batches to optimize it, is once again justified.

To demonstrate how non-linear GLMM can be fitted in the LMMNN framework, we perform a simulation in which $y$ is binary, and its expectation depends on $X$ in a very similar way to \eqref{eq:f}:
\begin{equation}\label{eq:glmm-f}
    \text{logit}(p_{ij}) = (X_1 + \dots + X_{10}) \cdot \cos(X_1 + \dots + X_{10}) + 2 \cdot X_1 \cdot X_2 + Zb
\end{equation}
We have a single categorical variable with $q$ varying in $\{100, 1000, 10000\}$, and $\sigma^2_b$ varying in $\{0.1, 1, 10\}$. As in Section~\ref{results:simulated} we sample different $n_j$ observations for each level $j$ of the categorical feature, the $X$ features come from a uniform distribution and $n = 100000$ always. We split the data to 80\% training and 20\% testing and use the same network architecture, batch size and early stopping details as in Section~\ref{results:simulated}. The loss for regular DNNs is the standard binary cross-entropy, and for LMMNN the NLL in \eqref{eq:glmm-nll4} is used. For Gauss-Hermite quadrature we use $K = 5$ roots. We use the area under the ROC curve (AUC) to compare LMMNN's results to ignoring the categorical feature, using OHE and entity embeddings. We also compare results to the lme4's glmer function. Table~\ref{tab:simulated:glmm:auc} summarizes the mean test AUC and Table~\ref{tab:simulated:glmm:sigmas} in Appendix~\ref{appx:tables:variance} summarizes the $\sigma^2_b$ estimates of LMMNN and glmer. As can be seen, for low cardinality $q$ LMMNN's performance is not significantly better than the best competitors, while for high $q$ it performs better, though with a considerable cost in runtime, as can be seen in Table~\ref{tab:simulated:glmm:times} in Appendix~\ref{appx:tables:times}.

\begin{table}
  \caption{Simulated binary GLMM model with $g(Z) = Z$, mean test AUCs and standard errors in parentheses (higher is better). Bold results are non-inferior to the best result in a paired t-test.}
  \label{tab:simulated:glmm:auc}
  \centering
  \begin{tabular}{llllllll}
\toprule
$\sigma^2_b$ & $q$ & Ignore & OHE & Embeddings & lme4 & LMMNN \\
	\midrule
0.1 & $10^2$   & \textbf{0.79 (.001)} & \textbf{0.79 (.002)} & \textbf{0.79 (.002)} & 0.67 (.003) & \textbf{0.80 (.001)} \\
    & $10^3$  & \textbf{0.79 (.001)} & 0.75 (.001) & 0.77 (.001) & 0.66 (.001) & \textbf{0.79 (.001)} \\
    & $10^4$ & \textbf{0.79 (.001)} & 0.67 (.002) & 0.67 (.002) & 0.66 (.001) & \textbf{0.79 (.001)} \\
\midrule
1 & $10^2$ & 0.77 (.002) & \textbf{0.82 (.003)} & \textbf{0.83 (.003)} & 0.73 (.005) & \textbf{0.82 (.003)} \\
    & $10^3$  & 0.76 (.003) & 0.79 (.002) & 0.81 (.002) & 0.73 (.002) & \textbf{0.83 (.001)} \\
    & $10^4$ & 0.76 (.002) & 0.71 (.002) & 0.71 (.002) & 0.70 (.001) & \textbf{0.80 (.001)} \\
\midrule
10 & $10^2$ & 0.67 (.005) & \textbf{0.93 (.002)} & \textbf{0.93 (.001)} & 0.90 (.004) & \textbf{0.92 (.001)} \\
    & $10^3$ & 0.67 (.003) & \textbf{0.91 (.002)} & \textbf{0.92 (.001)} & 0.90 (.001) & \textbf{0.92 (.001)} \\
    & $10^4$ & 0.66 (.002) & 0.87 (.001) &0.87 (.001) & 0.87 (.001) & \textbf{0.90 (.001)} \\
\bottomrule
\end{tabular}
\end{table}

We further tested LMMNN on real datasets encountered in Section~\ref{results:real:spatial}. For the Airbnb dataset we predict whether a listing has air conditioning or not (84\% do). The categorical feature here is the listing's host with $q = 40K$, and $p = 196$ features as before. For the Cars dataset we predict whether a car is located at the west of USA or not, by checking whether its longitude coordinate is larger than 100 (66\% are). The categorical feature here is the car's model with $q = 15K$, and $p = 73$ features as before. We use the same two-layer architecture of 10 and 3 neurons and a 5-CV training procedure, with all other details identical to previous simulations. Table~\ref{tab:real:glmm:auc} summarizes the mean test AUC, where it is clear that our approach performs best. Table~\ref{tab:real:glmm:times} in Appendix~\ref{appx:tables:times} summarizes mean runtime and number of epochs.

\begin{table}
  \caption{Classification datasets with a single categorical feature: Mean test AUCs and standard errors in parentheses. Bold results are non-inferior to the best result in a paired t-test.}
  \label{tab:real:glmm:auc}
  \centering
  \begin{tabular}{llllll}
    \toprule
    Dataset & Ignore & OHE & Embed. & lme4 & LMMNN \\
    \midrule
    Airbnb       & 0.79 (.005) & -- & 0.76 (.002) & -- & \textbf{0.82 (.003)} \\
    Cars       & 0.70 (.001) & 0.68 (.003) & 0.69 (.002) & 0.66 (.002) & \textbf{0.72 (.002)} \\
    \bottomrule
 \end{tabular}
\end{table}

\section{Conclusion}\label{discussion}
In this paper we presented LMMNN as a general framework for dealing with covariance structures for correlated data, including clustering due to categorical variables, spatial and temporal structures and combinations of these. One important aspect of our contribution is the use of NLL loss within the deep learning framework. Since this loss does not naturally decompose to a sum over observations, the use of standard SGD approaches is challenging, and in Section~\ref{theory} we demonstrated that the approach of inverting small sub-matrices to make SGD practical has some theoretical justifications and works well in practice. We showed in extensive simulations and real data analyses that LMMNN's predictive performance is never inferior to common solutions for handling correlated data in DNNs, and in many cases superior to these solutions, especially when compared to OHE and entity embeddings for encoding categorical features, and LSTM for longitudinal datasets. We find LMMNN to be especially useful for handling tabular datasets as often encountered in business and healthcare applications, where a few features inject correlations of different nature into the data. In the Airbnb and Cars datasets for example, we showed how using LMMNN with a single covariance structure to handle both spatial and high-cardinality categorical features can perform very well, with a reasonable cost in running time. We also offered in Section~\ref{classification} preliminary methodology for extending LMMNN to classification settings, with promising results. 

In the future we hope to make LMMNN more efficient, easy to use in additional common DNN frameworks such as PyTorch, and relevant to complex classification settings. All simulations and code used for making the experiments and visualizations in this paper are available on Github at \url{https://github.com/gsimchoni/lmmnn/}.

% Acknowledgements should go at the end, before appendices and references

\acks{We thank the anonymous reviewers and Action Editor for their useful comments and suggestions. This study was supported in part by a fellowship from the Edmond J. Safra Center for Bioinformatics at Tel-Aviv University, by Israel Science Foundation grant 2180/20 and by Israel Council for Higher Education Data-Science Centers. UK Biobank research has been conducted using the UK Biobank Resource under Application Number 56885.}

\newpage

\appendix
\renewcommand{\thesection}{\arabic{section}}

\section{The eigendecay of the multiple categorical features covariance matrix}\label{appx:multiple}
Suppose we model $L$ uncorrelated features each having $q_l$ levels, using \eqref{eq:multipleCategoricalV}. Let $\sigma^2_0 = \sigma^2_e$ and $Z_0 = I$. Then we can write \eqref{eq:multipleCategoricalV} as a sum of $L + 1$ covariance matrices:
\begin{equation}\label{multipleCatVAsSum}
    V(\theta) = \sum_{l = 0}^L \sigma^2_{l}Z_lZ'_l
\end{equation}
Each of the $V_l = \sigma^2_{l}Z_lZ'_l$ \textit{could be} written as a block-diagonal matrix with $q_l$ blocks, if $Z_l$ is properly sorted, let this be $V^*_l$. In this case $V^*_l$'s eigenvalues are those blocks eigenvalues. Each block $j$ is of size $n_j \times n_j$, where $n_j$ is the number of observations of level $j$ ($j = 1, \dots, q_l)$, and can be written as $\sigma^2_{l}\mathbf{1}\mathbf{1}'$, where $\mathbf{1}$ is an all ones vector of length $n_j$. Hence each block is of rank 1 and has $n_j - 1$ zero eigenvalues, the remaining eigenvalue has to be positive and equal to the block's trace $\sigma^2_{l}n_j$. The entire spectrum of the block-diagonal $V^*_l$ then, are those $q_l$ eigenvalues $\sigma^2_{l}n_j$ and the remaining $n - q_l$ are zeros. The range of the block-diagonal $V^*_l$'s spectrum is therefore $[0, \sigma^2_l\max{n_j}]$, and its eigendecay depends on the decay of $n_{q_l}, \dots, n_1$ where we assume these are sorted. At any case the eigenvalues starting from the $q_l$-th location are all zeros. While $V_l$ isn't necessarily block-diagonal (since $Z_l$ isn't necessarily sorted), its eigenvalues and eigendecay remain unchanged from those of $V^*_l$. To see this consider the fact that $V^*_l$ is a symmetric matrix whose rows and columns have been permuted in the exact same manner, which is equivalent to left and right multiplying it by an orthogonal matrix $P$ of dimension $q_l \times q_l$. $V_l$ could be written as $PV^*_lP'$, and from here it is easy to see its characteristic polynomial and therefore its eigenvalues are identical to those of $V^*_l$. Finally as mentioned in the text since each of $V_l$ can be seen as a kernel with a fast eigendecay with rate $C_l \cdot i^{-p}$, their sum $V$ is also a kernel with a fast eigendecay with rate $C \cdot i^{-p}$, where $C_l, C$ are some constants. Therefore \citet{Chen2020} theorems apply to it.

\begin{figure}[H]
	\centering
	\includegraphics[width=1.0\linewidth]{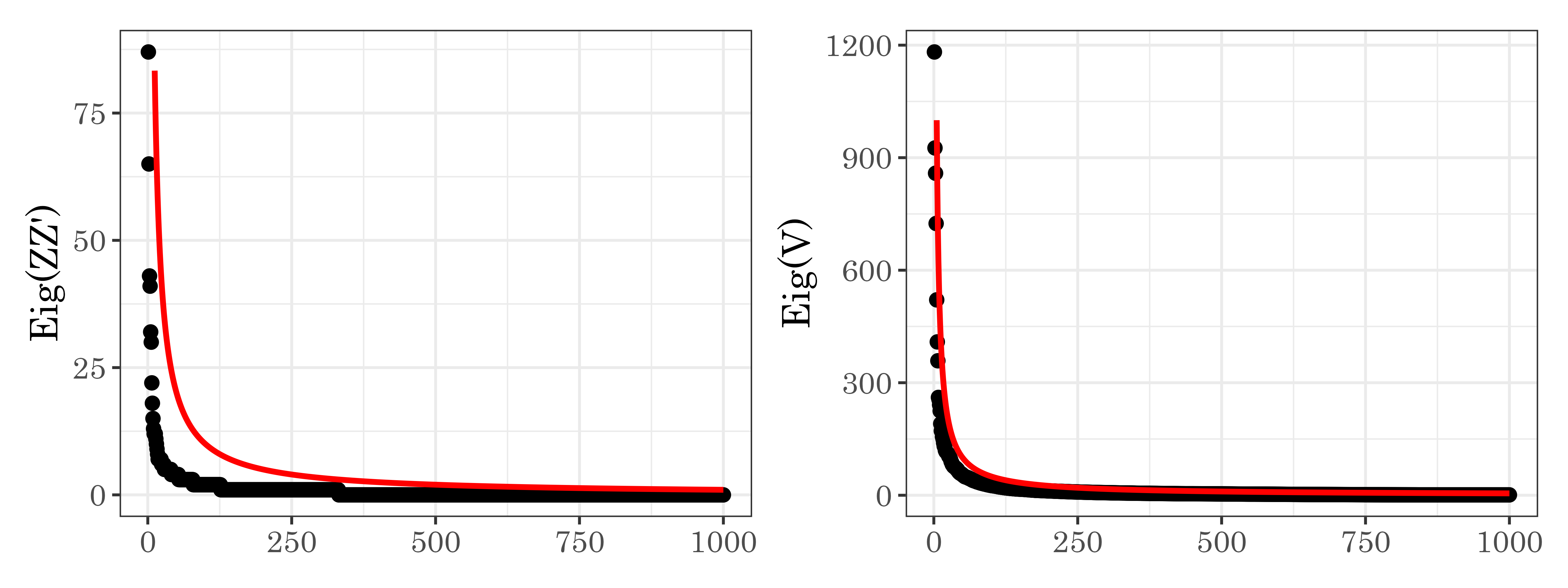}
	\caption{Eigendecay of covariance matrices of a sample of $n = 1000$ UK Biobank subjects with cancer history (black points) and a $C\cdot i^{-p}$ function (red line). All $\sigma^2_{bk}$ are 1. Left: a single categorical feature of diagnosis ($q = 338$ in sample), $C = 1000, p = 1$. Right: The entire $V(\theta)$ of 5 categorical features as described in Figure~\ref{fig:ukb_V_matrices}, $C = 5000, p = 1$}
	\label{fig:ukb_eigendecay}
\end{figure}

Figure~\ref{fig:ukb_eigendecay} presents actual eigendecays for the UKB sample described in Figure~\ref{fig:ukb_V_matrices}, with a simple decay function such as $C \cdot i^{-p}$, where $p = 1$ but can be larger. We see nicely how in realistic situations the number of observations for levels of a high-cardinality categorical feature decays fast. For these covariance matrices it is therefore suitable to apply \citet{Chen2020}'s theorems for bounding the NLL gradient by fitting the LMMNN model using SGD.

\section{Simulated Data: variance components estimates}
\label{appx:tables:variance}

\begin{table}[H]
  \caption{Simulated model with a single categorical feature, estimated variance components on average.}
  \label{tab:simulated:single:sigmas}
  \centering
  \begin{tabular}{ll|llll|llll|llll}
    \toprule
		&	& \multicolumn{4}{c|}{\textbf{g(Z) = Z}} & \multicolumn{4}{c|}{\textbf{g(Z) = ZW}} & \multicolumn{4}{c}{\textbf{g(Z) = ZW*cos(ZW)}} \\
		\cline{3-14}
        &    & \multicolumn{2}{c}{lme4}  & \multicolumn{2}{c|}{LMMNN} & \multicolumn{2}{c}{lme4}  & \multicolumn{2}{c|}{LMMNN} & \multicolumn{2}{c}{lme4}  & \multicolumn{2}{c}{LMMNN}\\
    \cline{3-14}
        $\sigma^2_b$ & $q$ & $\hat{\sigma}^2_e$ & $\hat{\sigma}^2_b$ & $\hat{\sigma}^2_e$ & $\hat{\sigma}^2_b$ & $\hat{\sigma}^2_e$ & $\hat{\sigma}^2_b$ & $\hat{\sigma}^2_e$ & $\hat{\sigma}^2_b$ & $\hat{\sigma}^2_e$ & $\hat{\sigma}^2_b$ & $\hat{\sigma}^2_e$ & $\hat{\sigma}^2_b$\\
    \hline
    0.1 & $10^2$   & 2.92 & 0.10 & 1.14 & 0.11 & 2.92 & 0.49 & 1.09 & 1.37 & 2.92 & 0.12 & 1.10 & 1.03  \\
        & $10^3$  & 2.90 & 0.10 & 1.12 & 0.10 & 2.91 & 3.52 & 0.28 & 1.28  & 2.90 & 1.71 & 0.28 & 1.12 \\
        & $10^4$ & 2.90 & 0.10 & 1.14 & 0.10 & 2.91 & 33.8 & 0.12 & 0.33  & 2.91 & 17.6 & 0.11 & 0.26 \\
    \hline
    1 & $10^2$   & 2.92 & 1.03 & 1.12 & 1.08 & 2.91 & 2.44 & 1.08 & 3.41   & 2.89 & 2.17 & 1.09 & 2.66 \\
        & $10^3$  & 2.91 & 0.98 & 1.12 & 1.03 & 2.90 & 32.0 & 0.29 & 2.49  & 2.90 & 20.4 & 0.29 & 2.10 \\
        & $10^4$ & 2.91 & 0.98 & 1.13 & 1.00 & 2.92 & 336.7 & 0.18 & 1.14  & 2.91 & 175.2 & 0.16 & 0.81 \\
    \hline
    10 & $10^2$  & 2.90 & 10.5 & 1.12 & 8.80 & 2.89 & 32.9 & 1.08 & 9.63  & 2.90 & 17.3 & 1.08 & 9.23  \\
        & $10^3$  & 2.89 & 10.0 & 1.12 & 8.68 & 2.89 & 337.8 & 0.33 & 5.48  & 2.91 & 179.7 & 0.31 & 4.36 \\
        & $10^4$ & 2.91 & 10.0 & 1.12 & 10.0 & 2.90 & 3305.6 & 0.30 & 4.73 & 2.92 & 1724.7 & 0.27 & 3.33 \\
    \bottomrule
  \end{tabular}

\end{table}

\begin{table}[H]
  \caption{Simulated model with 3 categorical features, with $q_1 = 1000, q_2 = 5000, q_3 = 10000$. Estimated variance components on average.}
  \label{tab:simulated:multiple:sigmas}
  \centering
  \begin{tabular}{lll|llll|llll}
    \toprule
		\multicolumn{11}{c}{\textbf{g(Z) = Z}} \\
		\hline
        & & & \multicolumn{4}{c|}{lme4} & \multicolumn{4}{c}{LMMNN} \\
    \cline{4-11}
        $\sigma^2_{b1}$ & $\sigma^2_{b2}$ & $\sigma^2_{b3}$ & $\hat{\sigma}^2_e$ & $\hat{\sigma}^2_{b1}$ & $\hat{\sigma}^2_{b2}$ & $\hat{\sigma}^2_{b3}$ & $\hat{\sigma}^2_e$ & $\hat{\sigma}^2_{b1}$ & $\hat{\sigma}^2_{b2}$ & $\hat{\sigma}^2_{b3}$ \\
    \hline
    0.3 & 0.3 & 0.3 & 2.89 & 0.30 & 0.29 & 0.31 & 1.12 & 0.29 & 0.31 & 0.30 \\
        & & 3.0 & 2.91 & 0.28 & 0.30 & 3.00 & 1.12 & 0.30 & 0.30 & 3.01 \\
        & 3.0 & 0.3 & 2.91 & 0.30 & 2.93 & 0.29 & 1.12 & 0.30 & 2.94 & 0.30 \\
        & & 3.0 & 2.90 & 0.29 & 3.04 & 3.00 & 1.12 & 0.31 & 2.94 & 3.08 \\
    \hline
    3.0 & 0.3 & 0.3 & 2.92 & 2.94 & 0.30 & 0.31 & 1.13 & 2.95 & 0.31 & 0.30 \\
        & & 3.0 & 2.90 & 3.09 & 0.32 & 3.01 & 1.13 & 2.93 & 0.30 & 2.96 \\
        & 3.0 & 0.3 & 2.91 & 2.90 & 2.98 & 0.30 & 1.12 & 2.97 & 3.12 & 0.31 \\
        & & 3.0 & 2.90 & 3.08 & 3.01 & 3.03 & 1.12 & 3.11 & 2.92 & 2.98 \\
    \midrule
	\multicolumn{11}{c}{\textbf{g(Z) = ZW}} \\
	\hline
    0.3 & 0.3 & 0.3 & 2.91 & 9.14 & 19.4 & 31.0 & 0.16 & 1.47 & 1.11 & 0.97 \\
        & & 3.0 & 2.9 & 10.2 & 20.1 & 291.6 & 0.19 & 1.09 & 0.73 & 2.21 \\
        & 3.0 & 0.3 & 2.93 & 10.4 & 189.3 & 29.2 & 0.17 & 1.15 & 3.02 & 0.67 \\
        & & 3.0 & 2.91 & 8.19 & 196.2 & 302.9 & 0.2 & 1.15 & 3 & 2.11 \\
    \hline
    3.0 & 0.3 & 0.3 & 2.91 & 101.2 & 18.0 & 31.6 & 0.16 & 5.07 & 0.89 & 0.85 \\
        & & 3.0 & 2.92 & 99.5 & 20.8 & 304.9 & 0.19 & 4.62 & 0.76 & 2.18 \\
        & 3.0 & 0.3 & 2.93 & 98.1 & 195.7 & 28.6 & 0.17 & 4.58 & 2.92 & 0.64 \\
        & & 3.0 & 2.93 & 97.9 & 205.5 & 293.9 & 0.2 & 4.08 & 2.51 & 1.84 \\
    \midrule
	\multicolumn{11}{c}{\textbf{g(Z) = ZW * cos(ZW)}} \\
	\hline
    0.3 & 0.3 & 0.3 & 2.89 & 5.28 & 10.3 & 16.33 & 0.15 & 0.90 & 0.74 & 0.70 \\
        & & 3.0 & 2.90 & 5.20 & 9.70 & 162.1 & 0.18 & 0.70 & 0.56 & 1.85 \\
        & 3.0 & 0.3 & 2.90 & 4.64 & 104.3 & 15.9 & 0.16 & 0.77 & 2.45 & 0.65 \\
        & & 3.0 & 2.90 & 5.32 & 104.6 & 150.8 & 0.19 & 0.62 & 2.29 & 1.74 \\
    \hline
    3.0 & 0.3 & 0.3 & 2.90 & 45.7 & 11.54 & 15.6 & 0.16 & 4.03 & 0.76 & 0.71 \\
        & & 3.0 & 2.92 & 49.8 & 10.03 & 158.8 & 0.18 & 3.24 & 0.57 & 1.71 \\
        & 3.0 & 0.3 & 2.88 & 51.2 & 96.61 & 16.3 & 0.16 & 3.47 & 2.22 & 0.54 \\
        & & 3.0 & 2.91 & 50.9 & 97.98 & 169.3 & 0.19 & 3.25 & 2.07 & 1.57 \\
    \bottomrule
  \end{tabular}
\end{table}

\begin{table}[H]
  \caption{Simulated model with longitudinal data for $q = 10000$ subjects. Estimated variance components on average.}
  \label{tab:simulated:longitudinal:sigmas}
  \centering
  \begin{tabular}{lll|llllll|llllll}
    \toprule
		\multicolumn{15}{c}{\textbf{Mode: Random}} \\
		\hline
        & & & \multicolumn{6}{c|}{lme4} & \multicolumn{6}{c}{LMMNN} \\
    \cline{4-15}
        $\sigma^2_{b0}$ & $\sigma^2_{b1}$ & $\sigma^2_{b2}$ & $\hat{\sigma}^2_e$ & $\hat{\sigma}^2_{b0}$ & $\hat{\sigma}^2_{b1}$ & $\hat{\sigma}^2_{b2}$ & $\hat{\rho}_{01}$ & $\hat{\rho}_{02}$ & $\hat{\sigma}^2_e$ & $\hat{\sigma}^2_{b0}$ & $\hat{\sigma}^2_{b1}$ & $\hat{\sigma}^2_{b2}$ & $\hat{\rho}_{01}$ & $\hat{\rho}_{02}$ \\
    \hline
    0.3 & 0.3 & 0.3 & 2.90 & 0.32 & 1.93 & 3.2 & 0.04 & -0.25 & 1.14 & 0.31 & 0.47 & 0.33 & 0.17 & 0.18 \\
        & & 3.0 & 2.91 & 0.32 & 1.88 & 5.50 & -0.11 & 0.08 & 1.12 & 0.31 & 0.76 & 2.16 & 0.08 & 0.30 \\
        & 3.0 & 0.3 & 2.92 & 0.33 & 4.13 & 4.59 & 0.21 & -0.14 & 1.13 & 0.31 & 2.74 & 1.88 & 0.32 & 0.12 \\
        & & 3.0 & 2.92 & 0.31 & 4.44 & 4.85 & 0.24 & 0.28 & 1.14 & 0.31 & 2.85 & 2.84 & 0.29 & 0.40 \\
    \hline
    3.0 & 0.3 & 0.3 & 2.89 & 3.02 & 1.44 & 4.68 & 0.15 & 0.02 & 1.13 & 3.01 & 0.34 & 0.57 & 0.36 & -0.01 \\
        & & 3.0 & 2.9 & 2.99 & 2.42 & 5.38 & 0.15 & 0.51 & 1.11 & 2.98 & 0.59 & 2.33 & 0.29 & 0.24 \\
        & 3.0 & 0.3 & 2.91 & 2.96 & 4.62 & 3.02 & 0.42 & -0.12 & 1.11 & 3.00 & 2.71 & 1.78 & 0.32 & 0.12 \\
        & & 3.0 & 2.89 & 3.00 & 4.37 & 8.26 & 0.28 & 0.49 & 1.13 & 3.04 & 3.02 & 3.45 & 0.34 & 0.16 \\
    \midrule
	\multicolumn{15}{c}{\textbf{Mode: Future}} \\
	\hline
    0.3 & 0.3 & 0.3 & 2.89 & 0.32 & 1.36 & 16.09 & 0.28 & -0.25 & 1.12 & 0.31 & 0.69 & 0.99 & 0.11 & 0.21 \\
        & & 3.0 & 2.90 & 0.31 & 1.32 & 8.84 & -0.5 & 0.49 & 1.12 & 0.31 & 0.69 & 1.19 & 0.13 & 0.36 \\
        & 3.0 & 0.3 & 2.91 & 0.31 & 3.12 & 17.91 & 0.52 & -0.01 & 1.13 & 0.31 & 2.62 & 2.29 & 0.28 & 0.29 \\
        & & 3.0 & 2.90 & 0.32 & 4.09 & 24.28 & 0.28 & -0.14 & 1.12 & 0.30 & 2.65 & 2.28 & 0.31 & 0.53 \\
    \hline
    3.0 & 0.3 & 0.3 & 2.88 & 3.02 & 1.05 & 10.97 & -0.09 & 0.47 & 1.12 & 3.02 & 0.53 & 0.73 & 0.27 & 0.11 \\
        & & 3.0 & 2.92 & 3.00 & 0.52 & 33.25 & -0.31 & 0.42 & 1.12 & 3.06 & 0.66 & 0.78 & 0.27 & 0.05 \\
        & 3.0 & 0.3 & 2.90 & 2.99 & 3.54 & 31.5 & 0.69 & -0.22 & 1.13 & 2.99 & 2.85 & 2.77 & 0.35 & -0.02 \\
        & & 3.0 & 2.94 & 2.97 & 3.34 & 12.67 & 0.69 & -0.31 & 1.12 & 2.99 & 2.70 & 2.82 & 0.39 & 0.17 \\
    \bottomrule
  \end{tabular}
\end{table}

\begin{table}[H]
  \caption{Simulated model with spatial data with a RBF kernel. Estimated variance components on average.}
  \label{tab:simulated:spatial:sigmas}
  \centering
  \begin{tabular}{lll|lll}
    \toprule
        $\sigma^2_{b0}$ & $\sigma^2_{b1}$ & $q$ & $\hat{\sigma}^2_e$ & $\hat{\sigma}^2_{b0}$ & $\hat{\sigma}^2_{b1}$ \\
    \hline
    0.1 & 0.1 & $10^2$   & 1.12 & 0.12 & 0.71  \\
        & & $10^3$   &  1.12 & 0.10 & 0.27  \\
        & & $10^4$   &  1.13 & 0.11 & 0.12  \\
        \cline{2-6}
        & 1.0 & $10^2$   &  1.13 & 0.11 & 1.77  \\
        & & $10^3$   &  1.12 & 0.10 & 1.12  \\
        & & $10^4$   &  1.13 & 0.10 & 1.08  \\
        \cline{2-6}
        & 10.0 & $10^2$   &  1.13 & 0.13 & 2.16 \\
        & & $10^3$   &  1.15 & 0.12 & 3.11 \\
        & & $10^4$   &  1.13 & 0.11 & 7.71 \\
    \midrule
    1.0 & 0.1 & $10^2$   &  1.13 & 0.90 & 1.29  \\
        & & $10^3$   &  1.12 & 0.99 & 0.48  \\
        & & $10^4$   &  1.13 & 0.98 & 0.10  \\
        \cline{2-6}
        & 1.0 & $10^2$   &  1.13 & 0.93 & 1.11  \\
        & & $10^3$   &  1.12 & 1.10 & 1.49  \\
        & & $10^4$   &  1.15 & 0.91 & 0.83 \\
        \cline{2-6}
        & 10.0 & $10^2$   &  1.12 & 0.91 & 3.05 \\
        & & $10^3$   &  1.11 & 0.74 & 4.93 \\
        & & $10^4$   &  1.11 & 1.13 & 8.69  \\
    \midrule
    10.0 & 0.1 & $10^2$   &  1.12 & 8.07 & 0.50 \\
        & & $10^3$   &  1.11 & 8.99 & 0.12 \\
        & & $10^4$   &  1.13 & 10.11 & 0.11 \\
        \cline{2-6}
        & 1.0 & $10^2$   &  1.12 & 8.39 & 1.01 \\
        & & $10^3$   &  1.12 & 9.24 & 0.86 \\
        & & $10^4$   &  1.12 & 9.00 & 0.99 \\
        \cline{2-6}
        & 10.0 & $10^2$   &  1.13 & 7.04 & 2.68 \\
        & & $10^3$   &  1.12 & 6.54 & 4.51 \\
        & & $10^4$   &  1.11 & 9.42 & 8.24 \\
    \bottomrule
  \end{tabular}
\end{table}

\begin{table}[H]
  \caption{Simulated model with 2 high-cardinality categorical features and a spatial feature with 2-D locations with a RBF kernel. Estimated variance components on average.}
  \label{tab:simulated:spatialcategorical:sigmas}
  \centering
  \begin{tabular}{lll|lllll}
\toprule
$\sigma^2_{b}$ & $\sigma^2_{c}$ & $\sigma^2_{d_0}$ & $\hat{\sigma}^2_{e}$ & $\hat{\sigma}^2_{b}$ & $\hat{\sigma}^2_{c}$ & $\hat{\sigma}^2_{d_0}$ & $\hat{\sigma}^2_{d_1}$ \\
	\midrule
    0.3 & 0.3 & 0.3 & 1.12 & 0.30 & 0.31 & 0.28 & 1.06 \\
    & & 3.0 & 1.14 & 0.29 & 0.29 & 2.98 & 0.95 \\
    & 3.0 & 0.3 & 1.12 & 0.30 & 3.03 & 0.30 & 1.03 \\
    & & 3.0 & 1.12 & 0.29 & 3.04 & 3.06 & 1.04 \\
	\midrule
    3.0 & 0.3 & 0.3 & 1.13 & 2.97 & 0.31 & 0.28 & 1.03 \\
    & & 3.0 & 1.12 & 2.98 & 0.30 & 2.76 & 0.97 \\
    & 3.0 & 0.3 & 1.13 & 3.05 & 3.10 & 0.32 & 1.16 \\
    & & 3.0 & 1.14 & 2.91 & 2.90 & 3.01 & 0.98 \\
\bottomrule
\end{tabular}
\end{table}

\begin{table}[ht]
  \caption{Simulated binary GLMM model with a single categorical feature, estimated variance components on average.}
  \label{tab:simulated:glmm:sigmas}
  \centering
  \begin{tabular}{llll}
    \toprule
    $\sigma^2_b$ & $q$ & lme4 & LMMNN \\
    	\midrule
    0.1 & $10^2$  & 0.06 & 0.08 \\
        & $10^3$  & 0.06 & 0.09 \\
        & $10^4$  & 0.06 & 0.1 \\
    \midrule
    1 & $10^2$ & 0.52 & 0.77 \\
        & $10^3$ & 0.6 & 0.87 \\
        & $10^4$ & 0.56 & 0.95 \\
    \midrule
    10 & $10^2$ & 6.47 & 3.64 \\
        & $10^3$ & 6.22 & 3.59 \\
        & $10^4$ & 5.77 & 5.35 \\
    \bottomrule
  \end{tabular}
\end{table}

\section{Mean runtime and number of epochs}
\label{appx:tables:times}

\begin{table}[H]
  \caption{Simulated model with a single categorical feature, mean runtime (minutes) and number of epochs in parentheses.}
  \label{tab:simulated:single:times}
  \centering
  \begin{tabular}{ll|llllll}
\toprule
\multicolumn{8}{c}{\textbf{g(Z) = Z}} \\
\midrule
$\sigma^2_b$ & $q$ & Ignore & OHE & Embeddings & lme4 & MeNets & LMMNN \\
	\midrule
0.1 & $10^2$  & 0.5 (26) & 0.7 (31) & 0.7 (24) & 0.01 (--) & 26.4 (96) & 2.2 (40) \\
    & $10^3$  & 0.7 (35) & 0.6 (16) & 0.6 (20) & 0.01 (--) & 48.3 (259) & 2.9 (56) \\
    & $10^4$ & 0.6 (29) & 1.4 (12) & 0.4 (14) & 0.02 (--) & 47.5 (275) & 2.4 (43) \\
\midrule
1 & $10^2$   & 0.5 (22) & 0.6 (31) & 0.7 (26) & 0.01 (--) & 21.2 (82) & 2.5 (47) \\
    & $10^3$ & 0.8 (34) & 0.4 (16) & 0.6 (21) & 0.01 (--) & 79.3 (434) & 2.5 (47) \\
    & $10^4$ & 0.5 (26) & 1.7 (13) & 0.5 (15) & 0.02 (--) & 51.1 (300) & 2.2 (41) \\
\midrule
10 & $10^2$ & 0.6 (32) & 0.4 (20) & 0.8 (29) & 0.01 (--) & 17.6 (65) & 2.1 (41) \\
    & $10^3$ & 0.6 (31) & 0.5 (18) & 0.6 (21) & 0.01 (--) & 34.5 (196) & 2.2 (37) \\
    & $10^4$ & 0.7 (33) & 1.8 (16) & 0.6 (20) & 0.02 (--) & 50.9 (300) & 2.8 (50) \\
	\midrule
	\multicolumn{8}{c}{\textbf{g(Z) = ZW}} \\
	\midrule
    0.1 & $10^2$ & 0.8 (33) & 0.5 (24) & 1.0 (37) & 0.01 (--) & 13.3 (63) & 1.8 (31) \\
        & $10^3$ & 0.8 (42) & 0.5 (17) & 0.6 (22) & 0.01 (--) & 44.1 (279) & 0.9 (14) \\
        & $10^4$ & 0.8 (36) & 1.6 (17) & 1.0 (30) & 0.02 (--) & 54.9 (300) & 1.5 (17) \\
    \midrule
    1 & $10^2$ & 0.7 (32) & 0.4 (19) & 0.7 (26) & 0.01 (--) & 15.3 (76) & 2.3 (42) \\
        & $10^3$ & 0.7 (34) & 0.9 (33) & 1.0 (37) & 0.01 (--) & 23.9 (148) & 1.0 (16) \\
        & $10^4$ & 0.7 (33) & 2.0 (21) & 0.7 (26) & 0.02 (--) & 27.1 (146) & 2.2 (27) \\
    \midrule
    10 & $10^2$ & 0.7 (34) & 0.6 (24) & 0.6 (21) & 0.01 (--) & 10.8 (55) & 2.5 (46) \\
        & $10^3$ & 0.5 (25) & 0.4 (14) & 0.3 (11) & 0.01 (--) & 2.9 (17) & 1.4 (24) \\
        & $10^4$ & 0.4 (18) & 2.2 (23) & 0.7 (22) & 0.02 (--) & 6.0 (32) & 3.1 (39) \\
        \midrule
        \multicolumn{8}{c}{\textbf{g(Z) = ZW * cos(ZW)}} \\
	\midrule
    0.1 & $10^2$ & 0.8 (35) & 0.7 (28) & 1.0 (29) & 0.02 (--) & 20.7 (97) & 1.6 (30) \\
        & $10^3$ & 1.0 (44) & 0.6 (17) & 0.6 (20) & 0.01 (--) & 20.0 (148) & 0.8 (14) \\
        & $10^4$ & 0.8 (34) & 1.7 (17) & 0.9 (27) & 0.02 (--) & 51.7 (300) & 1.6 (14) \\
    \midrule
    1 & $10^2$ & 0.9 (39) & 0.6 (24) & 0.7 (23) & 0.01 (--) & 12.5 (68) & 1.9 (38) \\
        & $10^3$ & 0.7 (29) & 0.7 (25) & 1.0 (32) & 0.01 (--) & 23.7 (172) & 0.9 (15) \\
        & $10^4$ & 0.7 (29) & 1.9 (21) & 1.5 (47) & 0.01 (--) & 23.0 (146) & 2.5 (23) \\
    \midrule
    10 & $10^2$ & 0.7 (30) & 0.7 (29) & 1.1 (35) & 0.02 (--) & 13.8 (58) & 2.0 (41) \\
        & $10^3$ & 0.6 (26) & 0.4 (12) & 0.4 (11) & 0.01 (--) & 2.1 (15) & 1.1 (20) \\
        & $10^4$ & 0.4 (18) & 2.5 (32) & 0.5 (15) & 0.01 (--) & 3.1 (19) & 3.6 (35) \\
\bottomrule
\end{tabular}

\end{table}

\begin{table}[H]
  \caption{Simulated model with 3 categorical features, mean runtime (minutes) and number of epochs in parentheses.}
  \label{tab:simulated:multiple:times}
  \centering
  \begin{tabular}{lll|lllll}
\toprule
\multicolumn{8}{c}{\textbf{g(Z) = Z}} \\
\midrule
$\sigma^2_{b1}$ & $\sigma^2_{b2}$ & $\sigma^2_{b3}$ & Ignore & OHE & Embed. & lme4 & LMMNN \\
	\midrule
    0.3 & 0.3 & 0.3 & 0.8 (40) & 1.0 (16) & 0.6 (16) & 0.07 (--) & 4.1 (65) \\
    &  & 3.0 & 0.7 (31) & 1.1 (17) & 0.6 (16) & 0.07 (--) & 3.2 (44) \\
    & 3.0 & 0.3 & 0.6 (30) & 1.2 (21) & 0.6 (16) & 0.07 (--) & 3.4 (45) \\
    &  & 3.0 & 0.5 (24) & 1.2 (20) & 0.7 (19) & 0.07 (--) & 3.7 (47) \\
	\midrule
    3.0 & 0.3 & 0.3 & 0.8 (36) & 1.1 (18) & 0.7 (17) & 0.08 (--) & 3.3 (38) \\
    &  & 3.0 & 0.6 (27) & 1.2 (21) & 0.8 (20) & 0.06 (--) & 3.9 (45) \\
    & 3.0 & 0.3 & 0.6 (27) & 1.3 (23) & 0.8 (20) & 0.07 (--) & 4.4 (50) \\
    &  & 3.0 & 0.7 (32) & 1.1 (19) & 0.7 (18) & 0.07 (--) & 4.4 (49) \\
	\midrule
	\multicolumn{8}{c}{\textbf{g(Z) = ZW}} \\
	\midrule
    0.3 & 0.3 & 0.3 & 0.6 (28) & 1.1 (18) & 1.5 (40) & 0.1 (--) & 1.2 (16) \\
    &  & 3.0 & 0.6 (27) & 1.0 (15) & 0.6 (14) & 0.1 (--) & 2.2 (33) \\
    & 3.0 & 0.3 & 0.7 (34) & 0.9 (14) & 0.6 (16) & 0.1 (--) & 1.9 (29) \\
    &  & 3.0 & 0.5 (25) & 1.6 (32) & 0.8 (20) & 0.1 (--) & 2.4 (37) \\
	\midrule
    3.0 & 0.3 & 0.3 & 0.5 (22) & 1.0 (17) & 0.9 (24) & 0.1 (--) & 1.5 (22) \\
    &  & 3.0 & 0.6 (27) & 1.0 (14) & 0.5 (12) & 0.1 (--) & 2.1 (33) \\
    & 3.0 & 0.3 & 0.7 (32) & 0.9 (13) & 0.5 (12) & 0.1 (--) & 2.1 (32) \\
    &  & 3.0 & 0.5 (23) & 0.9 (13) & 0.5 (13) & 0.1 (--) & 2.8 (45) \\
    \midrule
	\multicolumn{8}{c}{\textbf{g(Z) = ZW * cos(ZW)}} \\
	\midrule
    0.3 & 0.3 & 0.3 & 0.8 (35) & 1.2 (21) & 1.3 (30) & 0.1 (--) & 1.2 (15) \\
    &  & 3.0 & 0.5 (22) & 1.1 (17) & 0.5 (12) & 0.1 (--) & 1.6 (24) \\
    & 3.0 & 0.3 & 0.6 (27) & 0.9 (14) & 0.7 (16) & 0.1 (--) & 1.4 (20) \\
    &  & 3.0 & 0.8 (33) & 0.9 (13) & 0.7 (16) & 0.1 (--) & 1.9 (28) \\
	\midrule
    3.0 & 0.3 & 0.3 & 0.9 (31) & 0.9 (13) & 1.4 (33) & 0.1 (--) & 1.2 (17) \\
    &  & 3.0 & 0.6 (25) & 1.0 (15) & 0.7 (15) & 0.1 (--) & 1.7 (26) \\
    & 3.0 & 0.3 & 0.7 (30) & 1.1 (18) & 1.0 (24) & 0.1 (--) & 1.6 (25) \\
    &  & 3.0 & 0.6 (25) & 1.2 (21) & 1.0 (23) & 0.1 (--) & 2.1 (32) \\
\bottomrule
\end{tabular}

\end{table}

\begin{table}[H]
  \caption{Simulated model with longitudinal data, mean runtime (minutes) and number of epochs in parentheses.}
  \label{tab:simulated:longitudinal:times}
  \centering
  \begin{tabular}{lll|llllll}
\toprule
\multicolumn{9}{c}{\textbf{Mode: Random}} \\
\midrule
$\sigma^2_{b0}$ & $\sigma^2_{b1}$ & $\sigma^2_{b2}$ & Ignore & OHE & Embed. & lme4 & LSTM & LMMNN \\
	\midrule
    0.3 & 0.3 & 0.3 & 0.9 (42) & 1.4 (13) & 0.5 (15) & 0.9 (--) & 27.1 (104) & 2.9 (47) \\
    &  & 3.0 & 0.7 (32) & 1.4 (13) & 0.4 (15) & 0.6 (--) & 31.2 (110) & 3.0 (50) \\
    & 3.0 & 0.3 & 0.7 (29) & 1.4 (13) & 0.4 (14) & 0.7 (--) & 35.8 (118) & 2.8 (46) \\
    &  & 3.0 & 0.8 (34) & 1.4 (13) & 0.5 (15) & 0.7 (--) & 29.9 (107) & 2.5 (41) \\
	\midrule
    3.0 & 0.3 & 0.3 & 0.7 (30) & 1.4 (13) & 0.5 (16) & 0.6 (--) & 30.1 (104) & 3.5 (58) \\
    &  & 3.0 & 0.6 (25) & 1.5 (14) & 0.5 (16) & 0.4 (--) & 35.3 (132) & 3.3 (55) \\
    & 3.0 & 0.3 & 0.6 (25) & 1.4 (14) & 0.5 (16) & 0.8 (--) & 41.5 (147) & 3.8 (63) \\
    &  & 3.0 & 0.6 (28) & 1.5 (14) & 0.5 (16) & 0.5 (--) & 41.0 (128) & 2.8 (45) \\
	\midrule
	\multicolumn{9}{c}{\textbf{Mode: Future}} \\
	\midrule
    0.3 & 0.3 & 0.3 & 0.5 (22) & 1.5 (13) & 0.5 (15) & 1.68 (--) & 39.1 (129) & 3.0 (48) \\
    &  & 3.0 & 0.7 (31) & 1.4 (13) & 0.4 (15) & 1.38 (--) & 32.5 (114) & 3.1 (51) \\
    & 3.0 & 0.3 & 0.6 (27) & 1.5 (13) & 0.4 (15) & 1.21 (--) & 29.7 (102) & 2.8 (46) \\
    &  & 3.0 & 0.8 (36) & 1.4 (13) & 0.4 (14) & 1.39 (--) & 40.1 (133) & 2.8 (45) \\
	\midrule
    3.0 & 0.3 & 0.3 & 0.7 (32) & 1.5 (14) & 0.5 (17) & 1.29 (--) & 35.6 (116) & 3.4 (55) \\
    &  & 3.0 & 0.7 (31) & 1.5 (14) & 0.5 (17) & 1.14 (--) & 42.0 (154) & 3.3 (54) \\
    & 3.0 & 0.3 & 0.7 (33) & 1.5 (14) & 0.5 (17) & 0.81 (--) & 32.0 (118) & 3.0 (48) \\
    &  & 3.0 & 1.0 (43) & 1.5 (13) & 0.6 (18) & 1.22 (--) & 42.2 (140) & 3.1 (51) \\
\bottomrule
\end{tabular}
\end{table}

\begin{table}[H]
  \caption{Simulated model with spatial data with a RBF kernel. Mean runtime (minutes) and number of epochs in parentheses.}
  \label{tab:simulated:spatial:times}
  \centering
  \begin{tabular}{lll|llllll}
\toprule
$\sigma^2_{b0}$ & $\sigma^2_{b1}$ & $q$ & OHE & Embed. & CNN & SVDKL & LMMNN-E & LMMNN-R \\
	\midrule
    0.1 & 0.1 & $10^2$ & 0.8 (34) & 0.7 (25) & 3.9 (31) & 8.9 (46) & 2.4 (36) & 3.0 (55) \\
    & & $10^3$ & 0.6 (20) & 0.6 (21) & 6.4 (52) & 8.0 (41) & 3.5 (54) & 2.8 (53) \\
    & & $10^4$ & 1.5 (14) & 0.5 (16) & 3.6 (27) & 9.6 (44) & 2.9 (45) & 3.2 (47) \\
    \cline{2-9}
    & 1.0 & $10^2$ & 0.7 (30) & 1.2 (44) & 4.4 (36) & 10.1 (51) & 3.0 (46) & 2.3 (43) \\
    & & $10^3$ & 0.5 (17) & 0.6 (21) & 5.5 (45) & 5.8 (29) & 3.3 (51) & 2.9 (54) \\
    & & $10^4$ & 1.4 (13) & 0.4 (15) & 3.3 (26) & 7.8 (36) & 1.8 (28) & 3.8 (59) \\
    \cline{2-9}
    & 10.0 & $10^2$ & 0.7 (33) & 0.7 (26) & 3.8 (31) & 6.3 (32) & 2.7 (42) & 2.4 (45) \\
    & & $10^3$ & 0.5 (17) & 0.6 (22) & 3.4 (28) & 5.8 (29) & 4.0 (62) & 1.9 (34) \\
    & & $10^4$ & 1.4 (13) & 0.5 (16) & 3.3 (25) & 8.4 (37) & 2.5 (38) & 2.8 (42) \\
    \midrule
    1.0 & 0.1 & $10^2$ & 0.6 (28) & 0.8 (28) & 5.3 (43) & 7.8 (40) & 2.3 (35) & 2.1 (38) \\
    & & $10^3$ & 0.5 (19) & 0.8 (29) & 6.2 (50) & 14.1 (71) & 5.0 (79) & 2.7 (48) \\
    & & $10^4$ & 1.5 (15) & 0.5 (19) & 7.1 (56) & 16.5 (76) & 5.7 (92) & 3.5 (55) \\
    \cline{2-9}
    & 1.0 & $10^2$ & 0.5 (22) & 1.0 (37) & 4.6 (37) & 7.7 (39) & 3.2 (51) & 2.3 (42) \\
    & & $10^3$ & 0.5 (18) & 0.7 (25) & 5.8 (47) & 8.8 (45) & 2.3 (36) & 3.0 (55) \\
    & & $10^4$ & 1.5 (15) & 0.5 (17) & 4.3 (34) & 9.8 (45) & 2.7 (42) & 2.4 (36) \\
    \cline{2-9}
    & 10.0 & $10^2$ & 0.5 (21) & 0.8 (27) & 5.0 (40) & 7.3 (37) & 3.7 (60) & 2.6 (47) \\
    & & $10^3$ & 0.6 (19) & 0.7 (26) & 5.2 (43) & 8.9 (45) & 3.6 (58) & 2.5 (44) \\
    & & $10^4$ & 1.4 (13) & 0.5 (17) & 5.5 (43) & 7.1 (33) & 3.0 (47) & 3.4 (56) \\
    \midrule
    10.0 & 0.1 & $10^2$ & 0.6 (26) & 1.4 (48) & 4.6 (38) & 8.2 (41) & 4 (60) & 3.4 (53) \\
    & & $10^3$ & 0.6 (21) & 1.0 (36) & 6.4 (52) & 29.0 (144) & 8.5 (127) & 2.9 (49) \\
    & & $10^4$ & 1.6 (16) & 0.9 (30) & 10.8 (85) & 51.2 (236) & 23.1 (100) & 2.7 (41) \\
    \cline{2-9}
    & 1.0 & $10^2$ & 0.7 (34) & 1.1 (39) & 5.8 (47) & 5.4 (29) & 3.9 (58) & 2.6 (45) \\
    & & $10^3$ & 0.7 (24) & 1.0 (34) & 6.2 (51) & 11.6 (60) & 5.4 (81) & 3.2 (58) \\
    & & $10^4$ & 1.9 (22) & 0.9 (31) & 7.9 (62) & 10.6 (49) & 4.4 (66) & 2.9 (45) \\
    \cline{2-9}
    & 10.0 & $10^2$ & 0.5 (24) & 1.2 (43) & 5.2 (42) & 5.8 (34) & 4.0 (60) & 2.7 (49) \\
    & & $10^3$ & 0.8 (28) & 0.7 (25) & 5.2 (42) & 5.7 (33) & 2.8 (40) & 2.4 (42) \\
    & & $10^4$ & 1.9 (22) & 1.0 (34) & 5.4 (42) & 6.3 (30) & 4.0 (61) & 3.1 (51) \\
\bottomrule
\end{tabular}
\end{table}

\begin{table}[H]
  \caption{Simulated model with 2 high-cardinality categorical features and a spatial feature with 2-D locations with a RBF kernel. Mean runtime (minutes) and number of epochs in parentheses.}
  \label{tab:simulated:spatialcategorical:times}
  \centering
  \begin{tabular}{lll|llll}
\toprule
$\sigma^2_{b}$ & $\sigma^2_{c}$ & $\sigma^2_{d_0}$ & Ignore & OHE & Embed. & LMMNN \\
	\midrule
  0.3 & 0.3 & 0.3 & 0.7 (32) & 2.3 (15) & 0.7 (18) & 3.7 (53) \\
  & & 3.0 & 1.3 (62) & 2.7 (19) & 0.8 (22) & 3.1 (42) \\
  & 3.0 & 0.3 & 0.8 (37) & 2.7 (18) & 1.0 (26) & 3.8 (57) \\
  & & 3.0 & 1.5 (72) & 3.1 (22) & 1.1 (29) & 3.7 (54) \\
	\midrule
  3.0 & 0.3 & 0.3 & 0.7 (30) & 2.8 (17) & 0.9 (22) & 3.6 (52) \\
  & & 3.0 & 1.8 (84) & 3.3 (24) & 1.0 (27) & 3.6 (54) \\
  & 3.0 & 0.3 & 0.7 (34) & 2.9 (18) & 0.9 (23) & 2.8 (39) \\
  & & 3.0 & 1.4 (66) & 2.8 (17) & 1.1 (30) & 3.3 (47) \\
\bottomrule
\end{tabular}
\end{table}

\begin{table}[H]
  \caption{Simulated binary GLMM model with a single categorical feature. Mean runtime (minutes) and number of epochs in parentheses.}
  \label{tab:simulated:glmm:times}
  \centering
  \begin{tabular}{llllllll}
\toprule
$\sigma^2_b$ & $q$ & Ignore & OHE & Embeddings & lme4 & LMMNN \\
	\midrule
0.1 & $10^2$   & 2.1 (20) & 2.1 (20) & 3.0 (20) & 1.3 (--) & 8.4 (21) \\
    & $10^3$  & 2.4 (23) & 1.6 (14) & 2.1 (14) & 1.4 (--) & 10.8 (23) \\
    & $10^4$ & 2.9 (28) & 2.3 (11) & 1.9 (12) & 2.8 (--) & 28.4 (25) \\
\midrule
1 & $10^2$ & 2.8 (26) & 2.4 (22) & 3.8 (26) & 1.2 (--) & 12.5 (30) \\
    & $10^3$  & 2.7 (26) & 1.6 (14) & 2.4 (16) & 1.7 (--) & 15.9 (34) \\
    & $10^4$ & 3.2 (30) & 2.3 (11) & 2.0 (12) & 9.4 (--) & 35.3 (31) \\
\midrule
10 & $10^2$ & 3.2 (30) & 2.6 (24) & 3.6 (24) & 1.4 (--) & 9.4 (23) \\
    & $10^3$ & 2.8 (27) & 1.6 (14) & 2.5 (16) & 1.9 (--) & 17.5 (38) \\
    & $10^4$ & 3.1 (30) & 2.5 (12) & 2.0 (13) & 6.4 (--) & 33.5 (28) \\
\bottomrule
\end{tabular}
\end{table}

\begin{table}[H]
  \caption{Real datasets with $K$ categorical features: mean runtime (minutes) and number of epochs in parentheses.}
  \label{tab:real:multiple:times}
  \centering
  \begin{tabular}{llllll}
    \toprule
    Dataset & Ignore & OHE & Embed. & lme4 & LMMNN \\
    \midrule
    Imdb       &   0.9 (38) & -- & 0.9 (28) & 0.6 (--) & 2.41 (31) \\
    News       &   0.5 (16) & -- & 0.5 (23) & 0.7 (--) & 1.4 (25) \\
    InstEval   &   0.3 (37) & 0.5 (40) & 0.9 (55) & 0.2 (--) & 1.5 (23) \\
    Spotify    &  0.2 (13) & -- & 0.3 (39) & 0.1 (--) & 1.0 (39) \\
    UKB-blood  &  0.5 (28) & 0.4 (11) & 0.5 (12) & 0.6 (--) & 2.0 (34) \\
    \bottomrule
 \end{tabular}
\end{table}

\begin{table}[H]
  \caption{Longitudinal datasets: mean runtime (minutes) and number of epochs in parentheses.}
  \label{tab:real:longitudinal:times}
  \centering
  \begin{tabular}{lllllll}
    \toprule
    \multicolumn{7}{c}{\textbf{Mode: Random}} \\
    \midrule
    Dataset & Ignore & OHE & Embed. & lme4 & LSTM & LMMNN \\
    \midrule
    Rossmann & 1.3 (100) & 0.4 (24) & 0.9 (54) & 0.1 (--) & 8.6 (32) & 1.3 (42) \\
    AUimport &  1.4 (34) & 1.2 (17) & 1.8 (36) & 0.1 (--) & 15.8 (139) & 2.9 (34) \\
    UKB SBP & 5.0 (36) & -- & 4.6 (24) & 1.4 (--) & -- & 14.2 (49) \\
    \midrule
    \multicolumn{7}{c}{\textbf{Mode: Future}} \\
    \midrule
    Rossmann &  1.0 (90) & 0.3 (23) & 0.5 (38) & 0.1 (--)  & 10.0 (37) & 1.1 (46)\\
    AUimport &  1.3 (43) & 2.8 (55) & 1.9 (48) & 0.1 (--)  & 14.4 (129) & 2.4 (36) \\
    UKB SBP & 3.6 (33) & -- & 4.3 (28) & 1.3 (--) & -- & 11.6 (47) \\
    \bottomrule
 \end{tabular}
\end{table}

\begin{table}[H]
  \caption{Spatial datasets with an optionally high-cardinality categorical feature: mean runtime (minutes) and number of epochs in parentheses.}
  \label{tab:real:spatial:times}
  \centering
  \begin{tabular}{lllllll}
    \toprule
    \multicolumn{7}{c}{\textbf{Without high-cardinality categorical features}} \\
    \midrule
    Dataset & Ignore & Embed. & CNN & SVDKL & LMMNN-E & LMMNN-R \\
    \midrule
    Income     & 1.0 (55) & 1.2 (54) & 4.3 (44) & 17.6 (130) & 2.0 (46) & 1.8 (29) \\
    Asthma     & 0.7 (41) & 1.0 (46) & 2.5 (25) & 12.6 (109) & 1.6 (35) & 2.5 (25) \\
    AirQuality & 0.8 (39) & 1.5 (56) & 3.4 (34) & 24.6 (162) & 2.5 (51) & 1.5 (28) \\
    Radiation  & 1.3 (8) & 4.1 (21) & 13.2 (15) & 34.5 (30) & 6.9 (18) & 5.1 (10) \\
    Airbnb     & 0.4 (26) & 0.2 (12) & 3.0 (43) & 4.2 (42) & 1.3 (41) & 0.7 (18) \\
    Cars       & 1.3 (52) & 1.5 (47) & 4.7 (34) & 9.6 (54) & 3.7 (50) & 6.9 (78) \\
    \midrule
    \multicolumn{7}{c}{\textbf{With high-cardinality categorical features}} \\
    \midrule
    Airbnb     &  0.4 (33) & 1.5 (79) & -- & -- & -- & 2.8 (22) \\
    Cars       &  0.8 (26) & 1.8 (38) & -- & -- & -- & 6.5 (69) \\
    \bottomrule
 \end{tabular}
\end{table}

\begin{table}[H]
  \caption{Spatial datasets with an optionally high-cardinality categorical feature: mean runtime (minutes) and number of epochs in parentheses.}
  \label{tab:real:glmm:times}
  \centering
  \begin{tabular}{llllll}
    \toprule
    Dataset & Ignore & OHE & Embed. & lme4 & LMMNN \\
    \midrule
    Airbnb       & 0.4 (24) & -- & 0.3 (13) & -- & 8.9 (35) \\
    Cars       & 0.70 (21) & 2.5 (13) & 0.6 (12) & 133 (--) & 5.5 (32) \\
    \bottomrule
 \end{tabular}
\end{table}

\section{Real datasets additional details}\label{appx:realdetails}
\begin{table}[H]
  \caption{Real datasets description: Part I}
  \label{tab:real:datasets}
  \centering

\begin{tabular}{p{0.1\linewidth}p{0.15\linewidth}p{0.11\linewidth}p{0.15\linewidth}p{0.49\linewidth}}
    \toprule
    \multicolumn{5}{c}{\textbf{Multiple categorical features}} \\
    \midrule
    Dataset & Source & Availability & Reference & Description \\
    \midrule
    Imdb & Kaggle & Free & \citet{wrandrall_2021} & 86K movie titles scraped from imdb.com along with their genre, director, date of release a 1-10 mean score and a textual description which is processed to top 1-gram tokens count, see ETL. \\
    News & UCI ML & Free & \citet{Moniz2018} & 81K news items and their number of shares on Facebook. Headline is processed to top 1-gram tokens count, see ETL. \\
    InstEval & lme4 & Free & \citet{lme4}& 73K students 1-5 evaluations of professors from ETH Zurich \\
    Spotify & Tidy Tuesday & Free & \citet{tidytuesday} & 28K songs with their date release, genre, artist, album as well as 12 audio features from which we chose to predict the first one, danceability. \\
    UKB-blood & UK Biobank & Authorized & \citet{UKB} & Subset of 42K UK Biobank with cancer history. To predict triglycerides and other chemicals level in blood we use features such as gender, age, height, weight, skin color and more, see ETL.\\
    \midrule
    \multicolumn{5}{c}{\textbf{Longitudinal data and repeated measures}} \\
    \midrule
    Rossmann & Kaggle & Free & \citet{rossmann} & Total monthly sales in \$ from over 1.1K stores around Europe. Features include month, number of holiday days, number of days with promotion and more, see ETL.\\
    AUimport & Kaggle & Free & \citet{UN_commodity, owidco2andothergreenhousegasemissions} & Total yearly import in \$ of 5K commodities in Australia 1988-2016. Features come by joining to various yearly data from ourworldindata.org such as surface temperature, population size, CO2 emissions and wheat yield. See ETL. \\
    UKB-SBP & UK Biobank & Authorized & \citet{UKB} & 469K subjects of the UK Biobank cohort for which we have 1-4 SBP measures. Time-varying features include gender, age, height, different food intakes, smoking habits and many more, see ETL.\\
    \bottomrule
 \end{tabular}
\end{table}
\begin{table}[H]
  \caption{Real datasets description: Part II}
  \label{tab:real:datasets2}
  \centering
  \begin{tabular}{p{0.1\linewidth}p{0.15\linewidth}p{0.11\linewidth}p{0.15\linewidth}p{0.49\linewidth}}
    \toprule
    \multicolumn{5}{c}{\textbf{Spatial data and spatial-categorical combinations}} \\
    \midrule
    Income & Kaggle & Free & \citet{us_census} & Mean yearly income in \$ for 71K US census tracts, data was previously downloaded from the US Census Bureau. In addition to longitude and latitude features include population size, share of men, rate of employment and more, see ETL.\\
    Asthma & CDC & Free & \cite{asthma} & Adult asthma rate in 69K US census tracts according to CDC in 2019. Additional features come from the income data, see ETL. \\
    AirQuality & CDC & Free & \citet{pm25} & PM2.5 particles level in 71K US census tracts according to CDC, on 1/1/2016. Additional features come from the income data, see ETL. \\
    Radiation & Kaggle & Free & \citet{Radiation} & A 10\% sample from 6.5M radiation measurements in over 1K locations in Japan in 2017 by Safecast. \\
    Airbnb & Google Drive & Free & \citet{kalehbasti2019airbnb} & 50K Airbnb listings in NYC scraped by \citet{kalehbasti2019airbnb}, ETL follows their steps exactly. In addition to longitude and latitude, features include floor number, neighborhood, is there a bathtub, some top 1-ngram tokens counts from description and more, see ETL.\\
    Cars & Kaggle & Free & \citet{Cars} & 97K cars and trucks with unique VIN from Cragslist and their price in \$, price was filtered from 1K\$ to 300K\$. In addition to longitude and latitude, features include manufacturer, year of make, size, condition and more, see ETL. \\
    \bottomrule
 \end{tabular}
\end{table}

\section{Additional figures}\label{appx:realplots}
\begin{figure}[H]
	\centering
	\includegraphics[width=0.9\linewidth]{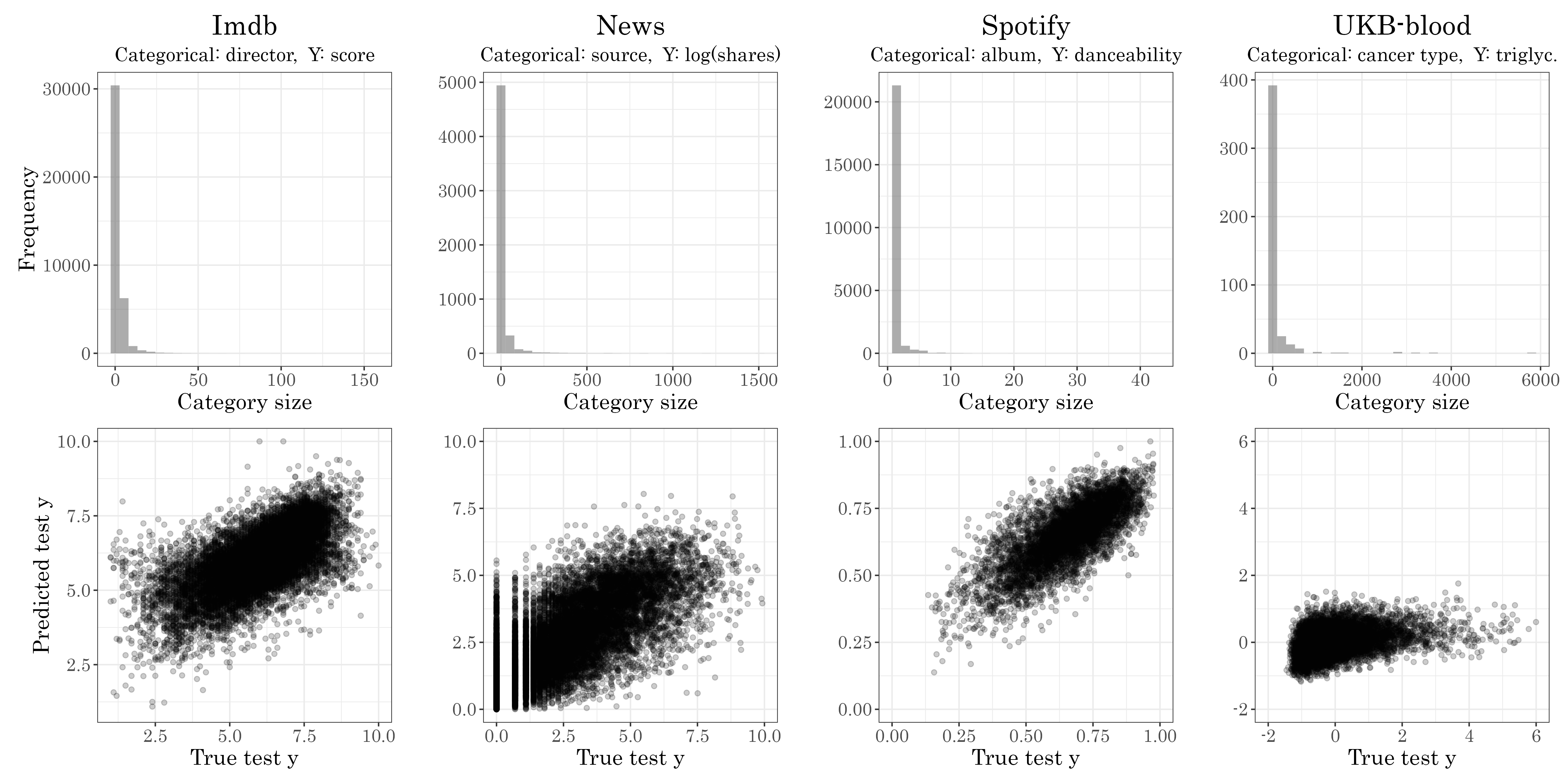}
	\caption{Selected multiple categorical datasets predicted vs. true results and category size distribution, only one categorical feature is presented.}
	\label{fig:real_viz_categorical}
\end{figure}

\begin{figure}[H]
	\centering
	\includegraphics[width=0.67\linewidth]{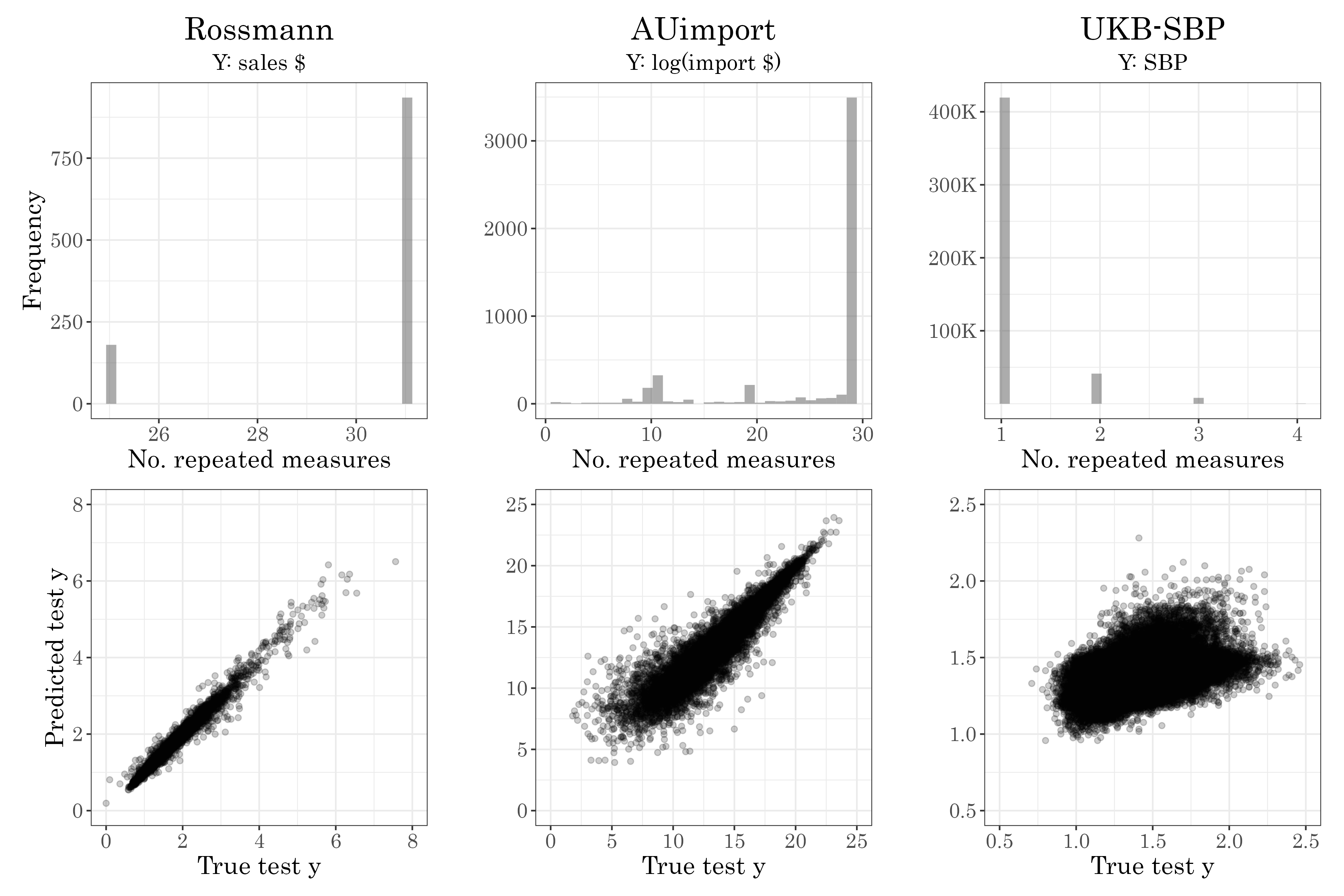}
	\caption{Longitudinal datasets predicted vs. true results and number of repeated measures distribution.}
	\label{fig:real_viz_longitudinal}
\end{figure}

\begin{figure}[H]
	\centering
	\includegraphics[width=0.9\linewidth]{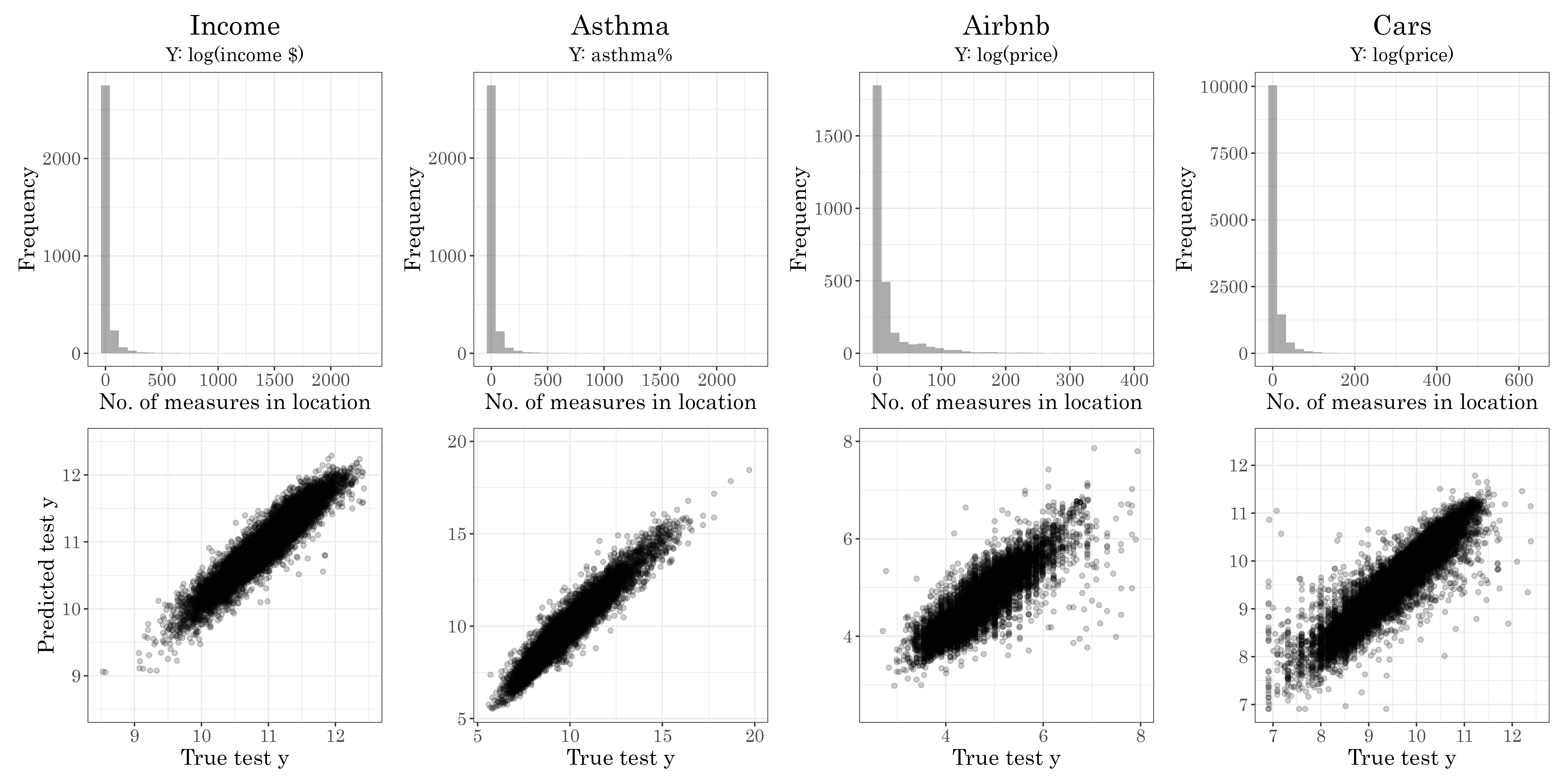}
	\caption{Selected spatial datasets predicted vs. true results and number of measurements in location distribution.}
	\label{fig:real_viz_spatial}
\end{figure}

% Note: in this sample, the section number is hard-coded in. Following
% proper LaTeX conventions, it should properly be coded as a reference:

%In this Appendix we prove the following theorem from
%Section~\ref{sec:textree-generalization}:
\nocite{*}
\vskip 0.2in
\newpage
\bibliography{references}

\end{document}